\documentclass{article}

\usepackage[preprint]{neurips_2026}

\usepackage[utf8]{inputenc} 
\usepackage[T1]{fontenc}    
\usepackage{hyperref}       
\usepackage{url}            
\usepackage{booktabs}       
\usepackage{amsfonts}       
\usepackage{nicefrac}       
\usepackage{microtype}      
\usepackage{xcolor}         
\usepackage{graphicx}
\usepackage{subcaption}
\usepackage[table]{xcolor}
\usepackage{amsmath}

\definecolor{rowblue}{HTML}{EAF2F8}

\title{Not All Tokens Learn Alike: Attention Entropy Reveals Heterogeneous Signals in RL Reasoning}

\author{
Gengyang Li$^{1,3}$,
Zheng-Fan Wu$^{1,\dagger}$,
Siqi Bao$^{1,\dagger}$,
Yunfang Wu$^{2,\dagger}$ \\
$^1$Baidu \\
$^2$School of Computer Science, Peking University \\
$^3$School of Software and Microelectronics, Peking University \\
\texttt{ligengyang@stu.pku.edu.cn} \\
\texttt{\{wuzhengfan, baosiqi\}@baidu.com}, \texttt{wuyf@pku.edu.cn}
}

\begin{document}

\maketitle

\begin{abstract}
Reinforcement-learning-based post-training has become a central paradigm for
improving the reasoning ability of large language models, yet the structure of
its token-level learning signals remains poorly understood. This work studies
such heterogeneity through attention entropy, a diagnostic that measures how
concentrated or diffuse the contextual support is for each response token.

We first show that token-level RL objectives are sparsely estimable: uniformly
random $20\%$ token subsets preserve a substantial fraction of full-token
held-out performance, suggesting considerable redundancy in token-level updates.
However, entropy-structured subsets exhibit markedly different optimization
behavior. Low-attention-entropy tokens, which we call anchors, rely on
concentrated support, produce stable gradients aligned with full-token updates,
and provide a reliable optimization backbone, but they tend to plateau on harder
benchmarks. High-attention-entropy tokens, which we call explorers, aggregate
more diffuse context and induce larger but directionally more volatile gradients.
Explorer-only training is unstable on average, yet the rare trajectories that
avoid collapse suggest that these tokens can contain useful hard-reasoning
signals when optimization remains stable.

We substantiate this anchor--explorer spectrum through evidence-gathering
analyses, entropy dynamics, gradient-geometry diagnostics, and controls showing
that position, predictive entropy, and loss normalization do not explain the
observed training asymmetry. Finally, a dynamic entropy-aware soft-reweighting
intervention improves Qwen3-8B-Base from $34.39$ to $37.40$ held-out average in
the strongest entropy-source setting, with representative shallow, middle, and
deep layers all outperforming full-token DAPO while favoring different
benchmarks. These findings suggest that attention entropy reveals
optimization-relevant structure in token-level RL signals, and that uniform token
averaging can obscure meaningful heterogeneity in RL-based reasoning
post-training.
\end{abstract}

\begin{figure}[t]
    \centering
    \includegraphics[width=\linewidth]{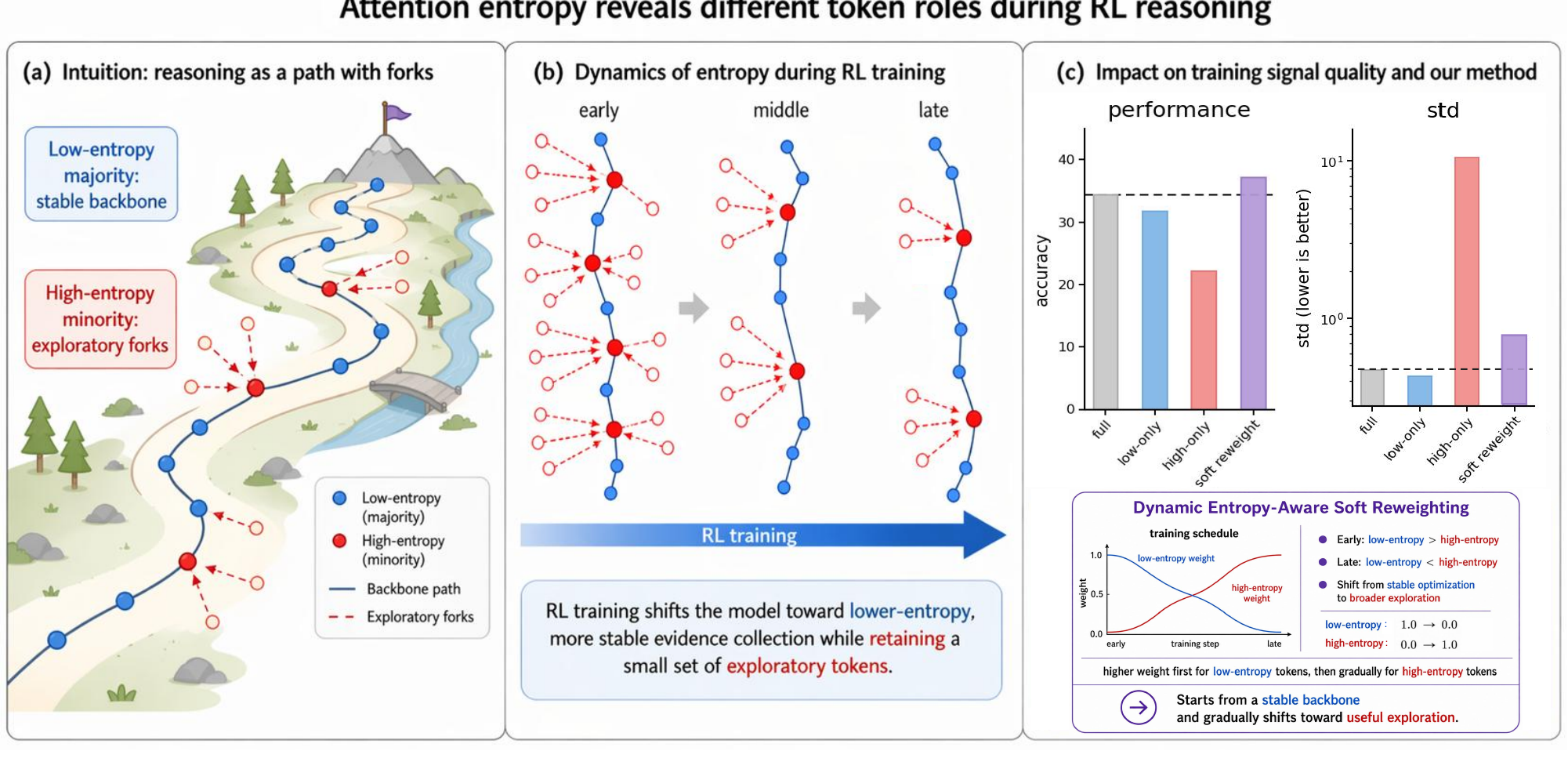}
    \caption{
        Attention entropy reveals an optimization spectrum in RL reasoning training.
        (a) Tokens span a spectrum from \textbf{anchor tokens} (low entropy, sparse selective support) to \textbf{explorer tokens} (high entropy, diffuse multi-position aggregation).
        (b) Selective training exposes the spectrum: anchor-only training is stable but plateaus; explorer-only training is fragile overall, while rare non-collapsed runs indicate potential signal that is difficult to use in isolation.
        (c) Full-token training implicitly averages over the spectrum. Entropy-aware soft reweighting is used as a validating intervention to test whether explicitly navigating this spectrum can combine anchor stability with explorer potential.
        }
    \label{fig:main_figure}
\end{figure}

\section{Introduction}

Reinforcement-learning-based post-training has become central to improving reasoning in large language models, but its token-level learning signal remains poorly understood. A single verifiable outcome reward is distributed across a long reasoning trace, yet the induced token updates need not have the same quality: some may preserve local fluency, others may stabilize intermediate states, and others may connect distant evidence. Uniform token averaging treats these contributions as exchangeable. We ask not only whether a sparse token subset can estimate the full objective, but which structured subsets provide stable, complementary, or volatile signal.

This question matters because sequence-level metrics can hide very different
token-level mechanisms. A run may look healthy while relying mostly on stable
but narrow updates, or it may destabilize when broader signals are emphasized
too early. Identifying which tokens supply reliable support and which tokens
carry harder but more volatile signal is therefore useful both for diagnosis and
for designing less brittle token allocation rules.

We study this question through \emph{attention entropy}, a forward-pass diagnostic of how concentrated or diffuse a token's contextual support is. Random $20\%$ token subsets first show that token-level RL objectives are \emph{sparsely estimable}; the main finding is that structured sparse subsets are not interchangeable. Attention entropy separates response tokens into an \emph{optimization spectrum} with two poles:

\begin{list}{\textbullet}{
    \setlength{\leftmargin}{1.2em}
    \setlength{\rightmargin}{0em}
    \setlength{\labelsep}{0.5em}
    \setlength{\labelwidth}{0.7em}
    \setlength{\itemindent}{0em}
    \setlength{\itemsep}{0.2em}
    \setlength{\topsep}{0.2em}
    \setlength{\parsep}{0pt}
}
    \item \textbf{Anchor tokens} (low attention entropy) rely on sparse selective support. They provide stable, well-aligned gradients and form a reliable optimization backbone, but anchor-only training tends to plateau on harder tasks.

    \item \textbf{Explorer tokens} (high attention entropy) aggregate diffuse multi-position context. Their gradients are strong but directionally volatile; explorer-only training usually collapses, yet non-collapsed trajectories suggest that these tokens can contain complementary hard-reasoning signal when optimization is stabilized.
\end{list}

This spectrum view reframes full-token training as an implicit average over heterogeneous token-level signals. We support it with evidence-gathering analyses, entropy dynamics, gradient geometry, and controls for position and next-token uncertainty, then use dynamic entropy-aware soft reweighting as a validating intervention rather than a tuned replacement for existing RL algorithms.

Our contributions:
\begin{list}{\textbullet}{
    \setlength{\leftmargin}{1.2em}
    \setlength{\rightmargin}{0em}
    \setlength{\labelsep}{0.5em}
    \setlength{\labelwidth}{0.7em}
    \setlength{\itemindent}{0em}
    \setlength{\itemsep}{0.2em}
    \setlength{\topsep}{0.2em}
    \setlength{\parsep}{0pt}
}
    \item We introduce attention entropy as a token-level diagnostic for RL reasoning and use it to identify an optimization spectrum with stable anchor tokens and volatile explorer tokens as its two poles.

    \item We provide optimization-spectrum evidence through selective-training outcomes, support-concentration statistics, entropy dynamics, and gradient geometry.

    \item We use control subsets to test alternative explanations based on position and predictive entropy, with additional controls in the appendix.

    \item We validate the preceding analysis with a simple entropy-aware soft-reweighting intervention, improving the held-out average on Qwen3-8B-Base from 34.39 to 37.40 in the strongest entropy-source setting under the strict exact-match evaluation protocol; representative shallow, mid, and deep entropy sources all outperform full-token DAPO but exhibit different benchmark profiles.
\end{list}

\section{Preliminaries and Training Setup}
\label{sec:background}

We study token-level training signals in RLVR using Qwen3 models optimized
with VeRL and the DAPO objective. Unless otherwise stated, our controlled
experiments use Qwen3-8B-Base, $n=8$ rollouts per prompt, DeepScaler plus
MATH level-3-to-5 prompts, and strict \texttt{\textbackslash boxed\{\}}
answer matching for evaluation. Full implementation and evaluation details
are provided in Appendix~\ref{app:experimental_details}. Let $x$ denote the
prompt and $y=(y_1,\ldots,y_T)$ denote the generated response. Our analysis
focuses on response tokens only.

We use two matched experiment families. Hard-masked selective training exposes
the role of token subsets by updating only selected response tokens, while the
full-data intervention replaces binary token selection with soft token weights
so that all valid response tokens remain in the objective. Therefore, absolute
scores are compared within a matched family.

Optimization uses a token-level objective:
\begin{equation}
\mathcal{L}_{\text{full}}=\frac{1}{T}\sum_{t=1}^{T}\ell_t,
\end{equation}
where $\ell_t$ denotes the effective loss at token $t$.

\paragraph{Attention entropy.} For response token $y_t$ at captured layer
$\ell$, we average attention probabilities over all heads in that layer to
obtain a layer-level attention distribution $a_{t,j}^{(\ell)}$ over visible
positions $j$. We define:
\begin{equation}
H_t^{\text{raw}}=-\sum_{j=1}^{N_t} a_{t,j}^{(\ell)}\log a_{t,j}^{(\ell)}, \qquad
H_t^{\text{norm}}=\frac{H_t^{\text{raw}}}{\log N_t}.
\end{equation}
The normalization removes the mechanical growth of raw entropy with visible
context length. We use $H_t^{\text{norm}}$ as the default metric and, unless the
entropy-source layer is explicitly varied, compute it from one fixed mid-layer
without cross-layer averaging. Details, variants, and layer-wise discussion are
in Appendices~\ref{app:entropy_details} and~\ref{app:layer_wise_attention}.
The fixed-layer choice is for controlled comparison rather than a claim of
optimality; the main intervention table also reports representative shallow and
deep entropy sources to show that the effect is not tied uniquely to Layer 20.

\paragraph{Token partition and naming.} For each response, we rank tokens by entropy \emph{within the same sample} and define:
\begin{list}{\textbullet}{
    \setlength{\leftmargin}{1.2em}
    \setlength{\itemsep}{0.1em}
    \setlength{\topsep}{0.2em}
    \setlength{\parsep}{0pt}
}
    \item \textbf{Anchor tokens}: the bottom $20\%$ by $H_t^{\text{norm}}$ within each sample.
    \item \textbf{Explorer tokens}: the top $20\%$ by $H_t^{\text{norm}}$ within each sample.
\end{list}
This within-sample partition avoids comparing entropy magnitudes across responses of different lengths or difficulty levels. Details in Appendix~\ref{app:entropy_based_partition}.

\paragraph{Weighted training objective.} We study token subsets with:
\begin{equation}
\mathcal{L}_{w}=\frac{\sum_{t=1}^{T} w_t \ell_t}{\sum_{t=1}^{T} w_t},
\end{equation}
where $w_t\ge 0$. This unifies hard masking and random sparse training; the
soft-reweighting intervention uses continuous advantage weights with all-token
normalization (Appendix~\ref{app:token_weighting_normalization}).

\paragraph{Training configurations.} We compare: (1) full-token training, (2) random-$20\%$ training, (3) anchor-only training (bottom $20\%$ entropy), and (4) explorer-only training (top $20\%$ entropy).

\paragraph{Control configurations.} We compare against controls based on
response position and next-token predictive entropy. Objective-magnitude,
token-type, and matched-random controls are included as additional axes for the
extended control analysis. Precise definitions are in
Appendix~\ref{app:control_configs}.

\section{Selective Training Reveals an Optimization Spectrum}
\label{sec:selective_training}

\subsection{Sparse baseline and entropy-selected departures}
\label{sec:sparse_context}

Uniformly random $20\%$ token training first establishes that the token-level RL
objective is sparsely estimable. In the selective-training diagnostic setup, it
reaches $42.60$ final mean@8 accuracy versus $43.55$ for matched full-token
DAPO, recovering $97.8\%$ of the full-token score; the last-five-checkpoint
average gives the same conclusion ($41.56$ versus $43.09$, or $96.4\%$).
Appendix~\ref{app:sparse_derivation} gives the sparse-gradient derivation, and
Appendix Figure~\ref{fig:sparse_estimability} shows the trajectory. These
numbers are used only within this diagnostic setup because the intervention
experiments in Table~\ref{tab:dynamic_soft_reweighting_main} use a different
dataset and step budget.

\subsection{Anchors are stable; explorers are fragile but informative}
\label{sec:anchor_training}
\label{sec:explorer_training}

\begin{figure}[t]
    \centering
    \begin{subfigure}[t]{0.32\linewidth}
        \centering
        \includegraphics[width=\linewidth]{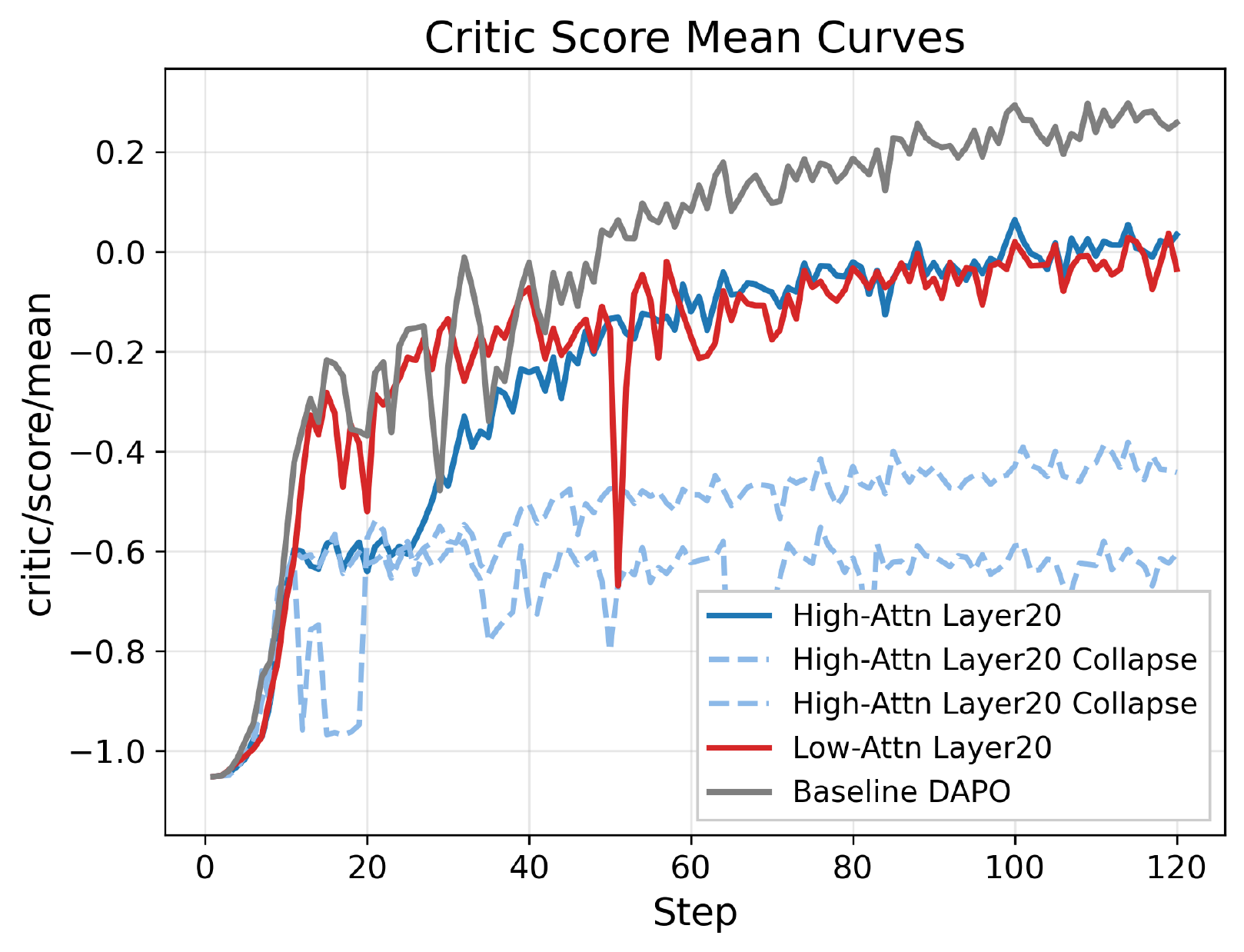}
        \caption{Training reward.}
        \label{fig:spectrum_training_reward}
    \end{subfigure}
    \hfill
    \begin{subfigure}[t]{0.32\linewidth}
        \centering
        \includegraphics[width=\linewidth]{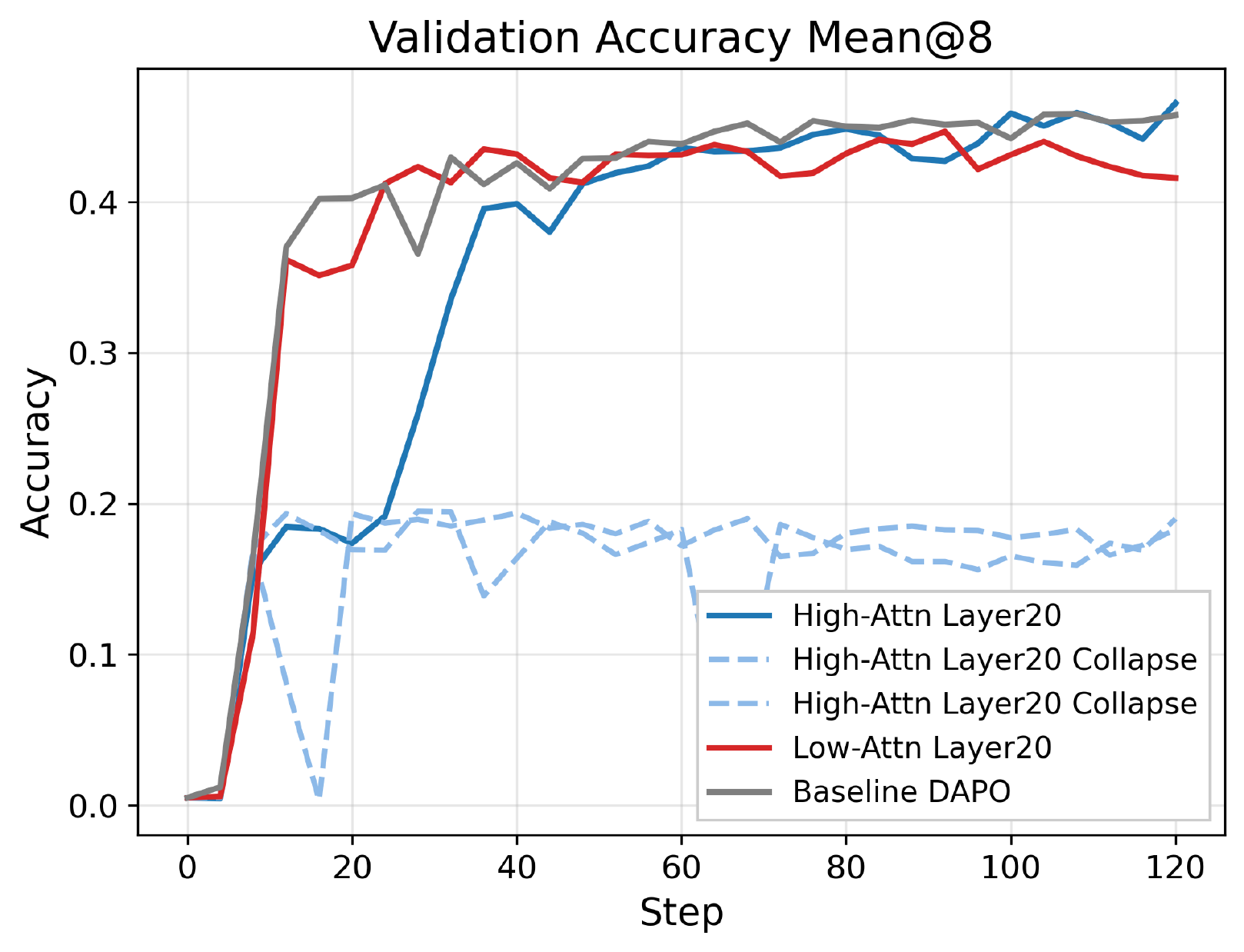}
        \caption{Held-out average.}
        \label{fig:spectrum_heldout}
    \end{subfigure}
    \hfill
    \begin{subfigure}[t]{0.32\linewidth}
        \centering
        \includegraphics[width=\linewidth]{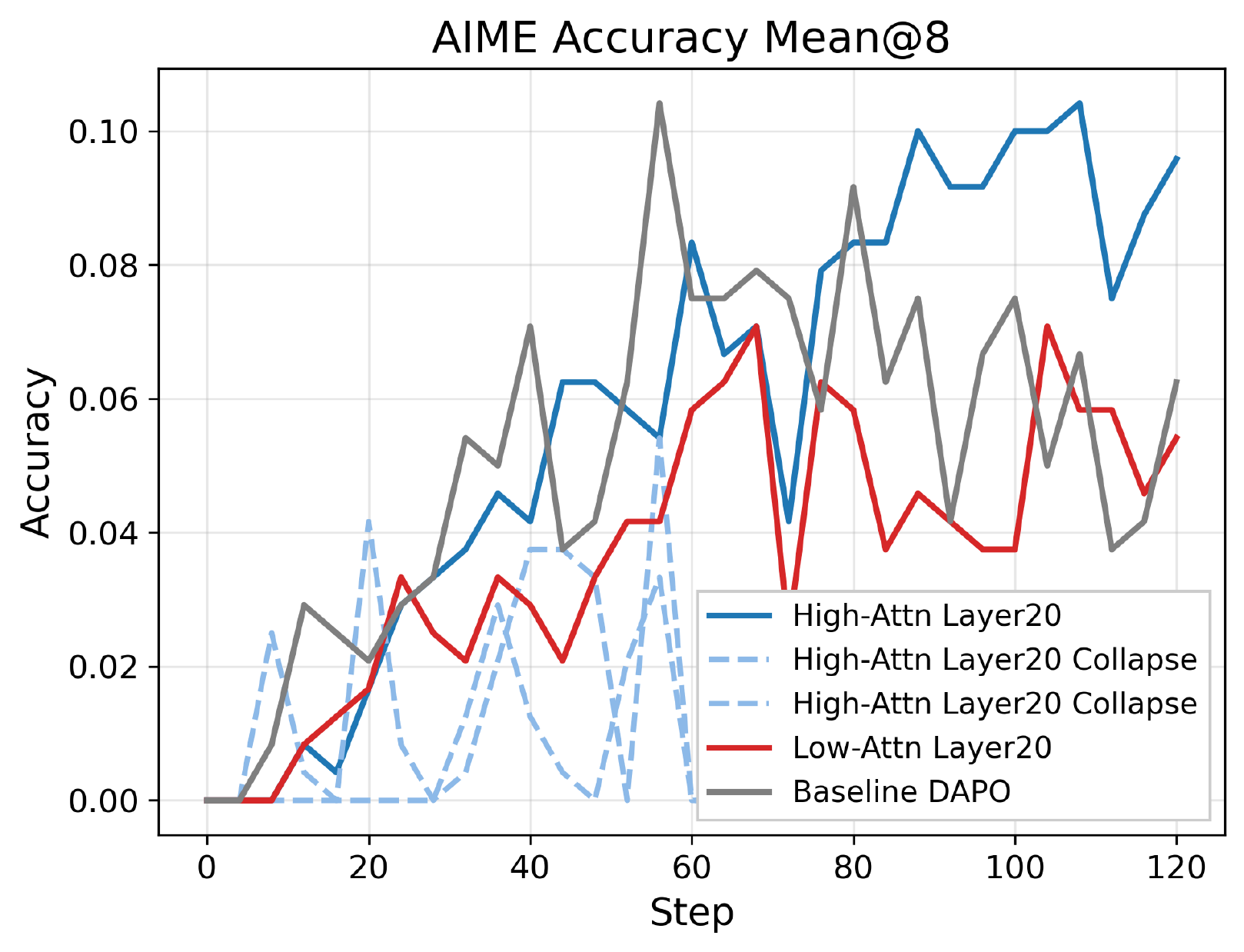}
        \caption{AIME.}
        \label{fig:spectrum_aime}
    \end{subfigure}
    \caption{
    Entropy-based selective training reveals an optimization spectrum. Anchors
    provide stable but ceiling-limited updates; explorers are fragile in
    isolation, while rare non-collapsed runs expose hard-benchmark signal.
    }
    \label{fig:selective_training_main}
\end{figure}

Figure~\ref{fig:selective_training_main} shows that entropy-defined subsets are
not interchangeable sparse estimators. Anchor-only training is stable and
competitive on most held-out benchmarks, with no reward or length collapse, but
it saturates earlier than full-token DAPO and lags most clearly on AIME. Anchors
therefore act as a reliable optimization backbone rather than a complete
substitute for all tokens.

Explorer-only training has the opposite profile. With the same $20\%$ token
budget, $5$ of $8$ independent runs collapse through short responses, length
instability, or reasoning degeneration
(Appendix~\ref{app:explorer_failure_modes}; Appendix
Table~\ref{tab:explorer_stats}). The $3$ non-collapsed runs are thus a
conditional diagnostic rather than a method comparison: they reach
$12.31 \pm 0.53$ on AIME, $6.48$ points above the matched full-token baseline
($5.83 \pm 0.89$), and average $46.13 \pm 1.58$ on the other held-out
benchmarks. The bimodality in Figure~\ref{fig:explorer_bimodal} indicates that
explorers are not merely noise, but their useful signal is trajectory-sensitive
and difficult to optimize without a stable backbone.

This asymmetry is the key empirical distinction. Anchors provide dependable
updates but do not cover all reasoning-relevant signal; explorers can contain
hard-benchmark signal, yet often destabilize optimization when used alone. We
therefore treat successful explorer-only seeds as evidence about signal quality
under stable trajectories, not as a competitive method. The intervention below
aims to retain the anchor backbone while making explorer-like signal usable
through softer allocation.

\subsection{Spectrum summary and controls}
\label{sec:spectrum_summary}

Random-$20\%$ training falls between anchors and explorers because it samples
across the spectrum without structural bias. Full-token training can therefore
be viewed as an implicit average over stable anchor-like updates and broader,
more volatile explorer-like updates. This also explains why hard masking is a
diagnostic tool rather than a recommended training rule: anchor-only training
loses coverage, while explorer-only training loses stability.

The pattern is not explained by simpler token attributes. Normalization changes
the severity but not the existence of explorer failure
(Appendix~\ref{app:normalization}). Position controls fail to reproduce the
entropy-defined behavior: front-only training largely fails, and back-only
training remains weaker than full-token DAPO. Prediction entropy is also
distinct from attention entropy, with only weak token-level correlation
($r=-0.1926$ over $97{,}995$ response tokens) and qualitatively different
off-diagonal token populations. Full control definitions and additional results
are in Appendix~\ref{app:control_configs}.

These controls rule out two shortcuts: entropy selection is not merely a
position mask, and attention entropy is not next-token uncertainty under another
name. A token can be uncertain but contextually selective, or confident but
supported by diffuse evidence.

This distinction is important for interpreting the spectrum. Predictive entropy
is about uncertainty over possible continuations, while attention entropy is
about how the generated token gathers contextual support. The two quantities can
therefore disagree on exactly the cases that matter for reasoning: a locally
uncertain operation may rely on a small set of premises, whereas a predictable
transition can still aggregate evidence from many positions.

\section{Mechanistic Evidence for the Optimization Spectrum}
\label{sec:evidence_gathering}

\paragraph{Support concentration.}
Attention entropy organizes tokens by support concentration rather than
locality. We measure how many positions are needed to accumulate a fixed
attention-mass threshold, along with spatial-span statistics. Anchors
concentrate mass on a small set of salient positions, nearby or distant;
explorers spread mass across many positions. The causal relevance of this
support structure is tested indirectly by selective training, controls, and
gradient geometry.

This evidence-gathering view is useful because it gives the entropy groups a
mechanistic interpretation beyond their scalar scores. Low-entropy tokens are
not simply ``easy'' tokens; they are tokens whose current generation can be
supported by a small effective set of context positions. High-entropy tokens are
not simply ``hard'' tokens either; they are tokens whose update aggregates a
broader support set and can therefore be more sensitive to which pieces of the
reasoning trace are emphasized. This distinction is why we treat attention
entropy as a support diagnostic rather than as a direct reward, difficulty, or
importance estimator.

\begin{figure*}[t]
    \centering
    \includegraphics[width=0.98\textwidth]{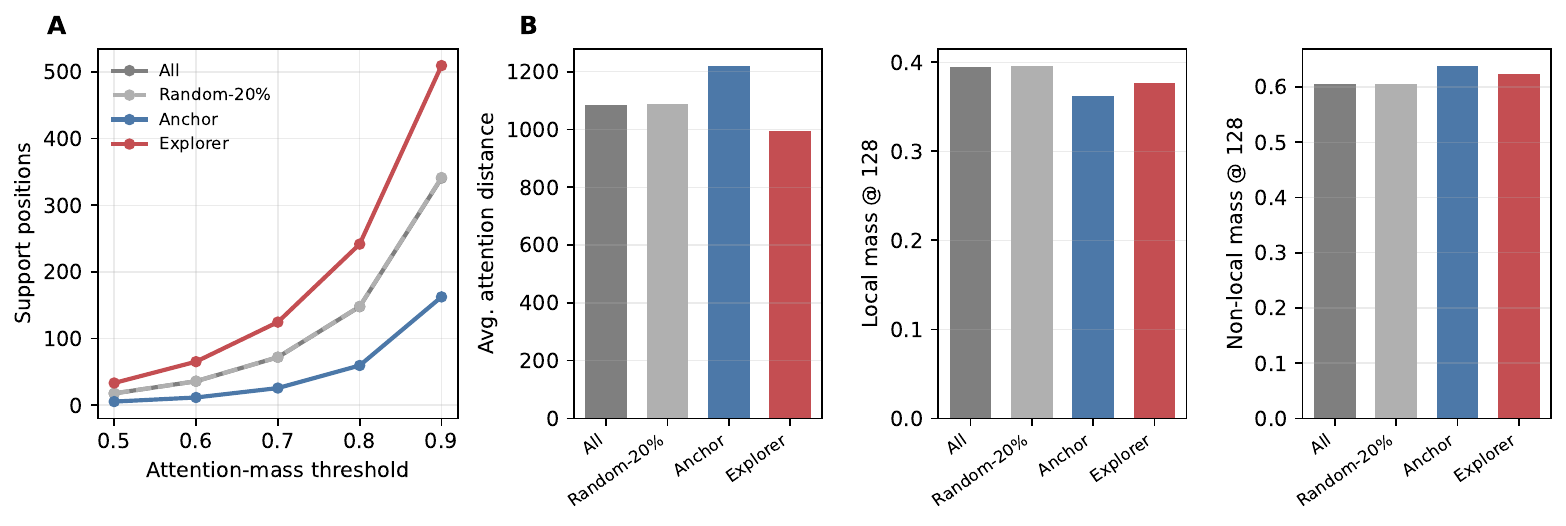}
    \caption{
    Evidence-gathering patterns under normalized attention entropy. Anchors
    require fewer support positions than explorers to accumulate the same
    attention mass; spatial-span statistics vary across entropy definitions, so
    support concentration is the stable interpretation.
    }
    \label{fig:evidence_gathering_concentration_distance}
\end{figure*}

Figure~\ref{fig:evidence_gathering_concentration_distance} shows a consistent
concentration separation: anchors rely on sparse selective support, whereas
explorers aggregate diffuse multi-position support. Random-$20\%$ tokens track
the all-token statistics, as expected. The concentration distinction persists
under alternative entropy definitions, while distance-based statistics are less
stable; additional variants and qualitative maps are in
Appendix~\ref{app:evidence_gathering}.

Thus the stable axis is sparse-versus-diffuse support, not local-versus-global
attention: under normalized entropy, anchors can even be slightly more
non-local because they selectively retrieve a few salient positions from a long
visible context.

\paragraph{Entropy dynamics.}
\label{sec:entropy_dynamics}
Because anchors and explorers are defined by within-response entropy
percentiles, their ordering is expected at each step; the relevant question is
whether RL collapses the separation. It does not. Across Qwen3-14B and
Qwen3-8B, Appendix Figure~\ref{fig:entropy_dynamics} shows explorers remaining
in a high-entropy support regime, anchors remaining in a low-entropy regime,
and full-token statistics between them even as means drift. Raw, top-$k$, and
fixed-position variants show the same qualitative separation
(Appendix~\ref{app:additional_entropy_dynamics}).

Together with selective-training results, these dynamics show that structurally
distinct token groups persist during RL and remain optimization-relevant.
Volatility is not inferred from entropy dispersion alone; it is established by
explorer collapse and the gradient instability below.

This persistence is important for using entropy as a training coordinate. If the
anchor/explorer distinction disappeared after a few updates, entropy-aware
allocation would only be a transient initialization heuristic. Instead, the
separation remains visible while the policy changes, so the current attention
pattern can provide a coarse but repeatedly available signal about which tokens
are supported by concentrated evidence and which require broader aggregation.

\paragraph{Gradient geometry.}
\label{sec:gradient_quality}
The same spectrum appears in update geometry. Let $g_{\mathrm{full}}$ be the
full-token gradient and $g_S$ the gradient from
$S\in\{\mathrm{anchor},\mathrm{explorer},\mathrm{rand}\}$. We compare:
\begin{align}
&\operatorname{NormRatio}(S)
=
\frac{\|g_S\|_2}{\|g_{\mathrm{full}}\|_2},
\label{eq:norm_ratio}\\
&\operatorname{Cosine}(S)
=
\frac{\langle g_S,g_{\mathrm{full}}\rangle}
{\|g_S\|_2\|g_{\mathrm{full}}\|_2},
\label{eq:cosine}\\
&\operatorname{ProjRatio}(S)
=
\frac{\langle g_S,g_{\mathrm{full}}\rangle}
{\|g_{\mathrm{full}}\|_2^2}.
\label{eq:proj_ratio}
\end{align}
Cosine measures directional agreement, norm ratio measures magnitude, and
projection ratio measures effective contribution along the full-token update.

\begin{figure}[t]
    \centering
    \begin{minipage}{0.32\linewidth}
        \centering
        \includegraphics[width=\linewidth]{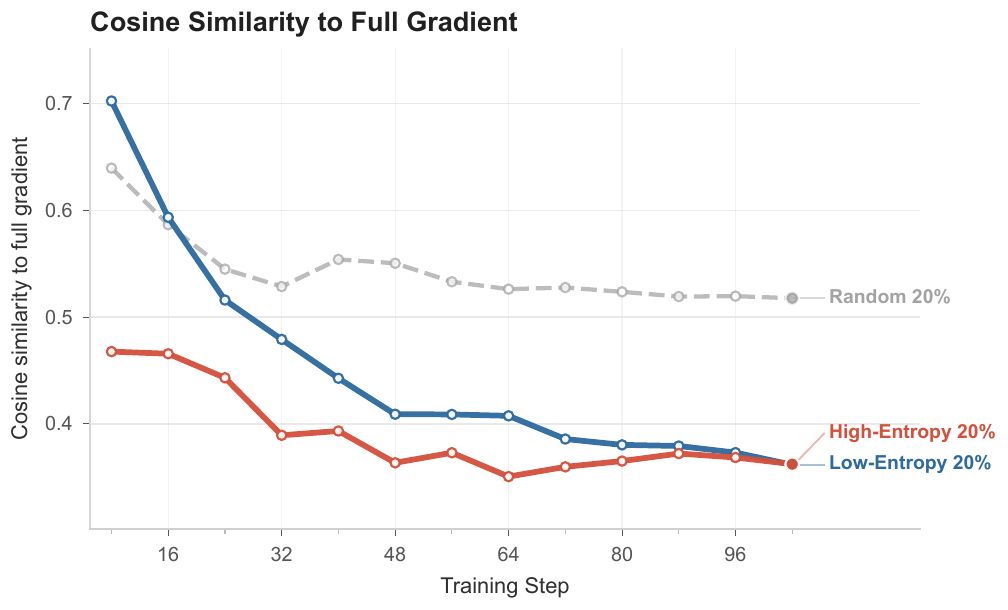}
        \centerline{\small (a) Cosine}
    \end{minipage}
    \hfill
    \begin{minipage}{0.32\linewidth}
        \centering
        \includegraphics[width=\linewidth]{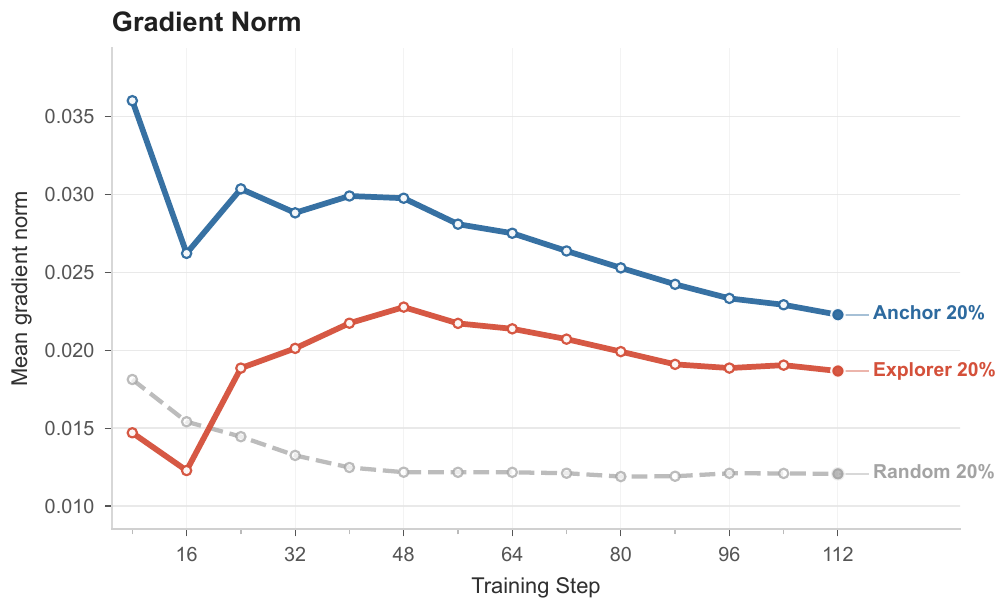}
        \centerline{\small (b) Norm}
    \end{minipage}
    \hfill
    \begin{minipage}{0.32\linewidth}
        \centering
        \includegraphics[width=\linewidth]{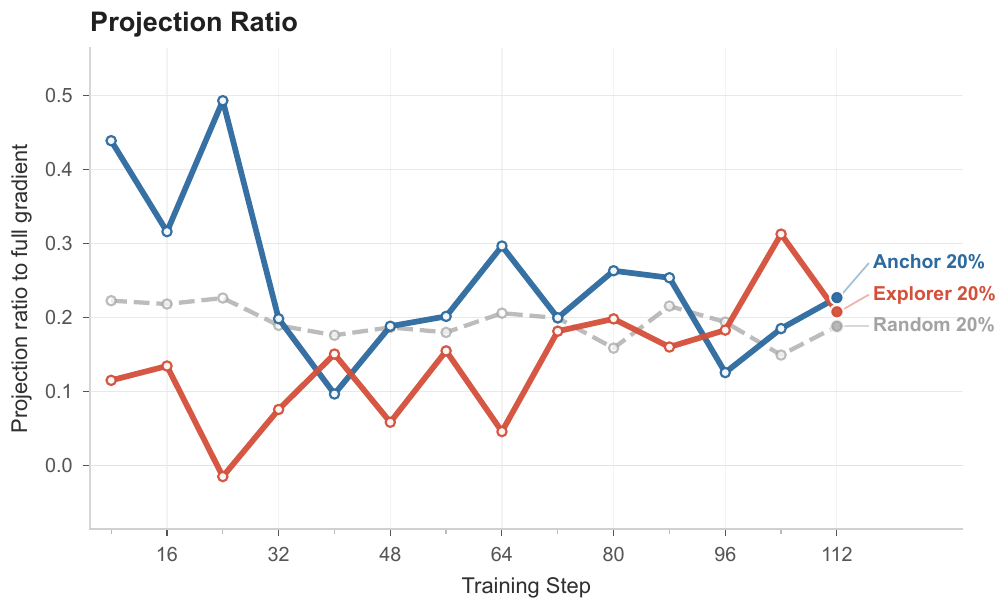}
        \centerline{\small (c) Projection}
    \end{minipage}
    \caption{
    Gradient diagnostics against the full-token update. Random subsets are most
    aligned, anchors are aligned early but drift, and explorers have substantial
    norm but volatile direction and projection.
    }
    \label{fig:gradient_probe}
\end{figure}

Figure~\ref{fig:gradient_probe} shows that the distinction is not simply
gradient magnitude. Random subsets remain the closest sparse approximation to
the full-token gradient. Anchor gradients are strongly aligned early (around
$0.7$ cosine similarity) but decline, matching stable training with a lower
ceiling. Explorer gradients can be large, yet pass through a low-alignment
middle phase and remain volatile in projection. Decile-level probes in
Appendix~\ref{app:decile_probe} show that useful update mass shifts across the
entropy spectrum, motivating soft reweighting rather than fixed hard masking.

This reconciles anchor stability with explorer potential: high-entropy tokens
can produce strong updates, but low cosine and volatile projection mean those
updates are not consistently aligned with the full-token trajectory. Useful
signal is distributed across entropy bands and changes over training, arguing
against a fixed hard partition as the final training rule.

The gradient evidence also explains why random sparse subsets are a strong
baseline. Random tokens approximate the full-token update because they sample
the whole spectrum, but they do not expose which parts of the spectrum are
stable, ceiling-limited, or volatile. Entropy-structured probes sacrifice some
sparse-estimation optimality in order to reveal those roles.

\section{Validation Intervention: Dynamic Entropy-Aware Soft Reweighting}
\label{sec:soft_weighting}

Hard masks reveal the spectrum but are too extreme as a training rule:
anchor-only training loses coverage and explorer-only training loses stability.
We therefore use dynamic entropy-aware soft reweighting as a validation
intervention. Each generated token receives a continuous advantage weight from
normalized decision attention entropy, so all valid response tokens remain in
the objective. The \textbf{Low2High} schedule emphasizes low-entropy tokens
early and shifts weight toward high-entropy tokens during warmup, testing
whether explorer-like signal becomes useful after a stable backbone forms.
Details and controls are in Appendix~\ref{app:dynamic_soft_reweighting}.

The intervention is deliberately simple. It does not introduce a new reward
model, discard middle-spectrum tokens, or tune a benchmark-specific schedule.
Its purpose is narrower: if attention entropy identifies actionable structure,
then smoothly changing token weights along that spectrum should improve the
matched training setup while preserving full-token coverage.

Low2High is also designed to match the empirical asymmetry rather than to
privilege one endpoint permanently. Early low-entropy emphasis follows the
selective-training evidence that anchors provide reliable directional support.
Later high-entropy emphasis tests whether explorer-like updates become useful
after the trajectory has accumulated enough stable structure. The reverse
High2Low schedule is therefore a meaningful control: it uses the same endpoints
and the same continuous weighting machinery, but exposes the volatile side of
the spectrum before the anchor-like backbone has formed.

\subsection{Empirical effect on Qwen3-8B-Base}
\label{sec:soft_weighting_empirical}

Table~\ref{tab:dynamic_soft_reweighting_main} evaluates Qwen3-8B-Base under the
matched VeRL-DAPO setup and strict boxed-answer exact-match protocol. Explorer
successful seeds are shown only as the conditional diagnostic from
Section~\ref{sec:selective_training}; additional seed, layer-source, schedule,
and model checks are in Appendix~\ref{app:soft_reweighting_multiseed}--
\ref{app:soft_reweighting_additional_models}.

\begin{table}[t]
\centering
\footnotesize
\caption{Qwen3-8B-Base results under the matched intervention setup. ``Avg.''
is the arithmetic mean over AIME, OlympiadBench, Minerva, and MATH. Reported
$\pm$ values summarize repeated runs under the same protocol; explorer
successful seeds are a conditional diagnostic; the high-prediction-entropy row
is a predictive-entropy control.}
\label{tab:dynamic_soft_reweighting_main}
\resizebox{\textwidth}{!}{
\begin{tabular}{lccccc}
\toprule
Method
& AIME
& OlympiadBench
& Minerva
& MATH
& Avg. \\
\midrule
Full-token DAPO
& $5.83{\pm}0.89$
& $34.24{\pm}1.07$
& $25.69{\pm}1.41$
& $71.80{\pm}1.31$
& $34.39{\pm}0.47$ \\

Random-$20\%$
& $8.12{\pm}0.96$
& $32.83{\pm}1.33$
& $22.94{\pm}1.52$
& $70.91{\pm}1.13$
& $33.70{\pm}0.71$ \\

High Prediction Entropy-only
& $1.84 \pm 1.34$
& $33.67 \pm 0.46$
& $27.47 \pm 1.52$
& $\textbf{74.41} \pm 0.98$
& $34.35 \pm 1.14$ \\

Anchor-only
& $5.60 \pm 0.77$
& $30.47 \pm 0.96$
& $24.39 \pm 0.88$
& $66.43 \pm 1.04$
& $31.72 \pm 0.43$ \\



Explorer-only, all seeds
& $4.62 \pm 5.96$
& $20.18 \pm 13.18$
& $16.82 \pm 4.60$
& $46.79 \pm 18.80$
& $22.10 \pm 10.64$ \\

Explorer-only, successful seeds (diagnostic)
& $\textbf{12.31} \pm 0.43$
& $37.18 \pm 0.99$
& $22.73 \pm 0.66$
& $70.98 \pm 1.21$
& $35.80 \pm 0.82$ \\

\rowcolor{rowblue}
Entropy reweighting, shallow layer (Layer 8)
& $7.37{\pm}1.18$
& $36.95{\pm}1.10$
& $\mathbf{31.41}{\pm}1.39$
& $73.88{\pm}1.49$
& $\mathbf{37.40}{\pm}0.81$ \\

\rowcolor{rowblue}
Entropy reweighting, mid layer (Layer 20)
& $9.75{\pm}1.21$
& $36.48{\pm}1.17$
& $28.22{\pm}1.34$
& $74.15{\pm}1.52$
& $37.15{\pm}0.77$ \\

\rowcolor{rowblue}
Entropy reweighting, deep layer (Layer 31)
& $9.14{\pm}1.19$
& $\mathbf{37.86}{\pm}1.24$
& $28.86{\pm}1.28$
& $73.54{\pm}1.44$
& $37.35{\pm}0.78$ \\
\bottomrule
\end{tabular}
}
\end{table}

Across entropy-source settings, entropy-aware reweighting improves over
full-token DAPO on the held-out average and on every benchmark, but the layer
source changes the gain profile. Shallow reaches the strongest average
($37.40$) and Minerva score; mid-layer is the original controlled setting and
gives the strongest AIME and MATH among entropy rows; deep gives the strongest
OlympiadBench. This supports actionable signal heterogeneity: different layers
expose useful but non-identical allocation signals. The high-prediction-entropy
control is a sharp contrast: it nearly matches full-token DAPO on average
($34.35$ versus $34.39$) and is strongest on MATH ($74.41$), but drops AIME to
$1.84$, so next-token uncertainty does not recover the hard-reasoning allocation
exposed by attention-support entropy.

We do not interpret the layer comparison as a full layer sweep or as evidence
that one depth should be tuned as a benchmark knob. Rather, the shallow, mid, and
deep rows test whether the diagnostic survives moving away from the original
Layer-20 controlled setting. The answer is positive, but not uniform: a
shallower source favors Minerva and the aggregate average, the mid-layer source
keeps the strongest AIME and MATH results among entropy rows, and the final-layer
source favors OlympiadBench. This pattern is consistent with different depths
mixing local computation, evidence integration, and output-facing structure in
different proportions. The shared result is that all three entropy-source
settings improve the held-out average over full-token DAPO while exposing
different allocation profiles.

\subsection{Robustness and schedule ablations}
\label{sec:soft_weighting_robustness}

The seed-robustness and schedule ablations use the original fixed Layer-20
entropy source. Over three seeds, full-token DAPO obtains $34.39 \pm 0.47$
average accuracy, while Layer-20 entropy-aware reweighting obtains
$37.15 \pm 0.77$ with benchmark-wise gains. The shallow- and deep-layer rows in
Table~\ref{tab:dynamic_soft_reweighting_main} test entropy-source sensitivity.

\begin{table}[t]
\centering
\small
\caption{
Schedule ablation for entropy-aware soft reweighting on Qwen3-8B-Base using the
fixed Layer-20 entropy source. Static endpoint biases are weaker than dynamic
temporal allocation. Low2High is the main Layer-20 schedule; High2Low is a
reverse-order control.
}
\label{tab:main_schedule_ablation}
\setlength{\tabcolsep}{3.5pt}
\resizebox{\linewidth}{!}{
\begin{tabular}{@{}lccrrrrrr@{}}
\toprule
Variant
& Early
& Late
& AIME
& OlympiadBench
& Minerva
& MATH
& Avg.
& Avg. w/o AIME \\
\midrule
Full-token DAPO
& uniform
& uniform
& 5.83
& 34.24
& 25.69
& 71.80
& 34.39
& 43.91 \\
Static anchor-biased
& low
& low
& 5.78
& 34.18
& 25.80
& 71.23
& 34.25
& 43.74 \\
Static explorer-biased
& high
& high
& 5.59
& 33.97
& 25.88
& 77.60
& 35.76
& 45.82 \\
\rowcolor{rowblue}
Dynamic Low2High
& low
& high
& \textbf{9.75}
& 36.48
& 28.22
& 74.15
& \textbf{37.15}
& 46.28 \\
\rowcolor{gray!15}
Dynamic High2Low
& high
& low
& 3.78
& \textbf{38.84}
& \textbf{29.58}
& 75.18
& 36.85
& \textbf{47.87} \\
\bottomrule
\end{tabular}
}
\end{table}

Table~\ref{tab:main_schedule_ablation} adds two qualifications. Fixed endpoint
biases miss either coverage or stability: anchor bias is stable but does not
improve the average, while explorer bias mainly helps MATH. Dynamic schedules
are stronger, but order matters. High2Low is strong without AIME, whereas
Low2High is much stronger on AIME and is the best Layer-20 schedule, consistent
with stabilizing early before increasing exposure to volatile explorer-like
signal.

This also clarifies the role of the hard masks used earlier. Anchor-only and
explorer-only training are intentionally extreme probes: they separate stability
from coverage by removing most tokens from the objective. The soft intervention
keeps all valid response tokens and changes only their relative influence. It
therefore tests whether attention entropy is actionable without turning the
analysis into a brittle token-selection rule.

\begin{figure}[t]
    \centering
    \begin{minipage}{0.48\linewidth}
        \centering
        \includegraphics[width=\linewidth]{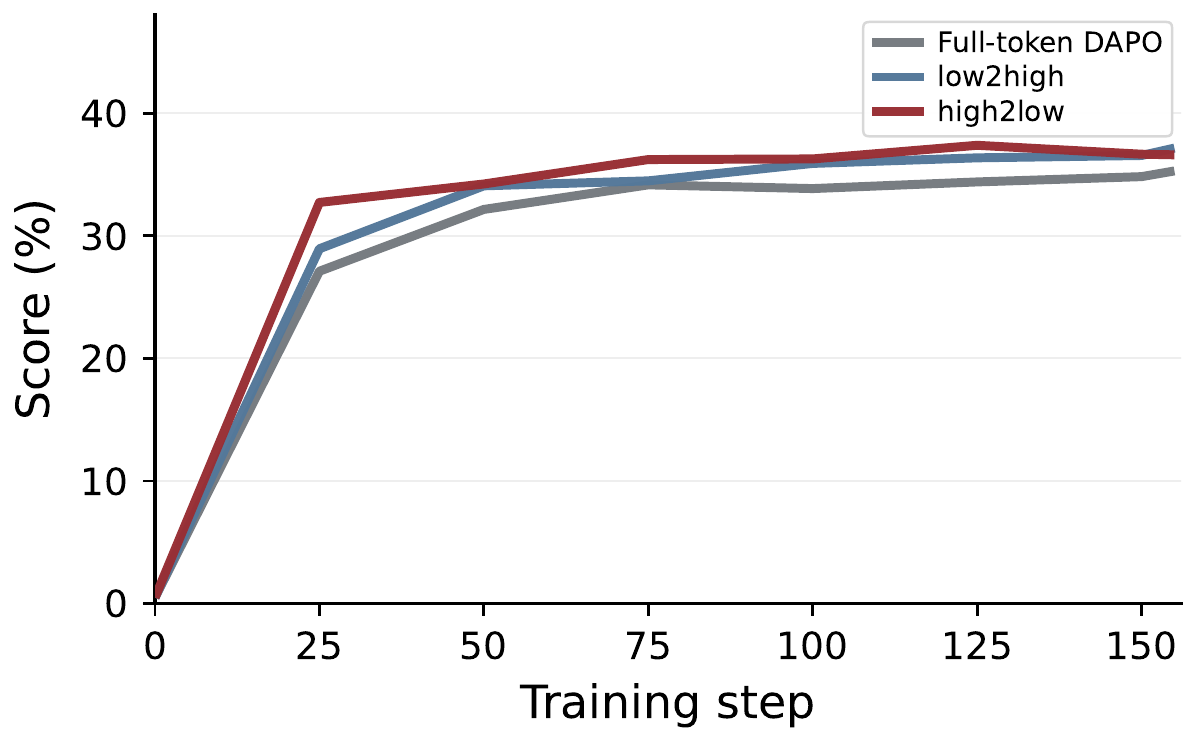}
        \vspace{-0.5em}
        \centerline{\small (a) Held-out average}
    \end{minipage}
    \hfill
    \begin{minipage}{0.48\linewidth}
        \centering
        \includegraphics[width=\linewidth]{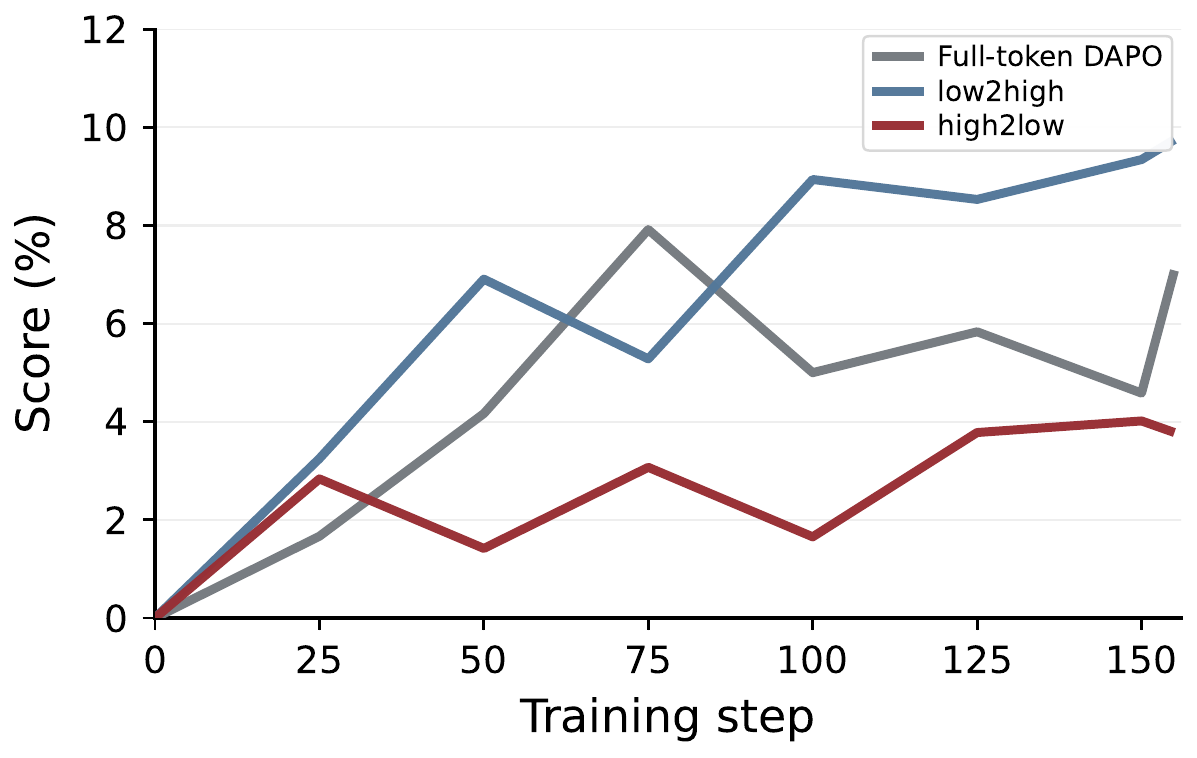}
        \vspace{-0.5em}
        \centerline{\small (b) AIME}
    \end{minipage}
    \caption{
    Training curves for the main Low2High schedule and reverse High2Low control.
    Both dynamic schedules improve the aggregate average, but Low2High gives a
    stronger late-stage AIME trajectory, consistent with stabilizing early
    training before increasing high-entropy emphasis.
    }
    \label{fig:main_schedule_curves}
\end{figure}

Figure~\ref{fig:main_schedule_curves} gives the trajectory view: High2Low rises
quickly on average but remains weak on AIME, while Low2High behaves as a
conservative allocation rule that begins with stable anchor-like updates and
then increases explorer-like coverage. Additional transfer checks on Qwen3-14B
and Qwen2.5-7B improve held-out average and AIME (Appendix
\ref{app:soft_reweighting_additional_models}), suggesting that the effect is
not isolated to Qwen3-8B-Base.

\section{Related Work}
\label{sec:related_work}

\paragraph{RLVR and token-level credit.}
RLVR methods such as GRPO, DeepSeek-R1, and DAPO improve reasoning with
verifiable outcome rewards~\citep{shao2024deepseekmath,guo2025deepseekr1,yu2025dapo},
but mostly optimize the response as a unit. Process-supervision and token-level
reward work adds denser feedback~\citep{lightman2023let,wang2023mathshepherd,cui2025prime,zhong2024rto}.
Our work is complementary: we analyze the native token-level objective induced
by outcome-supervised RLVR.
This distinction matters because no extra verifier, process label, or reward
model is introduced. The same rollout-level reward is still used, but its
token-level update terms are organized by an internal support diagnostic. The
resulting question is therefore not whether additional supervision can identify
better reasoning steps, but whether the policy's own attention structure reveals
which parts of an outcome-supervised update are stable, volatile, or
coverage-limited.

\paragraph{Entropy and attention support.}
Recent RLVR analyses connect non-uniform token updates to next-token policy
entropy or sparse critical positions~\citep{cui2025entropymechanism,wang2025beyond8020,he2026polarityentropy,meng2026sparsecritical}.
We instead study attention entropy, which measures contextual-support
concentration rather than uncertainty over possible next tokens. Attention
entropy has also been used to diagnose context utilization and information
seeking~\citep{zhang2025attentionentropy,su2024dragin}; we connect it to
selective-training behavior, entropy dynamics, gradient alignment, and a
validating reweighting intervention. Appendix~\ref{app:detailed_related_work}
gives a fuller discussion.

Our layer-source analysis also differs from standard layer probing. We do not use
shallow, middle, and final layers to claim a universal best depth, nor do we
optimize the layer choice per benchmark. Instead, the comparison is a sensitivity
check on whether attention entropy remains useful when computed from different
representational depths. The observed shifts across AIME, OlympiadBench, Minerva,
and MATH are useful because they expose different allocation profiles without
changing the reward, data, optimizer, or evaluation protocol. This makes the
main intervention closer to a diagnostic stress test than to a new architecture
or benchmark-specific training recipe.

A second difference is the validation criterion. Token-importance analyses can
remain correlational if they only rank positions or correlate scores with final
answers. Our protocol first uses hard selective training to expose asymmetric
failure modes, then checks support concentration and gradient alignment, and
finally asks whether a soft reweighting rule improves the matched RLVR setup.
This sequence is deliberately conservative: the diagnostic must explain behavior,
survive controls, and support an intervention that keeps full response coverage.
It also prevents the empirical claim from resting on successful explorer-only
seeds, which remain conditional diagnostics rather than method comparisons. This
matters for RLVR because optimization can fail through response-length collapse
or unstable reasoning even when some high-entropy tokens contain useful benchmark
signal.

\section{Conclusion}

We introduced attention entropy as a token-level diagnostic for RL reasoning and
showed that it reveals an optimization spectrum. Low-entropy anchors provide
stable but ceiling-limited updates, whereas high-entropy explorers use diffuse
support and can carry complementary but volatile signal. Evidence-gathering
statistics, entropy dynamics, gradient geometry, and position/prediction-entropy
controls support this heterogeneous-signal view.

The main implication is that useful RL signal is distributed but not
homogeneous. Hard masks expose the roles of different token groups, while dynamic
entropy-aware soft reweighting shows that the diagnostic can improve token
allocation without removing tokens from the objective. Attention entropy should
therefore be read as a support-structure coordinate, not as a standalone token
importance score.

\paragraph{Limitations.}
\label{para:limitation}
The results are a controlled case study on Qwen models with VeRL-DAPO, not a
claim of universal transfer. We use one fixed mid-layer probe for the controlled
mechanistic analyses, and the layer-source comparison checks only representative
shallow, middle, and final layers. The schedule ablation also leaves open
whether the allocation rule should adapt online to training dynamics.
Our evidence is also at the level of entropy-defined token groups and allocation
rules, rather than a head-by-head causal localization of attention behavior.
Future work should separate which depth best exposes token-level support, when
training should move from anchor-like stability to explorer-like coverage, and
whether allocation can react to collapse indicators such as response length,
reward stagnation, or gradient misalignment.

\clearpage

\bibliographystyle{plainnat}
\bibliography{main}

\clearpage

\appendix

\section{Detailed Related Work}
\label{app:detailed_related_work}

\paragraph{RLVR for reasoning language models.}
Reinforcement learning with verifiable rewards (RLVR) has recently become a central approach for improving the reasoning ability of large language models. DeepSeekMath introduced Group Relative Policy Optimization (GRPO), which removes the value model in PPO-style training and estimates advantages from groups of sampled responses~\citep{shao2024deepseekmath}. DeepSeek-R1 further demonstrated that large-scale RL can elicit long-chain reasoning behaviors from base models using verifiable outcome rewards~\citep{guo2025deepseekr1}. Subsequent systems such as DAPO improved the reproducibility and scalability of RL reasoning training through practical algorithmic choices including decoupled clipping and dynamic sampling~\citep{yu2025dapo}. While these works establish RLVR as an effective training paradigm, they mainly operate at the sequence or response level. In contrast, our work studies the token-level structure of RLVR updates: rather than asking whether RLVR improves reasoning, we ask which response tokens provide stable or unstable learning signals during RL training.

\paragraph{Process rewards and token-level credit assignment.}
A major challenge in RLVR is that sparse outcome rewards must be assigned to many intermediate reasoning tokens. Prior work addresses this issue by introducing denser supervision. Process supervision and process reward models provide feedback for intermediate reasoning steps and can improve mathematical reasoning reliability~\citep{lightman2023let,wang2023mathshepherd}. PRIME further explores implicit process rewards, deriving dense token-level reward signals from outcome labels and policy rollouts without requiring explicit human process annotations~\citep{cui2025prime}. Other work formulates alignment and preference optimization from a token-level perspective, learning token-wise reward or value signals for more fine-grained policy optimization~\citep{zhong2024rto}. These methods aim to construct better token-level rewards. Our work is complementary: we do not introduce an external process reward model, but instead analyze the native token-level objective already used in RLVR and show that its learning signals are structurally heterogeneous.

\paragraph{Entropy-aware analyses of RLVR.}
Several recent works analyze RLVR through the lens of policy entropy. The Entropy Mechanism of Reinforcement Learning for Reasoning Language Models studies entropy collapse during RL reasoning training and argues that the exhaustion of policy entropy limits continued exploration~\citep{cui2025entropymechanism}. Beyond the 80/20 Rule shows that a small fraction of high-policy-entropy tokens act as reasoning forks and can dominate the effectiveness of RLVR updates~\citep{wang2025beyond8020}. Related work further connects token entropy to credit assignment, arguing that high-entropy positions carry more potential credit under outcome-supervised RLVR~\citep{he2026polarityentropy}, and shows that RLVR induces sparse but critical token-level distributional shifts~\citep{meng2026sparsecritical}. These studies reveal that RLVR updates are not uniformly distributed across tokens. However, they primarily define entropy over the next-token output distribution, measuring uncertainty over possible token choices. Our work studies a different internal signal: attention entropy. Instead of measuring how uncertain the model is about the next token, attention entropy measures how concentrated or diffuse a token's contextual support is. This distinction allows us to identify anchor tokens with sparse selective support and explorer tokens with diffuse multi-position aggregation.

\paragraph{Attention entropy and internal support structure.}
Attention entropy has also been used as a diagnostic for how language models use context. Prior studies show that abnormal attention entropy can reveal failures in context encoding and that attention-based signals can expose a model's information needs during generation~\citep{zhang2025attentionentropy,su2024dragin}. These works suggest that attention entropy is a useful internal indicator of context utilization. We extend this perspective to RL reasoning training by connecting attention support structure with selective training behavior, entropy dynamics, and gradient geometry in RLVR.

\section{Experimental Details and Hyperparameters}
\label{app:experimental_details}

This appendix provides the implementation details for the RLVR training
experiments. The paper contains two matched experiment families: an earlier
selective-training diagnostic setup and the full-data soft-reweighting
intervention setup. Within each family, compared variants use the same base
model, reward function, rollout configuration, data split, and step budget; the
main intended difference is the token-weighting rule applied to the
response-token loss. Across families, however, the data and step budgets differ,
so absolute scores should not be compared across tables.

\subsection{Experiment-family summary}
\label{app:experiment_summary}

Table~\ref{tab:experiment_family_summary} summarizes the experiment families
used in the paper. The selective-training experiments are validation-oriented
diagnostics: to make the initial comparisons faster, they use $40\%$ of the
full training set, a different training-step budget, and validation subsets
sampled from the full benchmark test sets. These validation subsets can be
biased relative to the full test sets. We therefore interpret the selective
training results only through comparisons against their matched full-token
baseline, not as absolute scores comparable to the full-data intervention table.

\begin{table}[h]
\centering
\small
\caption{Summary of experiment families and comparability.}
\label{tab:experiment_family_summary}
\setlength{\tabcolsep}{3pt}
\resizebox{\linewidth}{!}{
\begin{tabular}{@{}p{0.18\linewidth}p{0.20\linewidth}p{0.22\linewidth}p{0.18\linewidth}p{0.17\linewidth}@{}}
\toprule
Experiment family & Purpose & Training data and steps & Evaluation data & Intended comparison \\
\midrule
Selective-training diagnostic
& Validate sparse estimability and expose anchor/explorer behavior
& $40\%$ of the full training set; diagnostic step budget different from the full-data intervention setup
& Subsets sampled from the full AIME, OlympiadBench, Minerva, and MATH test sets; potentially biased relative to the full test sets
& Compare full-token, random-$20\%$, anchor-only, and explorer-only runs within this matched diagnostic setup only \\
\midrule
Soft-reweighting intervention
& Test whether attention-entropy structure can inform token allocation
& Full training set and the intervention step budget used for Table~\ref{tab:dynamic_soft_reweighting_main}
& Full benchmark evaluation used in Table~\ref{tab:dynamic_soft_reweighting_main}
& Compare the intervention against baselines in the same table; do not compare absolute scores to the diagnostic setup \\
\midrule
Additional-model transfer checks
& Supporting evidence for the soft-reweighting intervention
& Same protocol family as the full-data intervention setup, applied to additional Qwen model settings
& Same evaluation protocol as the intervention setup
& Supporting transfer checks, not a universal effectiveness claim \\
\bottomrule
\end{tabular}
}
\end{table}

\subsection{Model, framework, and systems setup}

All main experiments use the official Qwen3-8B-Base checkpoint with the original
Qwen tokenizer. We train the model using VeRL with the original DAPO recipe.
Training is performed in bf16 precision with FSDP. We enable gradient
checkpointing, remove padding during model execution, and use dynamic batch
sizing for actor, reference, and rollout log-probability computation. Unless
otherwise specified, filter-group sampling is disabled to make selective-token
training directly comparable with the full-token baseline.

\begin{table}[h]
\centering
\small
\caption{Model and systems configuration.}
\label{tab:systems_config}
\begin{tabular}{@{}p{0.34\linewidth}p{0.58\linewidth}@{}}
\toprule
Item & Setting \\
\midrule
Base model & Official Qwen3-8B-Base checkpoint \\
Tokenizer & Original Qwen tokenizer \\
RL framework & VeRL with the original DAPO recipe \\
Precision & bf16 \\
Distributed training & FSDP \\
Gradient checkpointing & Enabled \\
Remove padding & Enabled \\
Dynamic batch size & Enabled for actor, reference, and rollout log-probability computation \\
Parameter offload & Enabled \\
Optimizer offload & Enabled \\
Tensor model parallel size & 1 \\
Sequence parallel size & 1 \\
GPUs & 2 nodes for selective-training probes; 4 nodes for full-data intervention runs; 8 NVIDIA H800 80GB GPUs per node \\
Filter groups & Disabled \\
\bottomrule
\end{tabular}
\end{table}

\subsection{Training and validation data}

For the full-data intervention setup, we train on verifiable
mathematical-reasoning prompts from DeepScaler(\cite{luo2025deepscaler}) and MATH(\cite{hendrycks2021measuring}) data. In our
implementation, these correspond to the DeepScaler training split and the
SimpleRL/Qwen training split. The selective-training diagnostic setup uses
$40\%$ of this full training pool and evaluates on subsets sampled from the full
benchmark test sets, as summarized in
Table~\ref{tab:experiment_family_summary}. We do not apply additional data
filtering. Prompts are read from the \texttt{prompt} field, and prompts
exceeding the maximum prompt length are left-truncated.


\begin{table}[h]
\centering
\small
\caption{Full-data intervention configuration.}
\label{tab:data_config}
\begin{tabular}{ll}
\toprule
Item & Setting \\
\midrule
Training data & DeepScaler train split + MATH train split \\

Validation data &
\begin{tabular}[t]{@{}l@{}}
AIME + \texttt{OLYMPIAD\_BENCH}~\cite{he2024olympiadbench} \\
+ MINERVA~\cite{lewkowycz2022solving} \\
+ MATH test split
\end{tabular} \\

Prompt field & \texttt{prompt} \\
Data filtering & None \\
Prompt truncation & Left truncation \\
Maximum prompt length & 1024 \\
Maximum response length & 8192 \\
Maximum sequence length & 9216 \\
\bottomrule
\end{tabular}
\end{table}

\subsection{Reward function and overlong penalty}

We use the DAPO reward manager with a rule-based verifier. For each generated
response, we extract the content inside \texttt{\textbackslash boxed\{\}} and
compare it with the reference answer using string matching. The reward takes
three values: $-2$ if no boxed answer is found, $-1$ if a boxed answer is found
but does not match the reference answer, and $1$ if the boxed answer is correct.
All runs use the same overlong-response penalty.

\begin{table}[h]
\centering
\small
\caption{Reward configuration.}
\label{tab:reward_config}
\begin{tabular}{ll}
\toprule
Item & Setting \\
\midrule
Reward manager & DAPO \\
Answer extraction & Content inside \texttt{\textbackslash boxed\{\}} \\
Verifier & String matching against the reference answer \\
Reward for missing boxed answer & $-2$ \\
Reward for incorrect boxed answer & $-1$ \\
Reward for correct boxed answer & $1$ \\
Overlong buffer & Enabled \\
Overlong buffer length & 2048 \\
Overlong penalty factor & 1.0 \\
\bottomrule
\end{tabular}
\end{table}

\subsection{RL optimization hyperparameters}

We use the GRPO-style advantage estimator in DAPO. We disable KL penalties both
in the reward and in the actor loss in the main experiments, so the reference
model is used for log-probability computation but not for an explicit KL penalty.
The actor entropy coefficient is set to zero. All selective-training variants
use the same optimizer, clipping, and batch-size settings.

\begin{table}[h]
\centering
\small
\caption{RL optimization hyperparameters.}
\label{tab:rl_hparams}
\begin{tabular}{ll}
\toprule
Item & Setting \\
\midrule
Advantage estimator & GRPO \\
Use KL in reward & False \\
KL coefficient in reward & 0.0 \\
Use actor KL loss & False \\
Actor KL loss coefficient & 0.0 \\
Clip ratio low & 0.2 \\
Clip ratio high & 0.28 \\
Clip ratio constant $c$ & 10.0 \\
Actor entropy coefficient & 0.0 \\
Optimizer & AdamW \\
Learning rate & $1\times 10^{-6}$ \\
Warmup steps & 10 \\
Weight decay & 0.1 \\
Gradient clipping & 1.0 \\
Loss aggregation mode & Token mean \\
Training prompt batch size & 512 \\
Generation prompt batch size & 1536 \\
PPO mini-batch size & 32 \\
PPO micro-batch size & Dynamic / unset \\
Total epochs & 10 \\
Validation before training & True \\
Validation frequency & Every 4 training steps \\
\bottomrule
\end{tabular}
\end{table}

\subsection{Rollout and evaluation configuration}

During training, each prompt is expanded into multiple sampled responses. The
rollout temperature is set to $1.0$, with top-$p=0.95$ and no top-$k$ truncation.
For validation, we also use temperature $1.0$ but set top-$p=0.7$ to reduce
evaluation variance. We sample 8 responses per validation prompt and report
mean@8 unless otherwise specified.

\begin{table}[h]
\centering
\small
\caption{Rollout and evaluation configuration.}
\label{tab:rollout_eval_config}
\begin{tabular}{ll}
\toprule
Item & Setting \\
\midrule
Training responses per prompt & 8 \\
Training temperature & 1.0 \\
Training top-$p$ & 0.95 \\
Training top-$k$ & $-1$ \\
Validation responses per prompt & 8 \\
Validation temperature & 1.0 \\
Validation top-$p$ & 0.7 \\
Validation top-$k$ & $-1$ \\
Validation sampling & Enabled \\
Reported validation metric & mean@8 \\
\bottomrule
\end{tabular}
\end{table}

\paragraph{Metric definition.}
For each validation problem, we sample 8 responses and compute the fraction of
correct responses under the same boxed-answer verifier. We report mean@8, i.e.,
the average correctness over the 8 sampled responses and then over all problems
in the benchmark. This differs from majority voting or pass@8.

\subsection{Attention-entropy instrumentation}

For each response token, we compute attention entropy from the actor model
during training. We use one fixed mid-layer attention probe and normalized
attention entropy as the default score. Within the captured layer, we average
attention probabilities over all heads before computing entropy, yielding one
layer-level attention distribution per response token. We do not average entropy
across multiple layers in the main experiments. Attention entropy is used only to
construct token weights or diagnostic statistics; it is detached from the
optimization graph and is not used as an auxiliary loss.

\begin{table}[h]
\centering
\small
\caption{Attention-entropy configuration.}
\label{tab:attention_entropy_config}
\begin{tabular}{@{}p{0.34\linewidth}p{0.58\linewidth}@{}}
\toprule
Item & Setting \\
\midrule
Enable attention entropy & True \\
Captured layer & One fixed mid-layer probe (Layer 20 in our implementation) \\
Head aggregation & Average attention probabilities over all heads in the captured layer \\
Layer aggregation & None in main experiments; the layer is fixed before comparison within each setup \\
Default entropy score & Normalized attention entropy \\
Use normalized entropy & True \\
Store full attention probabilities & False \\
Compute attention entropy on CPU & False \\
Entropy variants & Normalized, raw, top-$k$, and fixed-window variants when enabled \\
Token entropy log frequency & Every 2 steps \\
Logged sequences per step & 3 \\
Maximum logged tokens per sequence & 256 \\
Entropy-detail save frequency & Every 2 steps \\
Support-size thresholds & $0.5$, $0.7$, $0.9$ \\
\bottomrule
\end{tabular}
\end{table}

For response token $y_t$, the entropy score is computed over visible positions
under the causal attention mask. The token partition is performed within each
generated response: anchor tokens are the bottom 20\% by normalized attention
entropy, and explorer tokens are the top 20\%. The entropy scores are recomputed
from the current rollout batch rather than fixed in advance.

\subsection{Token-weighting configurations}

All training variants optimize the same token-level RL objective but differ in
the response-token weights $w_t$. Full-token training uses all valid response
tokens. Random-sparse training samples a uniformly random 20\% subset of response
tokens. Anchor-only and explorer-only training select the bottom and top 20\%
tokens by normalized attention entropy within each response, respectively.

For the soft-reweighting validation intervention, we use attention entropy to
assign continuous advantage weights instead of binary masks. The implementation
uses the decision attention entropy for each generated token, applies a
temperature-scaled softmax over valid response tokens in the rollout batch,
min--max rescales the softmax scores, and maps them to scheduled low-entropy
and high-entropy endpoint weights. The main controlled schedule is Low2High: the
low-entropy endpoint is scheduled from $1.0$ to $0.0$, and the high-entropy
endpoint is scheduled from $0.0$ to $1.0$. All valid response tokens receive
continuous entropy-dependent weights rather than being split into a fixed
middle band and endpoint groups.

\begin{table}[h]
\centering
\small
\caption{Token-weighting and reweighting configuration.}
\label{tab:token_weighting_config}
\begin{tabular}{lll}
\toprule
Setting & Token selection / weight & Normalization \\
\midrule
Full-token & $w_t=1$ for all valid response tokens & Token mean \\
Random-20\% & Uniform Bernoulli mask with 20\% budget & Selected-token or control-specific \\
Anchor-only & Bottom 20\% by normalized attention entropy & Selected-token normalization \\
Explorer-only & Top 20\% by normalized attention entropy & Selected-token normalization \\
Soft intervention & Continuous entropy-based advantage weights & All-token-weighted normalization \\
\bottomrule
\end{tabular}
\end{table}

\paragraph{Normalization variants.}
For hard-masked selective training, the main normalization uses the selected
tokens as the denominator, preserving the per-token loss scale of the retained
tokens. For normalization controls and the smooth soft-reweighting intervention,
we use all-token-weighted normalization:
\begin{equation}
\mathcal{L}^{\mathrm{all}}_w
=
\frac{1}{T}
\sum_{t=1}^{T} w_t \ell_t ,
\end{equation}
where $T$ is the number of valid response tokens. This variant keeps the
response-level loss scale comparable across different weighting rules. We report
normalization-specific results separately when they are used.

\subsection{Control configurations}

To test whether the anchor--explorer distinction is explained by simpler token
properties, we additionally evaluate control masks. Position controls select the
first or last 20\% of response tokens. Prediction-entropy controls select tokens
by next-token predictive entropy. Objective-magnitude controls select tokens by
the magnitude of their effective loss contribution. Type-matched and
position-matched random controls preserve coarse token-type or relative-position
statistics while removing the attention-entropy ranking. These controls are
defined in detail in Appendix~\ref{app:control_configs}.

\subsection{Compute Resources}
\label{app:compute_resources}

All experiments were conducted on an internal GPU cluster. Each node contains
8 GPUs, so the 2-node and 4-node settings correspond to 16 and 32 GPUs per run,
respectively. The GPUs used in our experiments were NVIDIA H800 with 80GB memory.

For the full-data soft-reweighting intervention experiments, including the
matched full-token baselines and schedule/layer-source variants, each run used
4 nodes, i.e., 32 GPUs, and took approximately 18 wall-clock hours. This
corresponds to roughly $32 \times 18 = 576$ GPU-hours per run. For the
selective-training probe experiments on the smaller diagnostic data split, each
run used 2 nodes, i.e., 16 GPUs, and took approximately 22 wall-clock hours,
corresponding to roughly $16 \times 22 = 352$ GPU-hours per run.

Across the reported intervention runs, probe runs, multi-seed diagnostics,
schedule ablations, and failed or preliminary runs used to diagnose instability,
we estimate the total compute to be on the order of
$1\times 10^{4}$--$2\times 10^{4}$ GPU-hours. This estimate is intended
to characterize the scale of compute required to reproduce the experimental
study; small post-processing jobs for plotting, metric aggregation, and
visualization are excluded.

\subsection{Implementation details omitted from the paper}

We omit machine-specific paths, proxy settings, process-management commands, and
logging credentials from the paper and released configuration. These details do
not affect the algorithmic setup or the reported results.

\section{Sparse Estimability Derivation}
\label{app:sparse_derivation}

Let $g_t=\nabla_\theta \ell_t$ denote the per-token gradient and $g_{\mathrm{full}}=\frac{1}{T}\sum_{t=1}^T g_t$. Under uniform random subsampling with inclusion probability $p$, each token is selected by $m_t\sim\mathrm{Bernoulli}(p)$ and the subset gradient is $g_{\mathrm{rand}}=\frac{1}{\sum_t m_t}\sum_t m_t g_t$ when the subset is non-empty. Conditional on a fixed subset size, this estimator is unbiased by symmetry. For large $T$, $\sum_t m_t$ concentrates near $pT$, yielding the approximation
\begin{align}
\mathbb{E}[g_{\mathrm{rand}}] &\approx g_{\mathrm{full}},\\
\mathbb{E}\|g_{\mathrm{rand}}-g_{\mathrm{full}}\|^2 &\approx \frac{1-p}{pT}\cdot \overline{V},
\end{align}
where $\overline{V}=\frac{1}{T}\sum_t\|g_t\|^2$. The expected directional agreement can be approximated as
\begin{equation}
\mathbb{E}\bigl[\operatorname{Cosine}(g_{\mathrm{rand}},g_{\mathrm{full}})\bigr]
\approx
\frac{1}{\sqrt{1+(1-p)\overline{V}/(pT\|g_{\mathrm{full}}\|^2)}}.
\end{equation}
This argument treats token gradients as fixed while analyzing mask randomness and does not apply to structured subsets, whose inclusion probabilities depend on token properties.

\begin{figure}[t]
    \centering
    \begin{subfigure}[t]{0.45\linewidth}
        \centering
        \includegraphics[width=\linewidth]{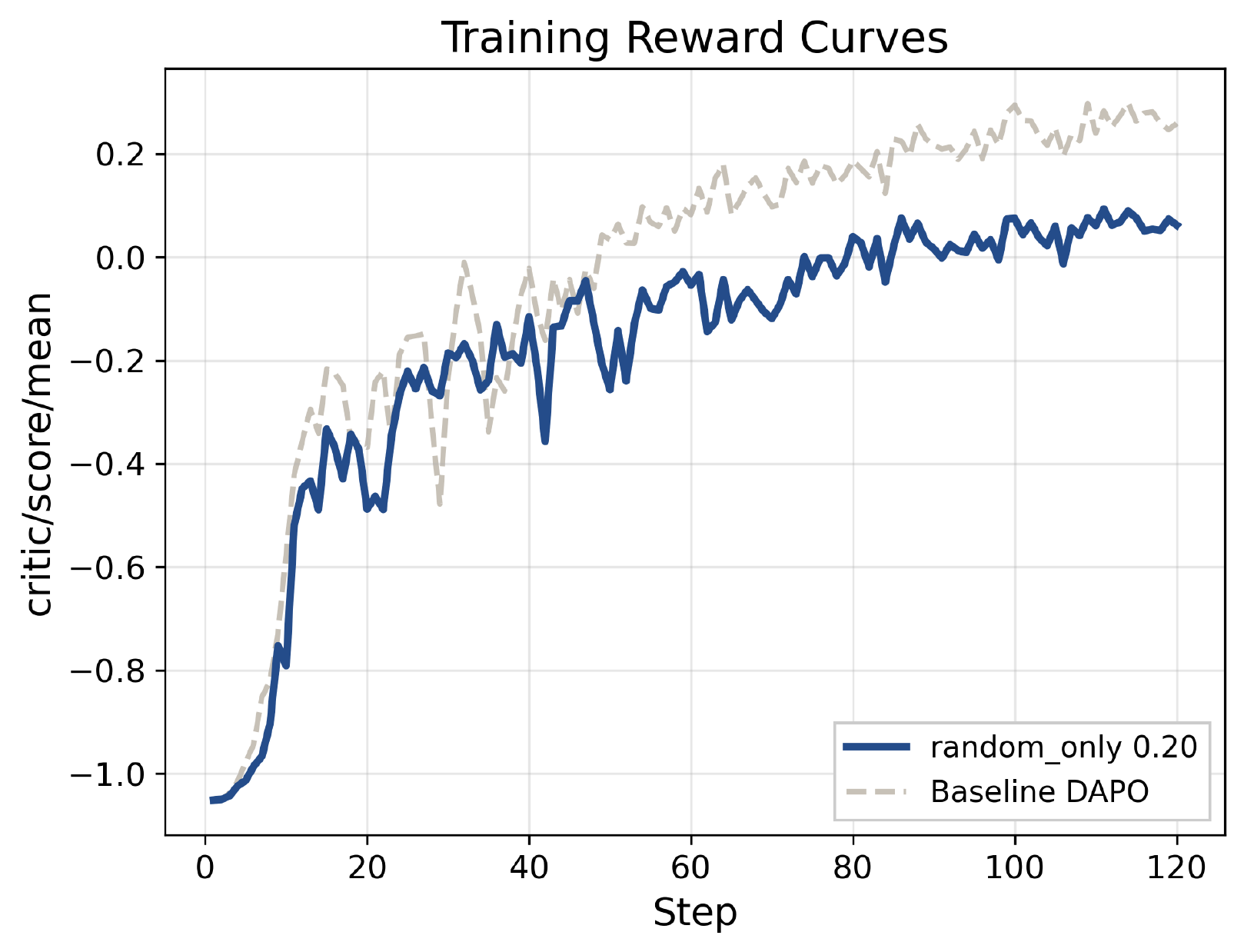}
        \caption{Training reward.}
        \label{fig:sparse_training_reward}
    \end{subfigure}
    \hfill
    \begin{subfigure}[t]{0.45\linewidth}
        \centering
        \includegraphics[width=\linewidth]{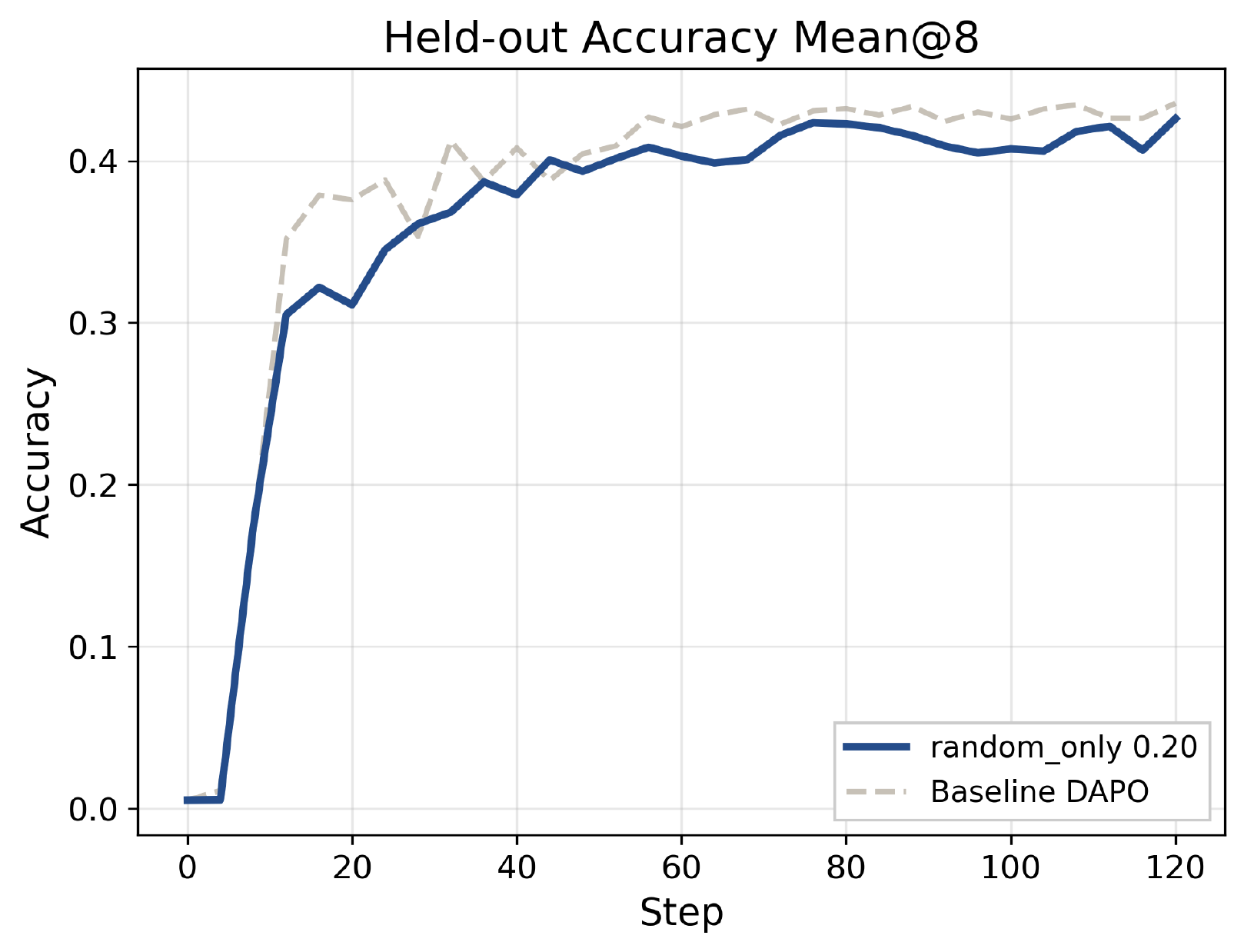}
        \caption{External held-out average.}
        \label{fig:sparse_heldout}
    \end{subfigure}

    \caption{
    Token-level RL objectives are sparsely estimable.
    (a) On training-side reward, random-$20\%$ token training (gray) lags behind full-token training (black), reflecting residual variance from sparse gradient estimation.
    (b) On external held-out benchmarks, random-$20\%$ training recovers much of full-token performance.
    }
    \label{fig:sparse_estimability}
\end{figure}

\section{Additional Entropy Definitions and Implementation Details}

\subsection{Additional Attention Entropy Variants}
\label{app:entropy_details}
In the main paper, we use normalized attention entropy as the default score.
Here we provide additional variants used in robustness analyses. For response
token $y_t$ at layer $\ell$, let $a_{t,j}^{(\ell)}$ denote the layer-level
attention probability assigned to visible position $j$ after averaging
attention probabilities over all heads in that layer, with
\begin{equation}
\sum_{j=1}^{N_t} a_{t,j}^{(\ell)}=1.
\end{equation}

\paragraph{Raw attention entropy.}
\begin{equation}
H_t^{\text{raw}}=-\sum_{j=1}^{N_t} a_{t,j}^{(\ell)}\log a_{t,j}^{(\ell)}.
\end{equation}

\paragraph{Normalized attention entropy.}
\begin{equation}
H_t^{\text{norm}}=\frac{H_t^{\text{raw}}}{\log N_t}.
\end{equation}

\paragraph{Top-$k$ attention entropy.}
Select the $k$ largest attention weights, renormalize, and compute entropy:
\begin{equation}
\tilde{a}^{(k)}_{t,j}
=
\frac{a_{t,j}^{(\ell)}}{\sum_{j' \in S_k(t)} a_{t,j'}^{(\ell)}},
\qquad
H_t^{\text{top-}k}
=
-\sum_{j\in S_k(t)} \tilde{a}^{(k)}_{t,j}\log \tilde{a}^{(k)}_{t,j}.
\end{equation}

\paragraph{Fixed-position attention entropy.}
Restrict to first $K=256$ visible positions and renormalize:
\begin{equation}
\hat{a}^{(K)}_{t,j}
=
\frac{a_{t,j}^{(\ell)}}{\sum_{j' \in F_K(t)} a_{t,j'}^{(\ell)}},
\qquad
H_t^{\text{fix-}K}
=
-\sum_{j\in F_K(t)} \hat{a}^{(K)}_{t,j}\log \hat{a}^{(K)}_{t,j}.
\end{equation}

Unless otherwise specified, analyses use the same fixed mid-layer probe across
compared variants. Appendix~\ref{app:layer_wise_attention} discusses the scope
of this implementation choice and layer-wise follow-up diagnostics.

\subsection{Entropy-based Token Partition}
\label{app:entropy_based_partition}
For each response, we compute $H_t^{\text{norm}}$ for response tokens only and rank within-sample. Given threshold $p$:
\begin{equation}
\mathcal{T}_{\text{explorer}}^{(p)}(y)
=
\mathrm{Top}\text{-}p\%\bigl(\mathcal{T}(y); H_t^{\text{norm}}\bigr),
\qquad
\mathcal{T}_{\text{anchor}}^{(p)}(y)
=
\mathrm{Bottom}\text{-}p\%\bigl(\mathcal{T}(y); H_t^{\text{norm}}\bigr).
\end{equation}
We use $p=20\%$ throughout. The within-sample partition focuses on relative functional role rather than absolute entropy magnitude.

\subsection{Token Weighting and Normalization Details}
\label{app:token_weighting_normalization}
For hard token masks, the weighted objective
\begin{equation}
\mathcal{L}_{w}
=
\frac{\sum_{t=1}^{T} w_t \ell_t}{\sum_{t=1}^{T} w_t}
\end{equation}
normalizes by weight sum rather than total token count, avoiding gradient shrinkage when only a small subset is selected.

\paragraph{Hard mask.} $w_t \in \{0,1\}$.

\paragraph{Random sparse mask.} $w_t \sim \mathrm{Bernoulli}(p)$ with $p=0.2$.

\paragraph{Smooth soft reweighting.} The intervention in
Section~\ref{sec:soft_weighting} uses continuous weights
$w_t \in \mathbb{R}_{\ge 0}$ to rescale token-level advantages, but keeps the
actor-loss denominator equal to the number of valid response tokens. This
all-token-weighted normalization is used to avoid changing the denominator when
the scheduled endpoint weights change.

\section{Layer-Wise Attention Patterns and Scope of the Fixed-Layer Probe}
\label{app:layer_wise_attention}

The main analysis uses attention entropy from one fixed mid-layer probe. This is
an implementation choice made to keep all compared variants matched, not a claim
that this layer is uniquely optimal or that all transformer layers expose the
same token partition. Different layers can play different roles during reasoning
generation. Earlier layers often retain more local lexical, syntactic, and
positional regularities; middle layers are more likely to mix local computation
with task-level state tracking; later layers can become more specialized toward
generation decisions, output format, and answer expression. Because attention
entropy measures the concentration of contextual support inside a particular
layer, its numerical value and the resulting anchor/explorer partition may vary
across layers.

This layer dependence is a feature of the diagnostic rather than a contradiction
of the main results. The paper uses attention entropy to expose token-level
heterogeneity in a controlled setting, not to claim that a single layer captures
all forms of reasoning support. The fixed-layer probe provides one consistent
view of support concentration that can be connected to selective training,
support-size statistics, entropy dynamics, and gradient geometry. A full
layer-wise intervention sweep is left for future work.

Several lightweight follow-up checks can be run before a full intervention
sweep. First, one can compute anchor/explorer mask agreement across a small set
of early, middle, and late layers using Jaccard overlap or rank correlation of
token entropies within each response. Low agreement would
show that different layers identify different support regimes, while high
agreement around neighboring middle layers would support the stability of the
fixed-layer probe. Second, the evidence-gathering statistics in
Section~\ref{sec:evidence_gathering} can be recomputed per layer: support-size
thresholds, average attention distance, local-window mass, and non-local mass
are inexpensive diagnostics once attention probabilities have been logged. Third,
one can run the gradient-geometry probe from Section~\ref{sec:gradient_quality}
with masks produced by neighboring layers. This is cheaper than full RL
training and directly tests whether the directional-alignment patterns are
specific to the chosen probe layer or stable across nearby layers. These checks
would clarify which aspects of attention entropy are layer-specific and which
are robust properties of token-level RL signal heterogeneity.

\section{Alternative-Explanation Control Configurations}
\label{app:control_configs}

This section defines the control masks used to test whether the optimization
roles of attention-entropy-defined token groups can be explained by simpler
token properties. We focus on three alternative explanations: token position,
prediction uncertainty, and token-level objective magnitude. All controls use
the same $20\%$ sparsity ratio as the anchor/explorer masks and are computed
within each response unless otherwise specified.

\subsection{Position Controls}
\label{app:position_controls}

\begin{figure*}[t]
    \centering

    \begin{subfigure}[t]{0.48\linewidth}
        \centering
        \includegraphics[width=\linewidth]{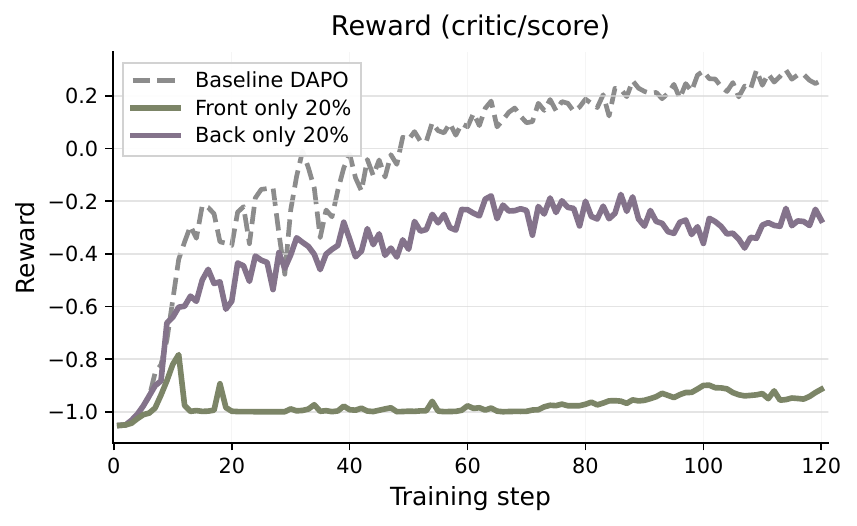}
        \caption{Training reward.}
    \end{subfigure}
    \hfill
    \begin{subfigure}[t]{0.48\linewidth}
        \centering
        \includegraphics[width=\linewidth]{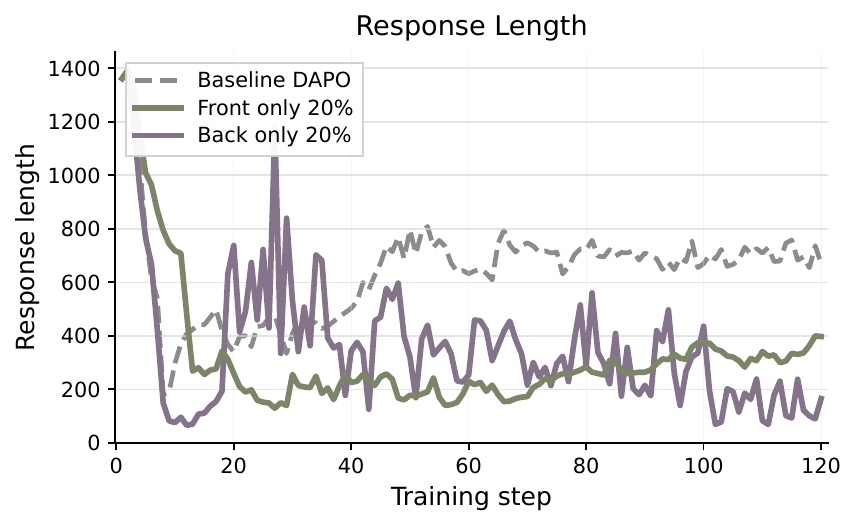}
        \caption{Response length.}
    \end{subfigure}

    \vspace{0.6em}

    \begin{subfigure}[t]{0.48\linewidth}
        \centering
        \includegraphics[width=\linewidth]{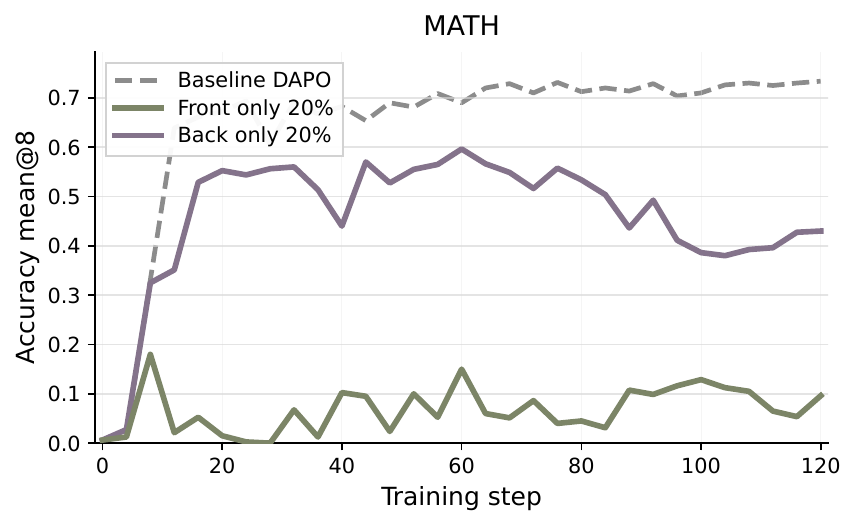}
        \caption{MATH.}
    \end{subfigure}
    \hfill
    \begin{subfigure}[t]{0.48\linewidth}
        \centering
        \includegraphics[width=\linewidth]{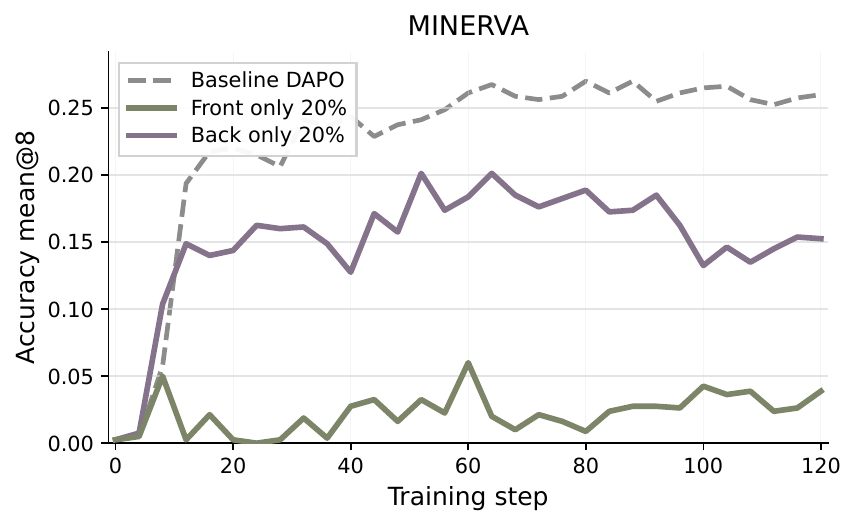}
        \caption{MINERVA.}
    \end{subfigure}

    \vspace{0.6em}

    \begin{subfigure}[t]{0.48\linewidth}
        \centering
        \includegraphics[width=\linewidth]{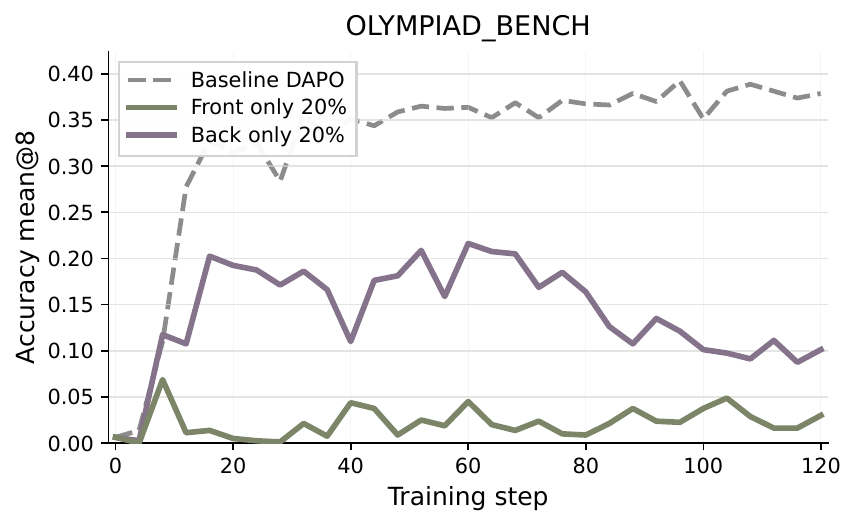}
        \caption{OLYMPIAD-BENCH.}
    \end{subfigure}
    \hfill
    \begin{subfigure}[t]{0.48\linewidth}
        \centering
        \includegraphics[width=\linewidth]{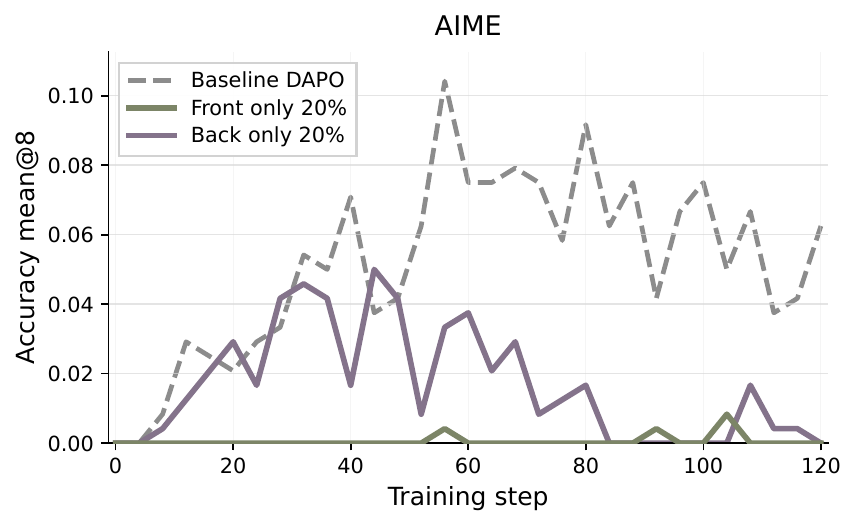}
        \caption{AIME.}
    \end{subfigure}

    \caption{
    Position-only controls on Qwen3-8B-Base. These controls test whether attention
    entropy is merely a proxy for response position. We compare full-token DAPO with
    training only the first $20\%$ or the last $20\%$ response tokens. Front-only
    training largely fails, suggesting that early response tokens alone cannot
    support effective RLVR optimization. Back-only training is stronger, consistent
    with answer-proximal tokens carrying more direct outcome-relevant signal, but it
    remains substantially below full-token DAPO and shows late-stage degradation on
    harder benchmarks. The failure of fixed-position masks to reproduce the
    entropy-based behavior indicates that attention entropy captures more than
    token position.
    }
    \label{fig:position_control_curves}
\end{figure*}

A key possible confound is that attention entropy may be a proxy for token
position. For example, low-entropy or high-entropy tokens might appear more
frequently in certain regions of the response, in which case entropy-based
selection could be explained by a simpler beginning-versus-end-of-response
effect. To test this alternative explanation, we construct position-only
controls that ignore attention entropy entirely.

For a response of length $T$, we define two fixed-position masks:
\begin{equation}
\mathcal{T}_{\mathrm{front}}^{20\%}
= \{t: t/T \le 0.2\},
\qquad
\mathcal{T}_{\mathrm{back}}^{20\%}
= \{t: t/T > 0.8\}.
\end{equation}
The front-only control updates only the first $20\%$ response tokens, while the
back-only control updates only the last $20\%$ response tokens. If attention
entropy were merely acting as a positional proxy, these simple position-based
masks should reproduce the main behaviors of entropy-defined anchor or explorer
training.

Figure~\ref{fig:position_control_curves} shows the training dynamics of these
position controls on Qwen3-8B-Base. Front-only training almost fails to learn:
the reward remains close to the low-reward regime, response length quickly
shrinks, and validation accuracy stays near zero across benchmarks. Back-only
training is stronger than front-only training, indicating that answer-proximal
tokens carry more direct outcome-relevant signal. However, it remains far below
full-token DAPO and often degrades in later training, especially on harder
benchmarks such as AIME and OLYMPIAD-BENCH.

These results rule out the interpretation that attention entropy is simply a
proxy for response position. Position-only selection chooses a contiguous
segment of the response and captures only where a token appears. In contrast,
attention entropy selects tokens according to how they aggregate information
from prior context. Low-entropy anchor tokens are therefore not merely early or
late tokens; they provide stable concentrated-support updates across the
response. Similarly, high-entropy explorer tokens are not equivalent to
answer-proximal tokens; they reflect diffuse evidence aggregation and stronger
but less stable optimization signals. Thus, the anchor/explorer distinction is
not reducible to a trivial positional bias.

\subsection{Prediction-Entropy Controls and Diagnostic Comparison}
\label{app:prediction_entropy_controls}

Another possible explanation is that attention entropy merely reflects output
uncertainty. To separate attention-support entropy from prediction uncertainty,
we define predictive entropy using the model's next-token distribution before
observing token $y_t$:
\begin{equation}
H_t^{\mathrm{pred}}
=
-\sum_{v \in \mathcal{V}}
p_\theta(v \mid x,y_{<t})
\log p_\theta(v \mid x,y_{<t}).
\label{eq:prediction_entropy}
\end{equation}
This quantity measures uncertainty over the next-token vocabulary distribution.
In contrast, attention entropy is computed over the attention distribution across
visible context positions and measures how concentrated or diffuse the token's
contextual support is:
\begin{equation}
H_t^{\mathrm{attn}}
=
-\sum_{j=1}^{N_t} a_{t,j}\log a_{t,j},
\qquad
H_t^{\mathrm{attn,norm}}
=
\frac{H_t^{\mathrm{attn}}}{\log N_t}.
\label{eq:attention_entropy_control_appendix}
\end{equation}
Thus, predictive entropy asks whether the model is uncertain about what token to
generate next, whereas attention entropy asks whether the generated token is
supported by a concentrated or diffuse set of prior context positions.

\paragraph{Prediction-entropy selective controls.}
We construct prediction-low and prediction-high controls by selecting the bottom
and top $20\%$ response tokens ranked by $H_t^{\mathrm{pred}}$ within each
response:
\begin{equation}
\mathcal{T}_{\mathrm{pred\text{-}low}}^{20\%}
=
\operatorname{Bottom}_{20\%}
\left(\{H_t^{\mathrm{pred}}\}_{t=1}^{T}\right),
\qquad
\mathcal{T}_{\mathrm{pred\text{-}high}}^{20\%}
=
\operatorname{Top}_{20\%}
\left(\{H_t^{\mathrm{pred}}\}_{t=1}^{T}\right).
\label{eq:prediction_entropy_controls}
\end{equation}
We then repeat the same selective-training setup using these
prediction-entropy-defined token groups in place of the attention-entropy-defined
groups.

\begin{figure*}[t]
\centering
\begin{subfigure}{0.45\linewidth}
    \centering
    \includegraphics[width=\linewidth]{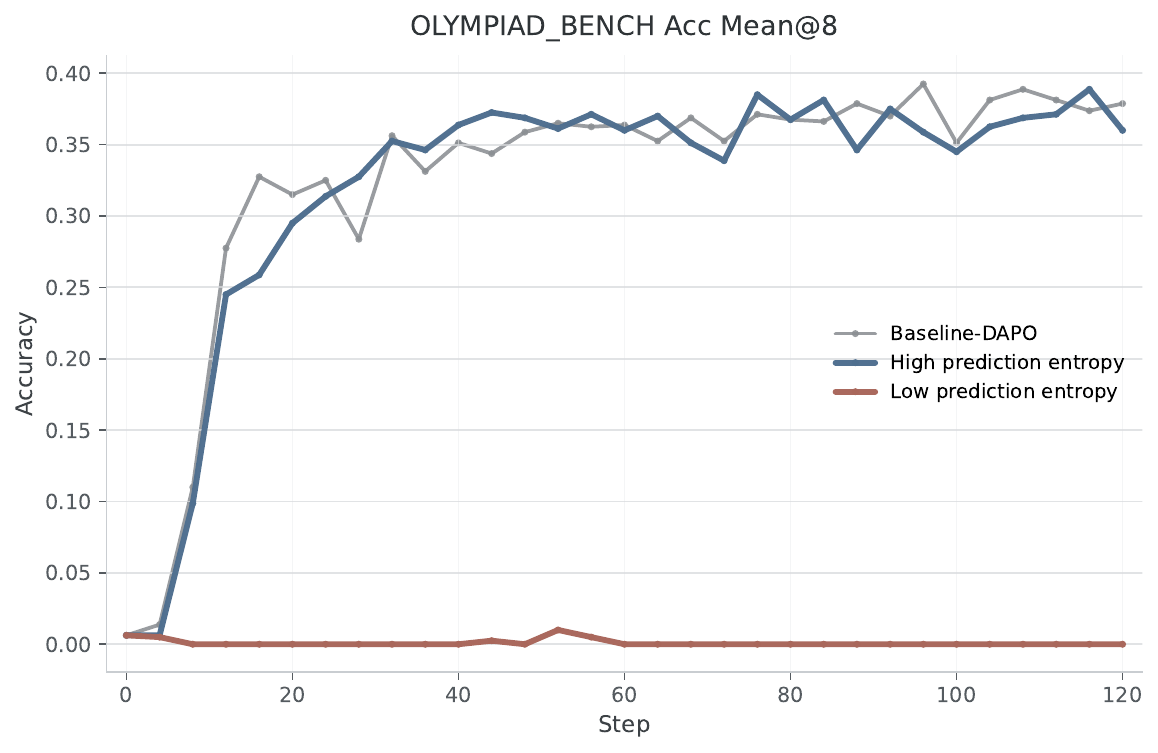}
    \caption{OlympiadBench}
\end{subfigure}
\hfill
\begin{subfigure}{0.45\linewidth}
    \centering
    \includegraphics[width=\linewidth]{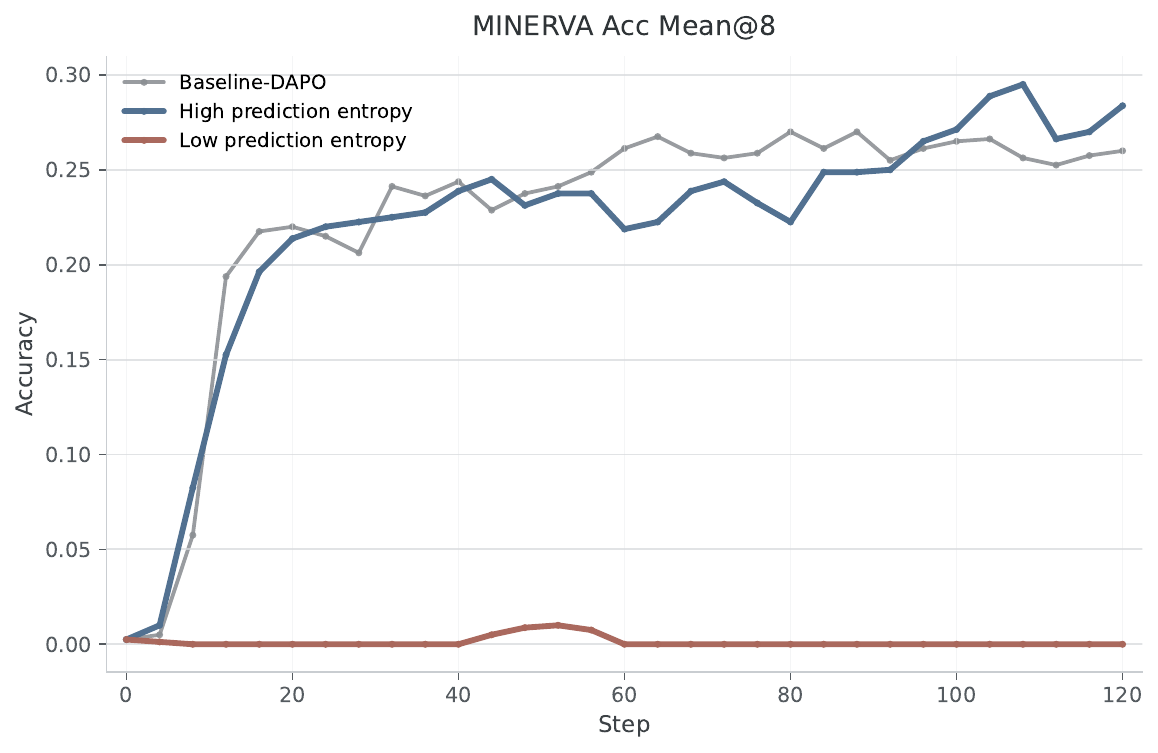}
    \caption{Minerva}
\end{subfigure}

\vspace{0.6em}

\begin{subfigure}{0.45\linewidth}
    \centering
    \includegraphics[width=\linewidth]{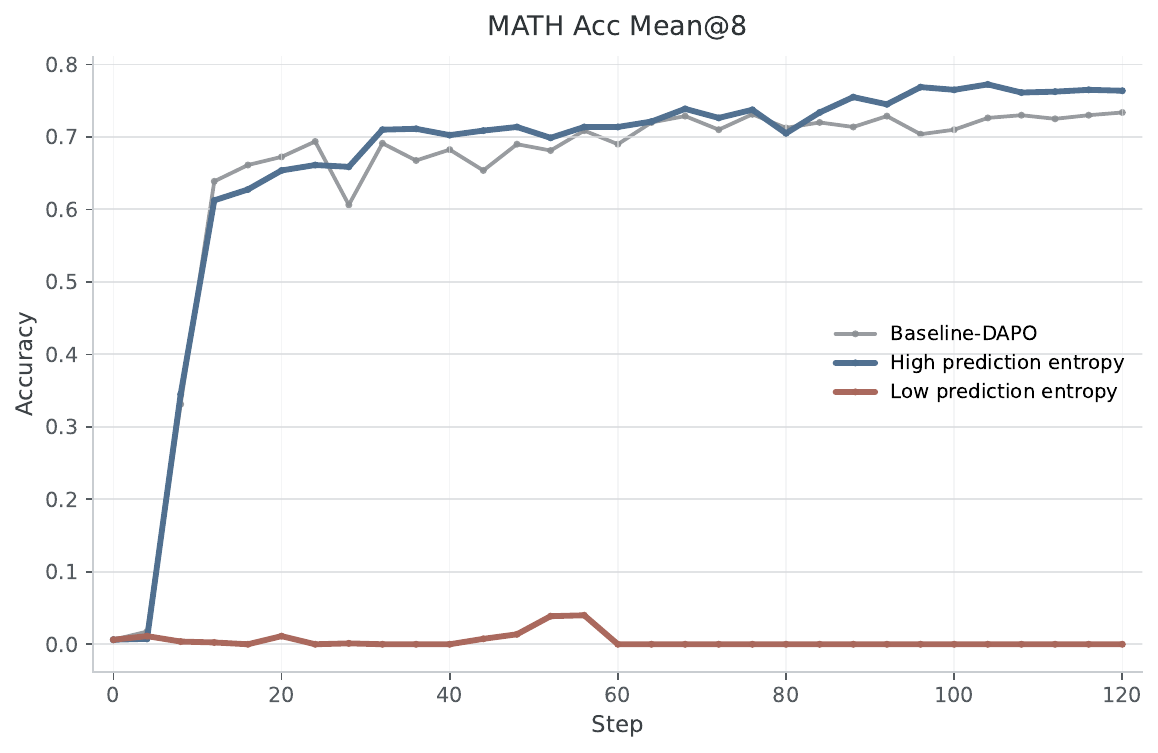}
    \caption{MATH}
\end{subfigure}
\hfill
\begin{subfigure}{0.45\linewidth}
    \centering
    \includegraphics[width=\linewidth]{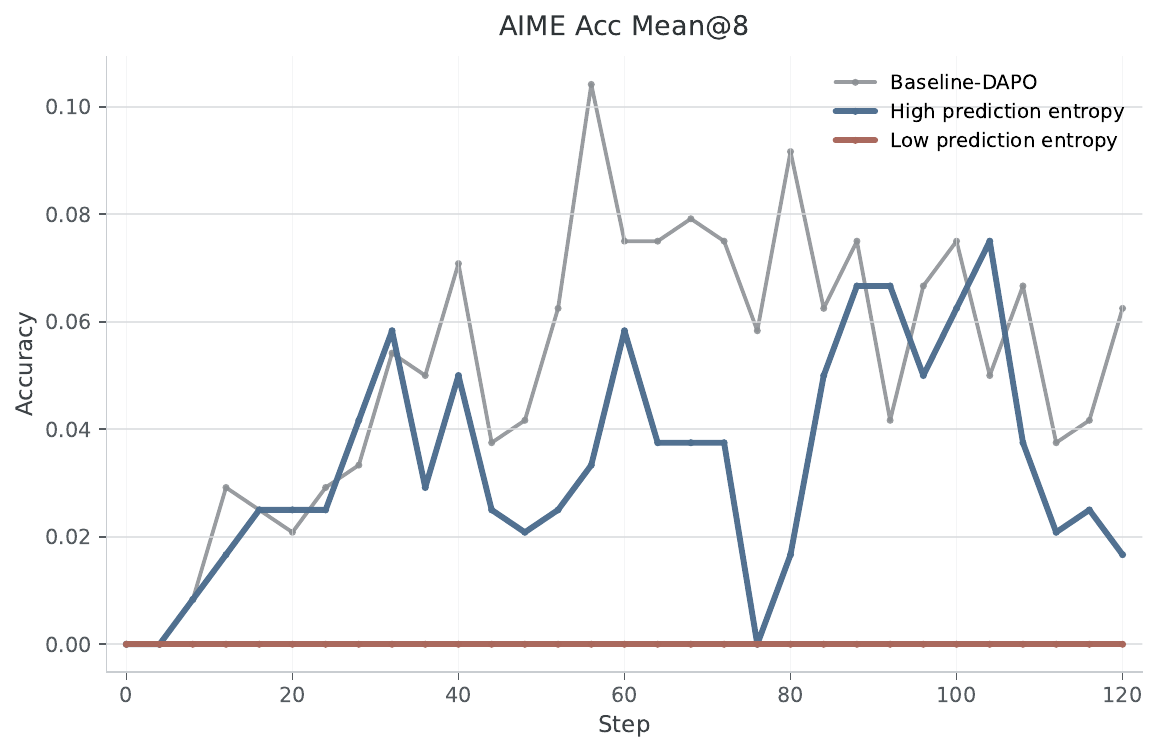}
    \caption{AIME}
\end{subfigure}

\vspace{0.6em}

\begin{subfigure}{0.45\linewidth}
    \centering
    \includegraphics[width=\linewidth]{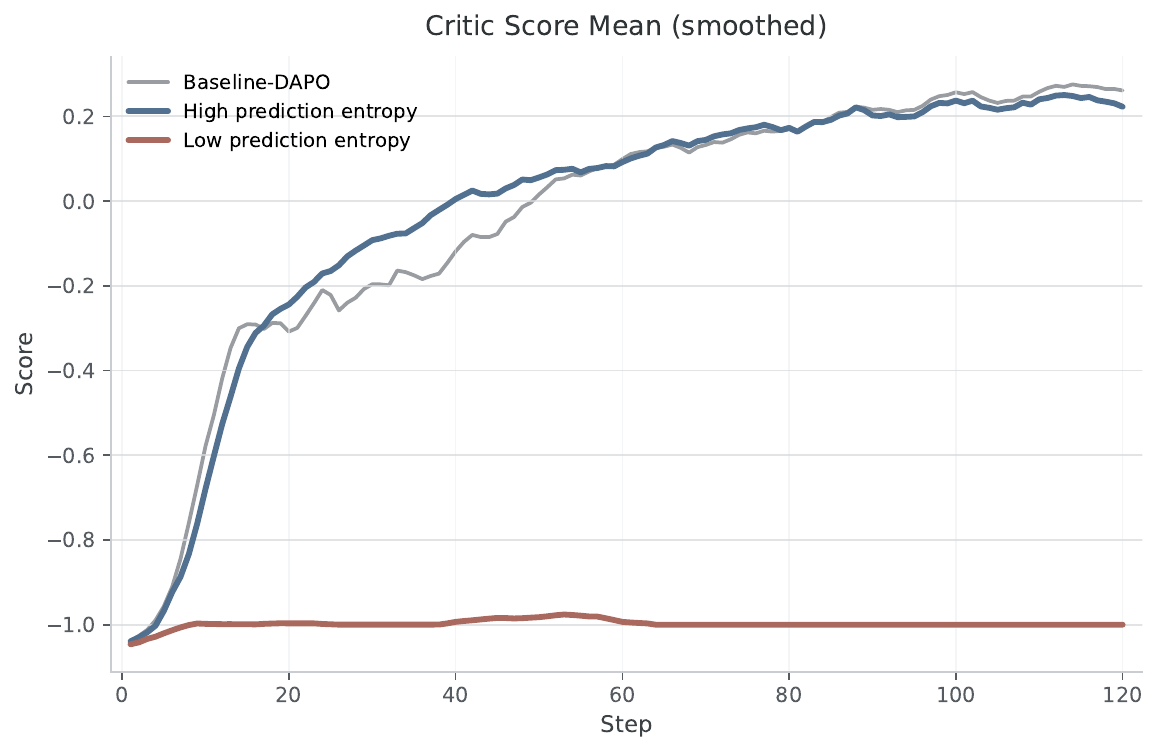}
    \caption{Critic score}
\end{subfigure}
\hfill
\begin{subfigure}{0.45\linewidth}
    \centering
    \includegraphics[width=\linewidth]{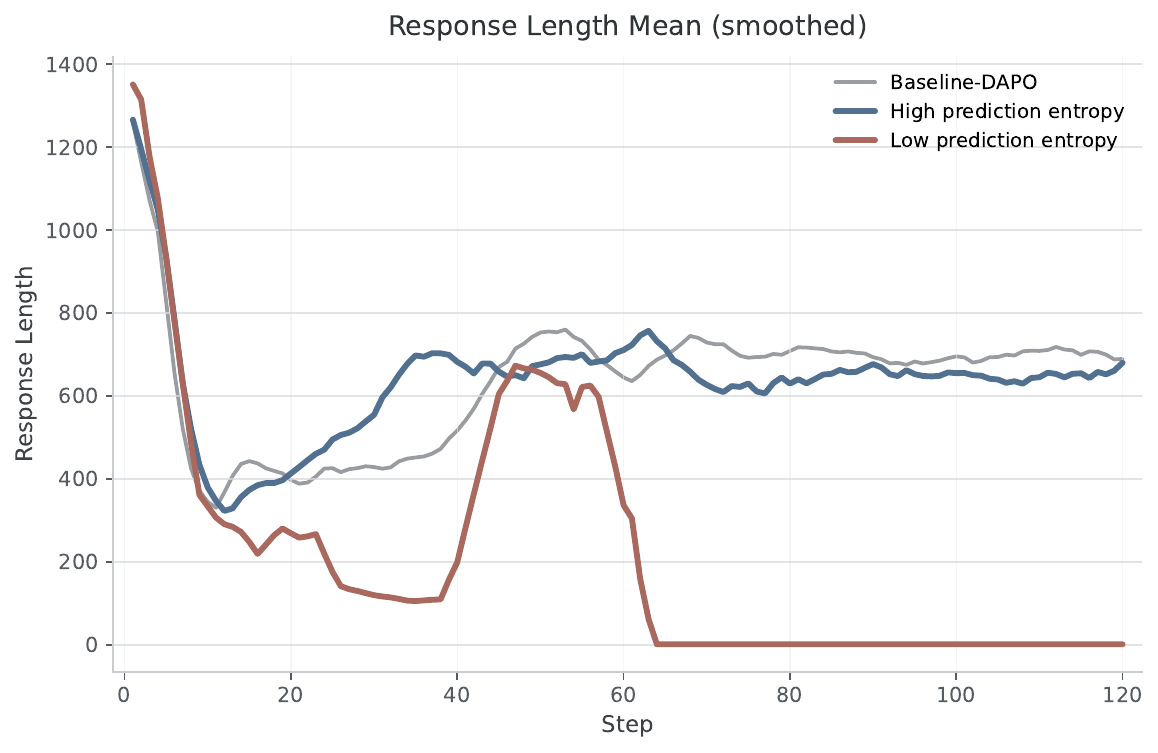}
    \caption{Response length}
\end{subfigure}

\caption{
Prediction-entropy controls for selective token training. Tokens are selected by
the entropy of the model's next-token distribution rather than by attention
entropy. The high-prediction-entropy and low-prediction-entropy subsets exhibit
different training trajectories, separating output-distribution uncertainty from
attention-support entropy.
}
\label{fig:prediction_entropy_controls}
\end{figure*}

Figure~\ref{fig:prediction_entropy_controls} shows that grouping tokens by
prediction entropy leads to a training profile that is different from the
attention-entropy selective-training curves in
Figure~\ref{fig:selective_training_main}. Under the prediction-entropy
partition, the high-prediction-entropy subset exhibits sustained changes in
benchmark accuracy and critic score, while the low-prediction-entropy subset
stays close to the floor on most metrics. Its response length also decreases
sharply and eventually collapses to nearly zero. This pattern indicates that
prediction-entropy selection mainly separates tokens according to uncertainty in
the next-token output distribution.

This differs from the behavior induced by the attention-entropy partition. In
the attention-entropy selective-training curves, low-attention-entropy and
high-attention-entropy token groups show distinct optimization dynamics in terms
of reward evolution, held-out accuracy, AIME behavior, response-length changes,
and collapse patterns. The prediction-entropy controls do not reproduce this
same form of separation. Instead, they reflect a different criterion: whether the
model's next-token distribution is sharp or uncertain at the current position.

Therefore, prediction entropy and attention entropy should be interpreted as two
different token-level quantities. Prediction entropy characterizes uncertainty
over the vocabulary distribution, whereas attention entropy characterizes the
structure of contextual support across prior positions. The difference between
the two sets of selective-training curves suggests that attention entropy is not
fully explained by output uncertainty alone.

\begin{figure}[t]
    \centering
    \includegraphics[width=0.78\linewidth]{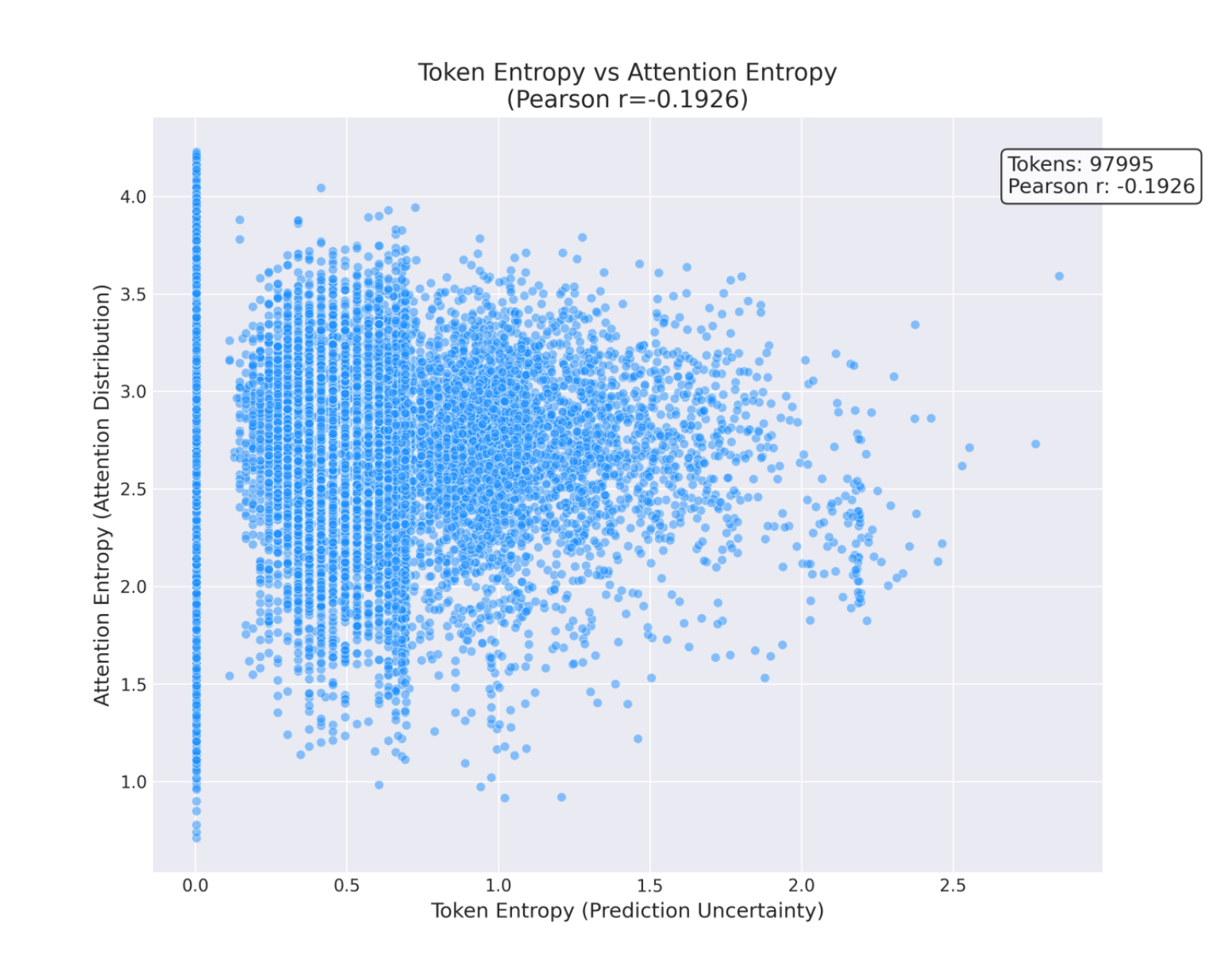}
    \caption{
    Token-level relation between predictive entropy and attention entropy.
    Each point corresponds to one response token. Across $97{,}995$ tokens,
    the Pearson correlation is $r=-0.1926$, indicating a weak negative
    relationship rather than a strong alignment between the two scores.
    The broad dispersion of points shows that tokens with similar predictive
    entropy can have substantially different attention entropy, and vice versa.
    Thus, attention entropy is not simply a relabeling of next-token prediction
    uncertainty.
    }
    \label{fig:attn_pred_scatter}
\end{figure}

\paragraph{Global relation between attention entropy and prediction entropy.}
We first examine whether the two entropy scores define the same token ordering.
Figure~\ref{fig:attn_pred_scatter} plots token-level predictive entropy against
attention entropy over response tokens. Across $97{,}995$ tokens, the Pearson
correlation is $r=-0.1926$, indicating only a weak negative linear relationship
between the two quantities. This weak correlation suggests that predictive
uncertainty and attention-support diffuseness are not interchangeable token-level
signals.

The scatter structure shows that attention entropy and predictive entropy are
related but non-equivalent token-level signals. Predictive entropy varies with
the uncertainty of the output distribution, while attention entropy varies with
the concentration or diffuseness of contextual support. Therefore, a token can be
prediction-uncertain but attention-concentrated, or prediction-certain but
attention-diffuse. The weak correlation in Figure~\ref{fig:attn_pred_scatter}
supports the view that attention-entropy-defined anchor/explorer groups cannot
be reduced to low- or high-prediction-entropy tokens.

\begin{figure}[t]
    \centering
    \includegraphics[width=\linewidth]{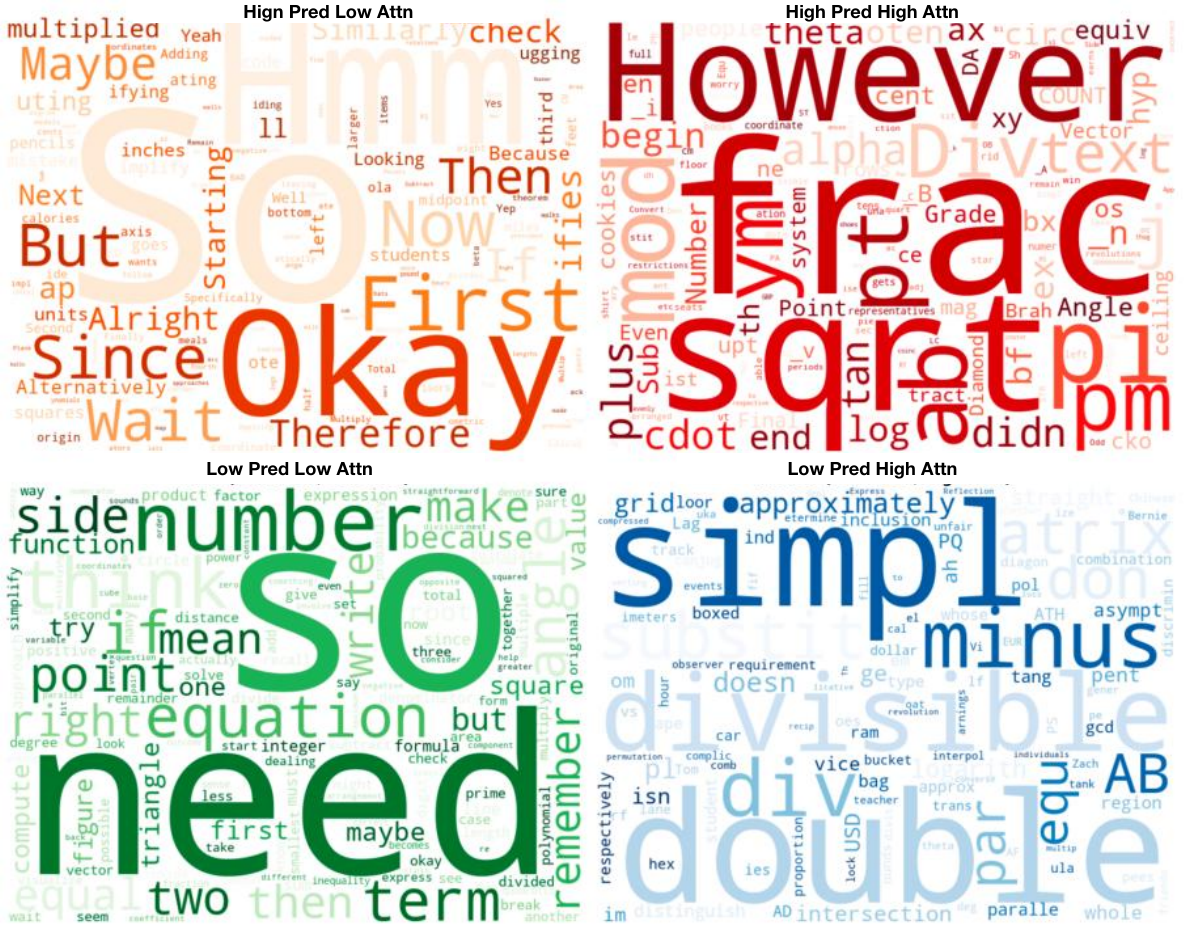}
    \caption{
    Quadrant analysis of predictive entropy and normalized attention entropy.
    Tokens are partitioned by within-response median thresholds. Word size
    indicates token frequency within each quadrant. The low-prediction /
    low-attention-entropy quadrant is dominated by routine reasoning-flow tokens;
    the high-prediction / low-attention-entropy quadrant contains discourse and
    planning tokens where the next-token choice is uncertain despite concentrated
    support; the low-prediction / high-attention-entropy quadrant contains
    predictable but context-diffuse operation tokens; and the high-prediction /
    high-attention-entropy quadrant contains many symbolic or formula-heavy
    tokens with both output uncertainty and diffuse support. These qualitatively
    different token populations show that attention entropy does not simply
    relabel prediction entropy.
    }
    \label{fig:attn_pred_quadrants}
\end{figure}

\paragraph{Quadrant analysis.}
To make the distinction more interpretable, we divide response tokens into four
quadrants using within-response percentile thresholds for both entropy scores.
Let $\mathrm{HighPred}$ and $\mathrm{LowPred}$ denote tokens above and below the
median predictive entropy within the same response, and let $\mathrm{HighAttn}$
and $\mathrm{LowAttn}$ denote tokens above and below the median normalized
attention entropy within the same response. This gives four token categories:
\begin{equation}
\begin{aligned}
Q_{\mathrm{LP,LA}} &= \mathrm{LowPred} \cap \mathrm{LowAttn}, \\
Q_{\mathrm{HP,LA}} &= \mathrm{HighPred} \cap \mathrm{LowAttn}, \\
Q_{\mathrm{LP,HA}} &= \mathrm{LowPred} \cap \mathrm{HighAttn}, \\
Q_{\mathrm{HP,HA}} &= \mathrm{HighPred} \cap \mathrm{HighAttn}.
\end{aligned}
\label{eq:entropy_quadrants}
\end{equation}

The resulting quadrants reveal qualitatively different token regimes rather
than a simple monotonic relation between the two entropy scores. As shown in
Figure~\ref{fig:attn_pred_quadrants}, the low-prediction / low-attention-entropy
quadrant is dominated by routine reasoning and problem-solving tokens, such as
``need'', ``number'', ``equation'', ``term'', ``so'', and ``because''. These
tokens are relatively predictable from the output distribution and rely on
concentrated contextual support, corresponding to a deterministic reasoning-flow
regime.

The high-prediction / low-attention-entropy quadrant contains many discourse,
planning, and transition tokens, such as ``Maybe'', ``Wait'', ``But'', ``Since'',
``First'', ``Then'', and ``okay''. These tokens are uncertain at the output
level, likely because multiple discourse continuations are plausible, yet their
attention support remains concentrated. This quadrant therefore represents
localized decision points: the model is uncertain about which token to emit, but
the contextual evidence used by the token is not diffuse.

The low-prediction / high-attention-entropy quadrant shows the opposite
off-diagonal pattern. It contains tokens such as ``simple'', ``possible'',
``divide'', ``minus'', ``part'', and other operation-related or relation-related
tokens. These tokens can be relatively easy to predict, while still aggregating
information from a broader set of prior positions. This indicates that low
predictive uncertainty does not imply concentrated attention support; some
apparently certain tokens still depend on diffuse contextual aggregation.

Finally, the high-prediction / high-attention-entropy quadrant contains many
formula-heavy, symbolic, or branching tokens, including fragments such as
``frac'', ``sqrt'', ``left'', ``right'', ``theta'', ``alpha'', ``begin'', and
``However''. This quadrant combines output uncertainty with diffuse contextual
support, and corresponds to a more ambiguous regime where both the next-token
choice and the contextual evidence distribution are less concentrated.

This quadrant view exposes cases that cannot be explained by output uncertainty
alone. The two off-diagonal quadrants are especially important. High-prediction /
low-attention-entropy tokens show that output uncertainty can arise even when a
token relies on concentrated contextual support. Conversely, low-prediction /
high-attention-entropy tokens show that a token can be easy to predict while
still drawing on diffuse contextual evidence. Therefore, prediction entropy and
attention entropy capture different axes of token behavior: the former measures
uncertainty over the next-token distribution, whereas the latter measures the
concentration of the contextual support used by the token representation.

\begin{figure}[t]
    \centering
    \includegraphics[width=\linewidth]{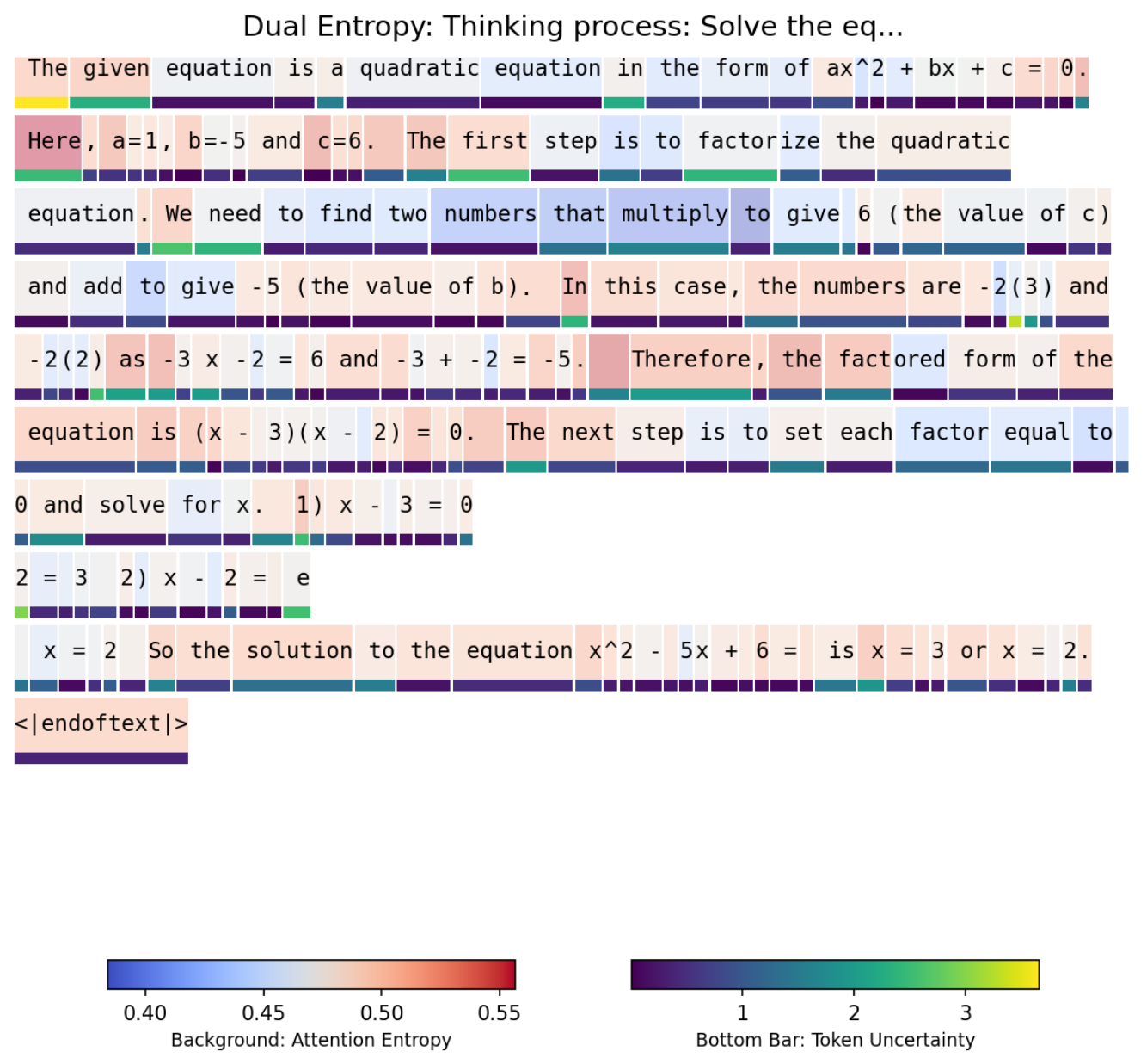}
    \caption{
    Sequence-level visualization of normalized attention entropy and predictive
    entropy along a reasoning trace. Background color denotes attention entropy,
    while the bottom bar denotes predictive entropy. Attention entropy is higher
    during initial state construction and reasoning-stage transitions, decreases
    during stable local computation, and can rise again when an intermediate
    result is integrated into the next step. Predictive entropy instead
    highlights local uncertainty over the next-token choice; for example, the
    erroneous token ``e'' in ``$x - 2 = e$'' appears with high predictive entropy.
    The two signals therefore capture different aspects of token-level behavior.
    }
    \label{fig:dual_entropy_sequence}
\end{figure}

\paragraph{Sequence-level dynamics.}
We also visualize both entropy scores along individual reasoning trajectories.
Figure~\ref{fig:dual_entropy_sequence} shows one representative response, where
the token background color indicates normalized attention entropy and the bottom
bar indicates predictive entropy. This visualization reveals that the two
signals follow different temporal patterns within the same reasoning trace.

At the beginning of the response, attention entropy is relatively high. This
corresponds to an initial state-construction phase, where the model identifies
the equation type, maps the coefficients, and establishes the problem-solving
context. As the response enters routine algebraic computation, attention entropy
tends to decrease, suggesting that these intermediate calculation tokens rely on
more concentrated contextual support. Within calculation segments, we further
observe a local rise--fall--rise pattern: attention entropy increases when a new
calculation step is initiated, decreases during the middle of a stable local
computation, and rises again when the computed result is integrated into the next
reasoning step.

Predictive entropy exhibits a different behavior. Its peaks mark positions where
the model is uncertain about the next-token choice, rather than positions where
contextual support is necessarily diffuse. In the example, the erroneous token
``e'' in the fragment ``$x - 2 = e$'' has high predictive entropy, suggesting
that prediction entropy can act as a local warning signal for unstable or
incorrect token generation. However, such prediction-entropy peaks do not
generally coincide with attention-entropy peaks, indicating that output
uncertainty and contextual-support diffuseness capture different aspects of
token-level behavior.

The sequence-level visualization is intended as a qualitative diagnostic rather
than a standalone statistical claim. Together with the global scatter and
quadrant analyses, it suggests that prediction entropy and attention entropy can
assign different roles to the same token positions: prediction entropy reflects
local output uncertainty, whereas attention entropy reflects the diffuseness of
contextual support.

\paragraph{Control conclusion.}
Overall, prediction entropy does not provide an equivalent substitute for
attention entropy. Predictive entropy measures output-layer uncertainty, whereas
attention entropy measures the internal concentration of contextual support.
Their scatter distribution, quadrant composition, and sequence-level trajectories
show only partial alignment. Moreover, prediction-entropy-based token selection
does not reproduce the stable-anchor / fragile-explorer asymmetry induced by
attention-entropy selection. This supports the interpretation that the
anchor/explorer spectrum is not merely a confidence or uncertainty effect at the
output layer, but is tied to how tokens aggregate evidence from prior context.

\subsection{Loss-Magnitude Controls}
\label{app:loss_controls}

Finally, we test whether the attention-entropy partition is simply a proxy for
token-level objective magnitude. Let $\ell_t$ denote the per-token policy-gradient
loss contribution. We rank tokens within each response by the magnitude of this
contribution and construct two matched $20\%$ token subsets:
\begin{equation}
\mathcal{T}_{\mathrm{loss\text{-}low}}^{20\%}
=
\operatorname{Bottom}_{20\%}
\left(\{|\ell_t|\}_{t=1}^{T}\right),
\qquad
\mathcal{T}_{\mathrm{loss\text{-}high}}^{20\%}
=
\operatorname{Top}_{20\%}
\left(\{|\ell_t|\}_{t=1}^{T}\right).
\end{equation}
We then repeat the same selective-training setup using these loss-magnitude
groups in place of the attention-entropy groups.

\begin{figure*}[t]
\centering
\begin{subfigure}{0.32\linewidth}
    \centering
    \includegraphics[width=\linewidth]{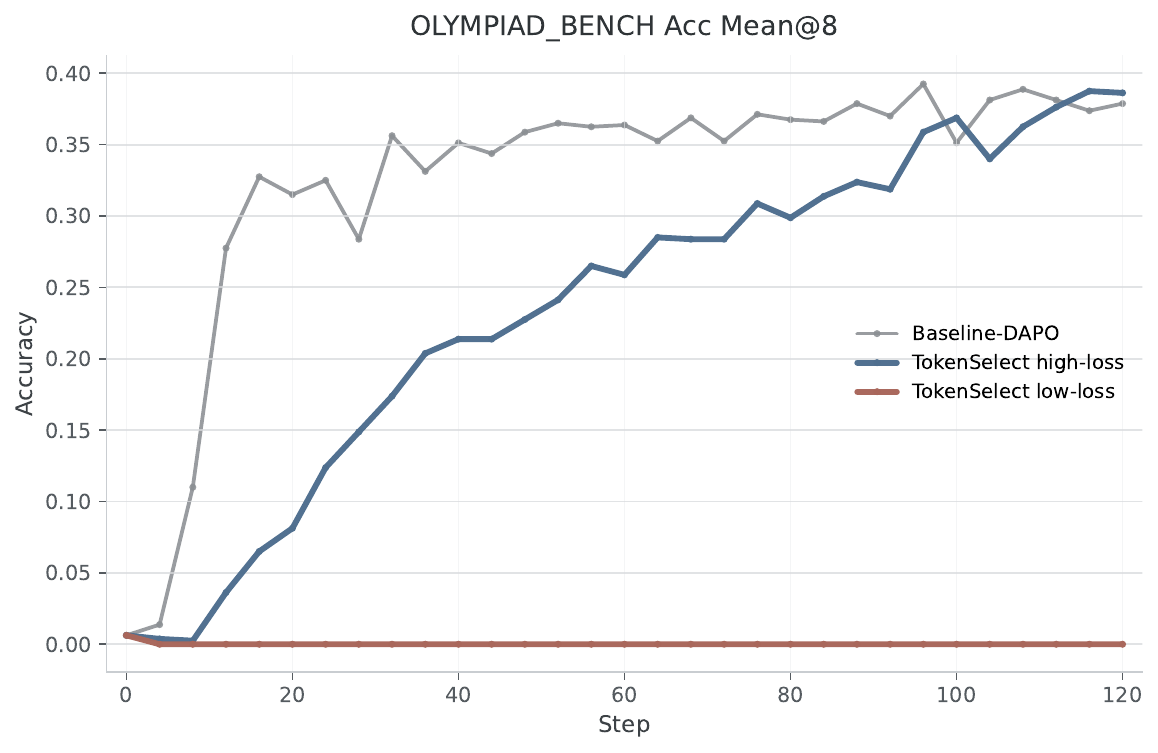}
    \caption{OlympiadBench}
\end{subfigure}
\hfill
\begin{subfigure}{0.32\linewidth}
    \centering
    \includegraphics[width=\linewidth]{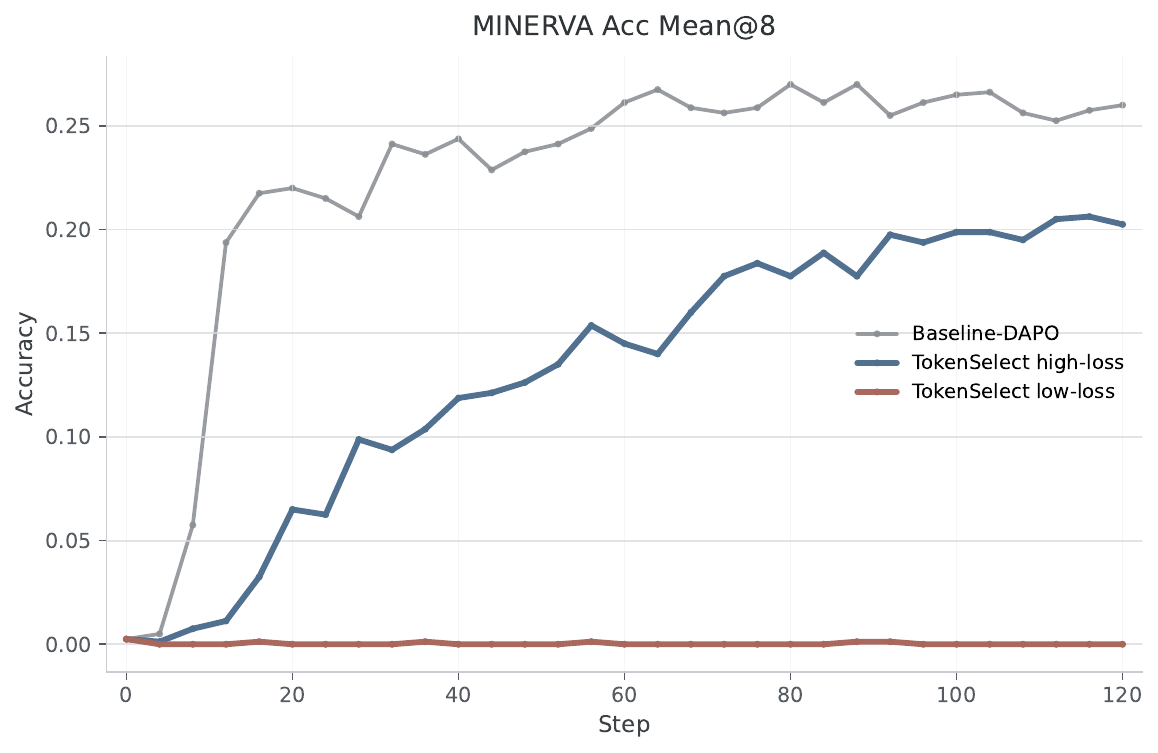}
    \caption{Minerva}
\end{subfigure}
\hfill
\begin{subfigure}{0.32\linewidth}
    \centering
    \includegraphics[width=\linewidth]{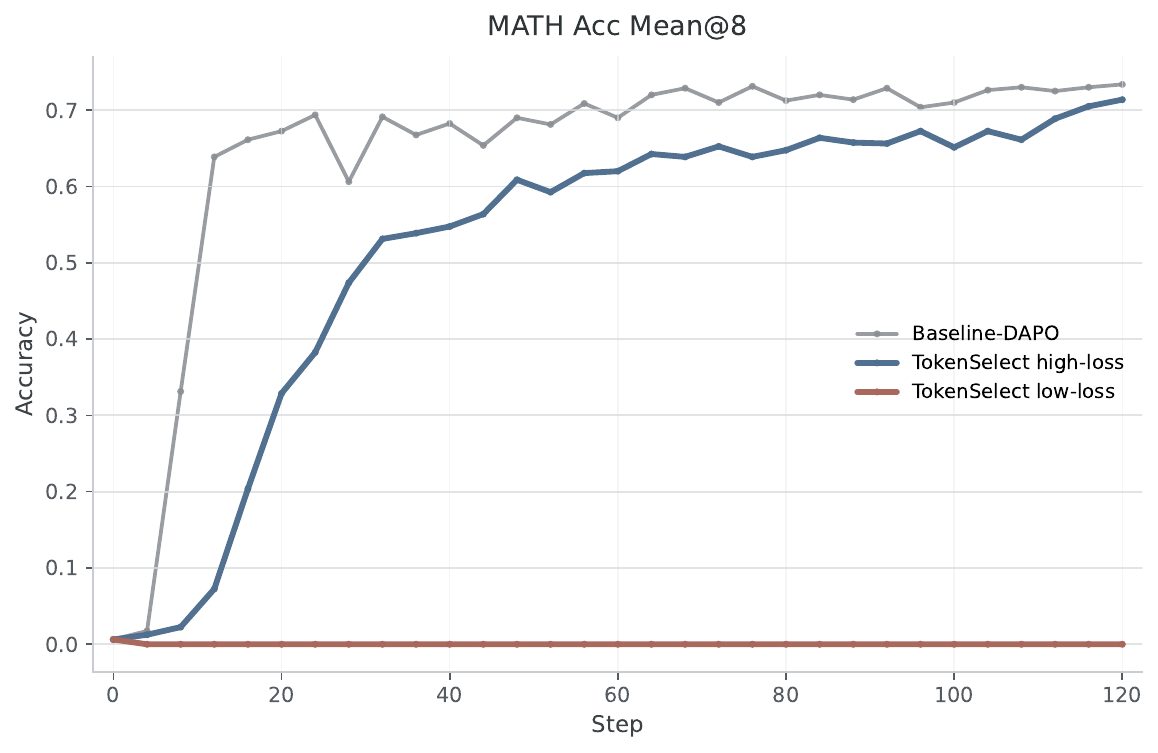}
    \caption{MATH}
\end{subfigure}

\vspace{0.6em}

\begin{subfigure}{0.32\linewidth}
    \centering
    \includegraphics[width=\linewidth]{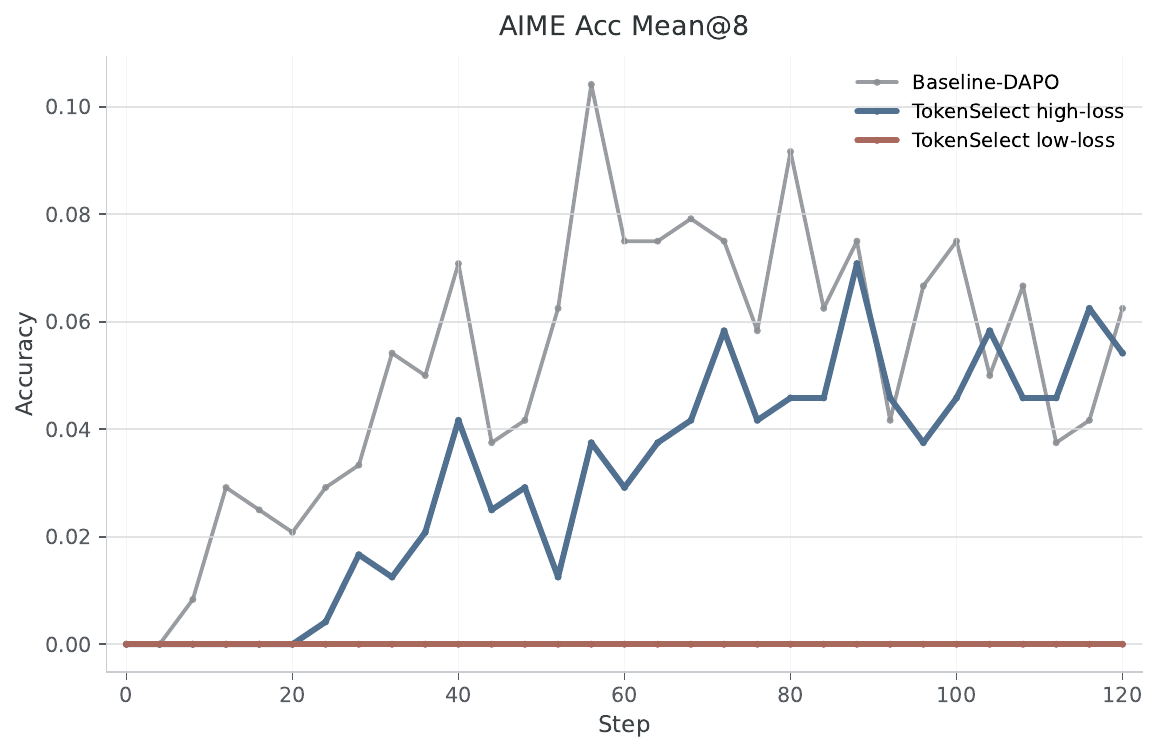}
    \caption{AIME}
\end{subfigure}
\hfill
\begin{subfigure}{0.32\linewidth}
    \centering
    \includegraphics[width=\linewidth]{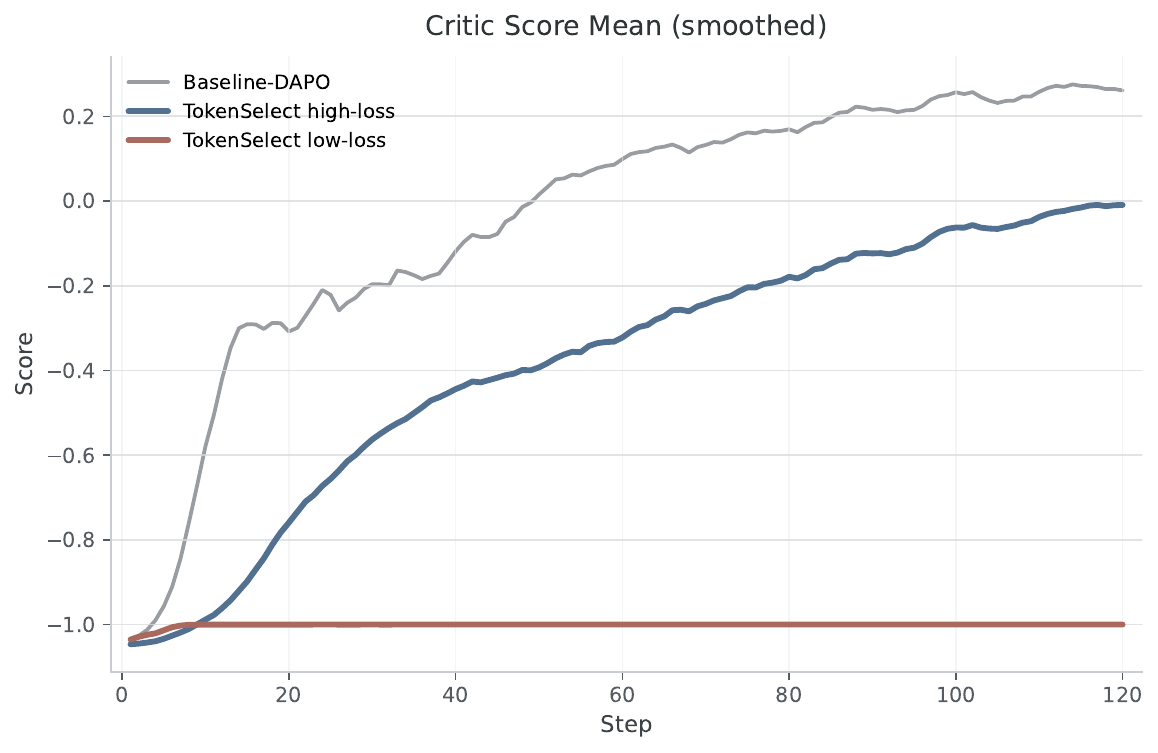}
    \caption{Critic score}
\end{subfigure}
\hfill
\begin{subfigure}{0.32\linewidth}
    \centering
    \includegraphics[width=\linewidth]{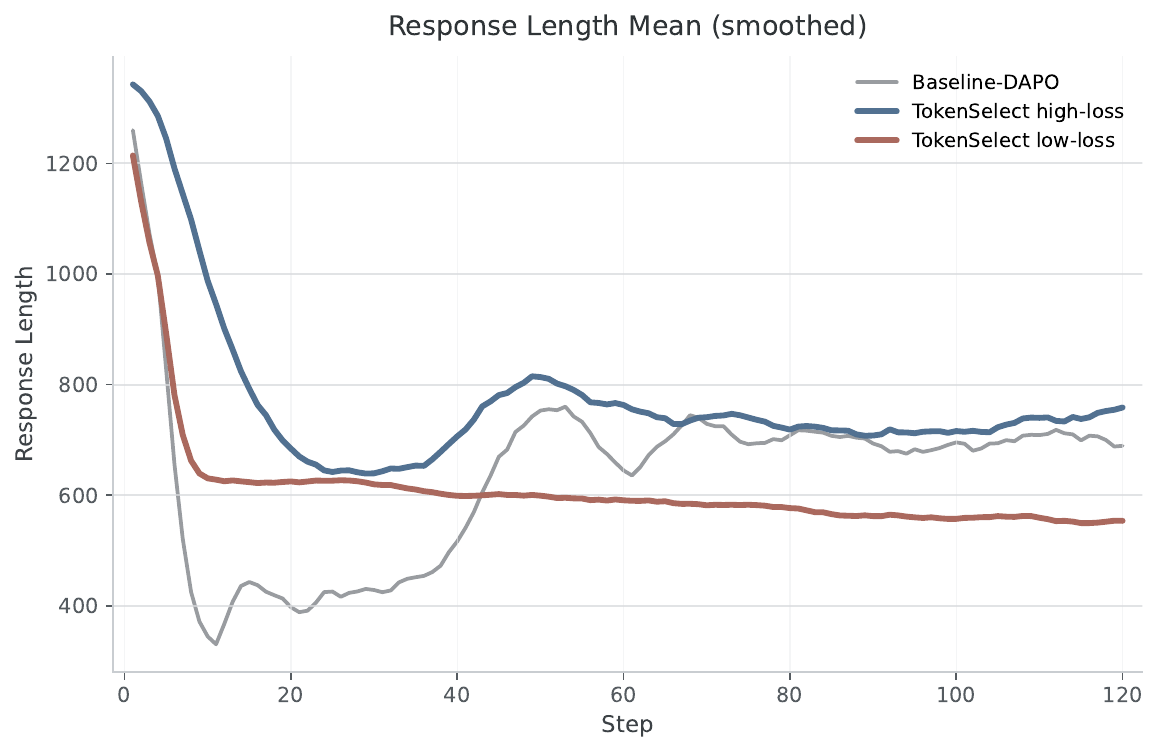}
    \caption{Response length}
\end{subfigure}

\caption{
Loss-magnitude controls for selective token training. Tokens are selected by the
magnitude of their per-token policy-gradient loss contribution rather than by
attention entropy. High-loss-only training shows gradual improvement, whereas
low-loss-only training remains close to the floor across benchmarks and critic
score.
}
\label{fig:loss_magnitude_controls}
\end{figure*}

Figure~\ref{fig:loss_magnitude_controls} shows the behavior of selective training
when tokens are grouped by instantaneous loss magnitude rather than by attention
entropy. The high-loss-only variant exhibits gradual improvement on
OlympiadBench, Minerva, MATH, AIME, and critic score, indicating that tokens with
large policy-gradient loss contributions contain direct update signal. In
contrast, the low-loss-only variant stays close to zero accuracy across all
benchmarks, while its critic score remains near the floor. This pattern is
consistent with the intended role of this control: loss-magnitude selection
primarily separates tokens by the strength of their immediate optimization
contribution.

This behavior is qualitatively different from the attention-entropy-based
selective-training curves in Figure~\ref{fig:selective_training_main}. The
entropy-based split does not simply reproduce a high-loss versus low-loss
separation. Instead, it yields a different set of training dynamics: low-entropy
and high-entropy token groups differ in stability, collapse behavior, response
length evolution, and benchmark-specific trajectories. By contrast, the
loss-magnitude split mainly distinguishes whether the selected tokens carry
large or small instantaneous policy-gradient contributions.

Therefore, these controls suggest that attention entropy is not merely acting as
a surrogate for per-token loss magnitude. Loss magnitude characterizes the
immediate strength of the token-level objective signal, whereas attention entropy
characterizes the structure of attention support used by a token when aggregating
contextual evidence. The difference between the two sets of curves indicates that
the entropy partition captures a token property that is distinct from
instantaneous loss scale.

\section{Normalization Controls and Full Optimization Trajectories}
\label{app:normalization}
\label{app:optimization_trajectories}

\subsection{Normalization variants}
\label{app:normalization_variants}

A potential confound in masked token-level training is the normalization
denominator. Let $m_t \in \{0,1\}$ denote the subset mask, $T$ the number of
response tokens, and $K=\sum_{t=1}^{T}m_t$ the number of selected tokens. For a
per-token loss contribution $\ell_t$, we compare two normalization schemes:
\begin{equation}
\mathcal{L}_{\mathrm{subset}}^{\mathrm{sel}}
=
\frac{1}{K}\sum_{t=1}^{T} m_t \ell_t ,
\label{eq:subset_selected_mean}
\end{equation}
and
\begin{equation}
\mathcal{L}_{\mathrm{subset}}^{\mathrm{all}}
=
\frac{1}{T}\sum_{t=1}^{T} m_t \ell_t .
\label{eq:subset_alltoken_mean}
\end{equation}

The selected-token mean in Eq.~\ref{eq:subset_selected_mean} preserves the
average loss scale over the retained tokens. In contrast, the all-token mean in
Eq.~\ref{eq:subset_alltoken_mean} keeps the denominator equal to the full
response length and therefore shrinks the total subset contribution. Since our
masks retain $20\%$ of response tokens, $K \approx 0.2T$, and selected-token
normalization increases the effective update scale of the retained tokens by
approximately $5\times$ relative to all-token normalization.

These two variants answer complementary questions. The all-token mean tests
whether a subset remains effective even when its contribution is diluted by the
full response length. The selected-token mean tests the intrinsic training
quality of the retained tokens after removing this dilution effect, but it also
introduces a larger effective update scale. Comparing both variants allows us to
separate genuine subset quality from a normalization-induced scale artifact.

\subsection{Full optimization trajectories}
\label{app:normalization_trajectories}

\begin{figure*}[t]
    \centering

    \begin{subfigure}[t]{0.45\linewidth}
        \centering
        \includegraphics[width=\linewidth]{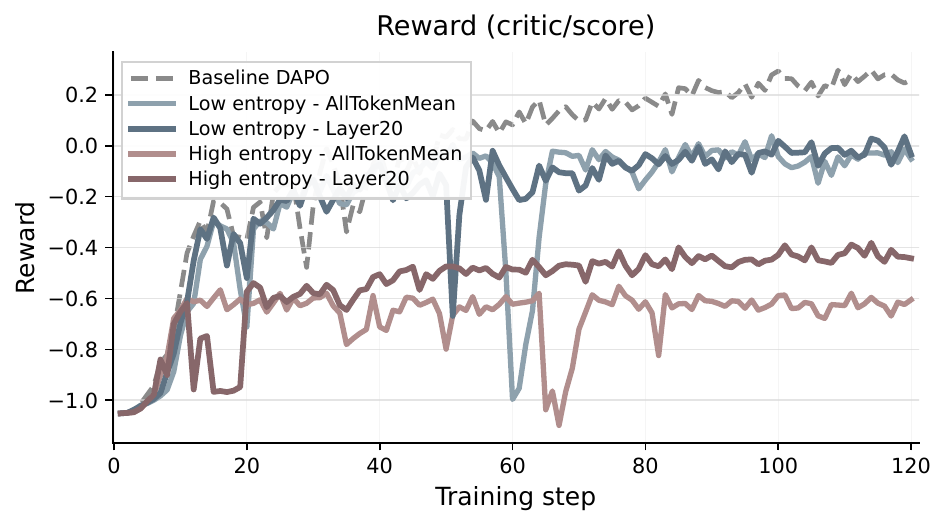}
        \caption{Training reward.}
        \label{fig:normalization_reward}
    \end{subfigure}
    \hfill
    \begin{subfigure}[t]{0.45\linewidth}
        \centering
        \includegraphics[width=\linewidth]{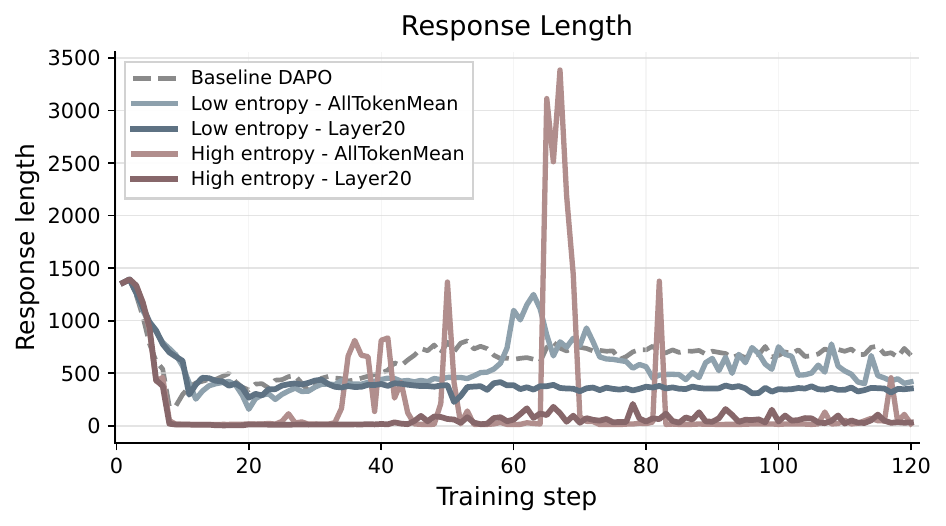}
        \caption{Response length.}
        \label{fig:normalization_response_length}
    \end{subfigure}
    
    \vspace{0.6em}
    
    \begin{subfigure}[t]{0.45\linewidth}
        \centering
        \includegraphics[width=\linewidth]{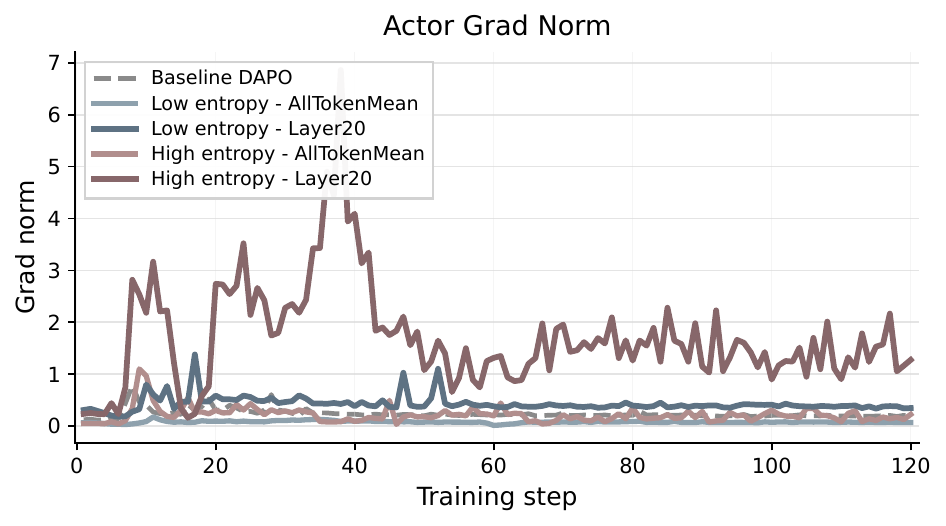}
        \caption{Actor gradient norm.}
        \label{fig:normalization_grad_norm}
    \end{subfigure}
    \hfill
    \begin{subfigure}[t]{0.45\linewidth}
        \centering
        \includegraphics[width=\linewidth]{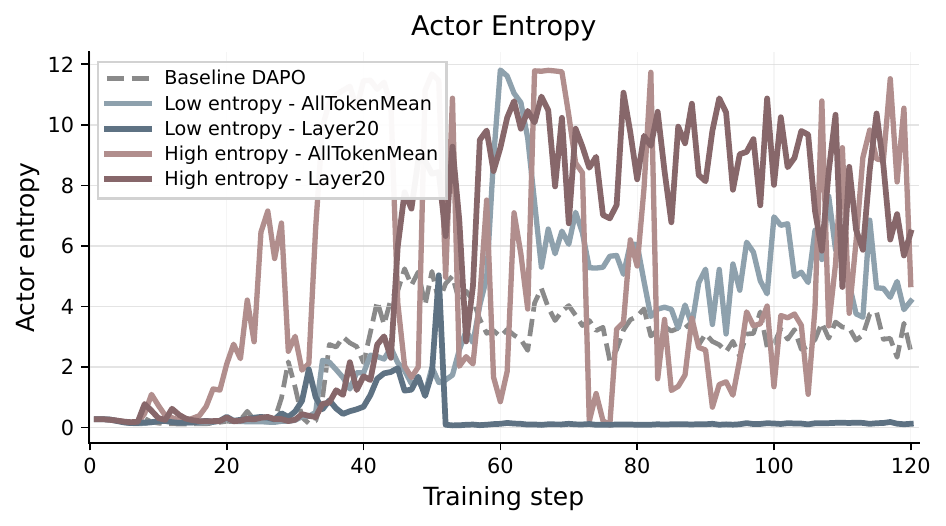}
        \caption{Policy entropy.}
        \label{fig:normalization_actor_entropy}
    \end{subfigure}

    \vspace{0.6em}
    
    \begin{subfigure}[t]{0.45\linewidth}
        \centering
        \includegraphics[width=\linewidth]{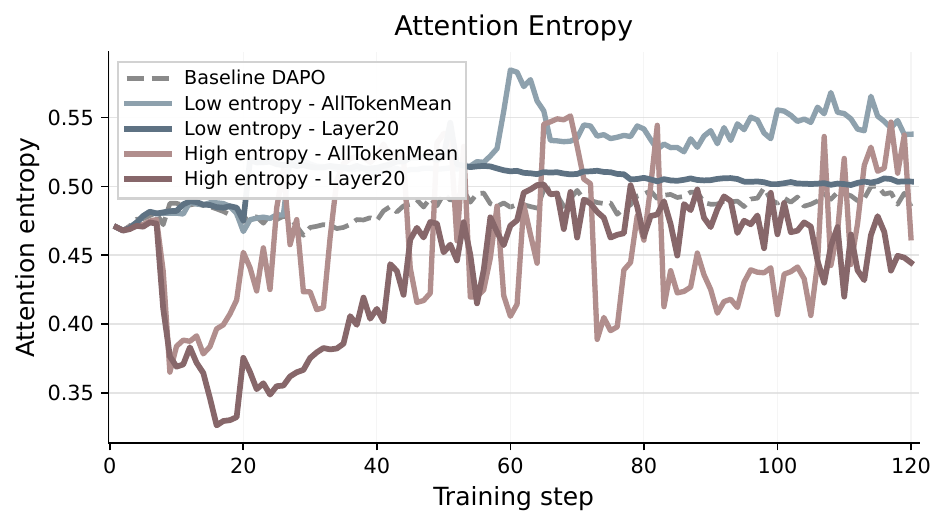}
        \caption{Attention entropy.}
        \label{fig:normalization_attention_entropy}
    \end{subfigure}
    \hfill
    \begin{subfigure}[t]{0.45\linewidth}
        \centering
        \includegraphics[width=\linewidth]{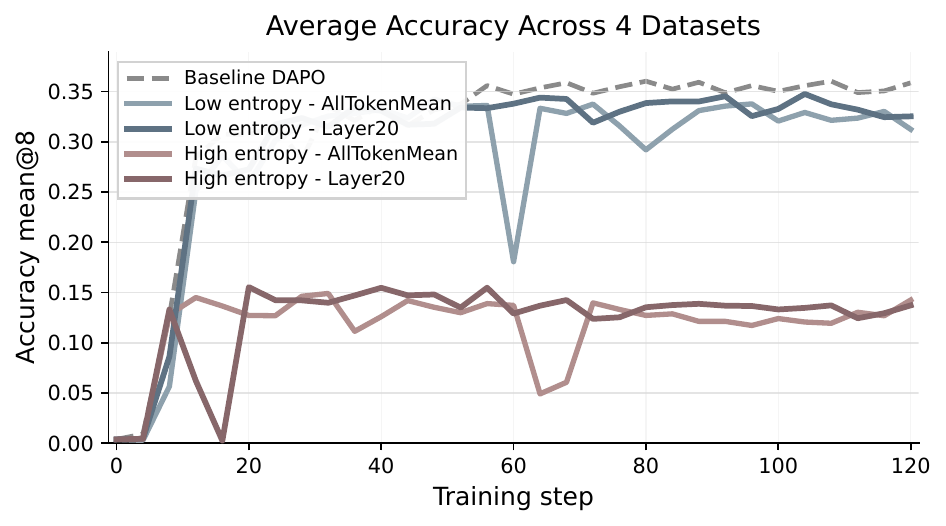}
        \caption{Average validation score.}
        \label{fig:normalization_avg_score}
    \end{subfigure}

    \caption{
    Normalization controls for low-attention-entropy and
    high-attention-entropy $20\%$ token subsets on Qwen3-8B-Base. Curves with
    the suffix \texttt{AllTokenMean} use all-token normalization
    $\mathcal{L}_{\mathrm{subset}}^{\mathrm{all}}$, while curves without this
    suffix use selected-token normalization
    $\mathcal{L}_{\mathrm{subset}}^{\mathrm{sel}}$. Low-attn training remains
    stable and competitive under both normalization schemes. High-attn training
    is weak under all-token normalization and becomes much more volatile under
    selected-token normalization, with large gradient spikes and erratic
    response-length behavior. These results indicate that the anchor/explorer
    distinction is not caused by the normalization denominator.
    }
    \label{fig:normalization_trajectories}
\end{figure*}

Figure~\ref{fig:normalization_trajectories} compares full-token DAPO with
low-attention-entropy and high-attention-entropy $20\%$ token subsets under both
normalization variants. Runs with the suffix \texttt{AllTokenMean} use
Eq.~\ref{eq:subset_alltoken_mean}; runs without this suffix use
Eq.~\ref{eq:subset_selected_mean}.

The main observation is that the low-vs-high attention-entropy gap persists
under both normalization schemes. Low-attention-entropy training remains strong
and stable under both selected-token and all-token normalization. It achieves
held-out performance close to full-token DAPO on the averaged validation score,
maintains relatively stable response lengths, and does not exhibit large
gradient spikes. This indicates that low-entropy anchor tokens provide robust
optimization signals rather than benefiting from a particular normalization
choice.

High-attention-entropy training behaves differently. Under all-token
normalization, high-attn training is weak: the subset contribution is diluted,
the reward remains low, response lengths often collapse to short traces, and
held-out accuracy stays far below both low-attn training and full-token DAPO.
However, under selected-token normalization, simply restoring the per-token
scale does not make high-attn training stable. Instead, high-attn training
exhibits much larger gradient norms, abrupt response-length spikes, and unstable
reward dynamics. Thus, high-attn tokens do not fail merely because their
gradients are too small under all-token normalization. When their update scale
is restored, their signals become stronger but not more reliable.

This comparison supports the interpretation that low-entropy anchors and
high-entropy explorers play different optimization roles. Low-entropy anchors
provide concentrated-support updates that are stable across normalization
choices. High-entropy explorers carry stronger and more volatile signals that
are difficult to use in isolation. Therefore, the anchor/explorer distinction is
not an artifact of the loss normalization denominator.

\subsection{Characteristic failure modes of explorer-only training}
\label{app:explorer_failure_modes}

The normalization comparison also clarifies the failure modes of explorer-only
training. High-attention-entropy tokens are not simply uninformative. Under
selected-token normalization, their gradient norms can become much larger than
those of full-token DAPO or low-attn training, indicating that these tokens
indeed carry strong optimization pressure. However, this pressure is difficult
to use in isolation and often manifests as unstable training dynamics.

We observe three recurring failure modes. First, some runs enter an early
short-response collapse, where response length rapidly shrinks and reward
growth stalls. Second, some runs show abrupt length instability, with sudden
spikes in generated length followed by reward degradation. Third, some runs
undergo gradual reasoning degeneration, where validation accuracy initially
improves but later declines as the policy drifts toward less reliable reasoning
patterns. These behaviors are consistent with the view that high-entropy
explorer tokens aggregate diffuse evidence and produce high-variance updates.

Rare successful explorer-only runs avoid these failure modes. They maintain
longer reasoning traces, recover reward growth, and can sometimes achieve strong
performance on difficult benchmarks such as AIME. We interpret this as evidence
of potential signal rather than method competitiveness: explorer tokens are not
simply uninformative, but their useful contribution is difficult to extract when
they are isolated from lower-entropy stabilizing updates. This motivates using a
stable optimization backbone or a softer weighting strategy rather than relying
on isolated hard masking.

\begin{table}[t]
\centering
\caption{
Explorer-only multi-seed statistics over $N=8$ independent runs.
A run is considered successful if it avoids collapse according to the criterion
defined in this appendix. Because most explorer-only runs collapse, the
successful-run statistics are reported only as a conditional diagnostic of
potential signal quality, not as an overall method comparison. Avg. w/o AIME
denotes the arithmetic mean over OlympiadBench, Minerva, and MATH.
}
\label{tab:explorer_stats}
\begin{tabular}{lc}
\toprule
Metric & Value \\
\midrule
Total runs & 8 \\
Successful runs & 3 \\
Success rate & 37.5\% \\
\midrule
\multicolumn{2}{l}{\textit{Successful runs}} \\
\quad AIME (mean $\pm$ std) & $12.31 \pm 0.53$ \\
\quad Avg. w/o AIME (mean $\pm$ std) & $46.13 \pm 1.58$ \\
\quad AIME $-$ full-token & $6.48$ \\
\midrule
\multicolumn{2}{l}{\textit{Failed runs}} \\
\quad First detected collapse step & $21.80 \pm 3.84$ \\
\bottomrule
\end{tabular}
\end{table}

\begin{figure*}[t]
    \centering
    \begin{subfigure}[t]{0.45\linewidth}
        \centering
        \includegraphics[width=\linewidth]{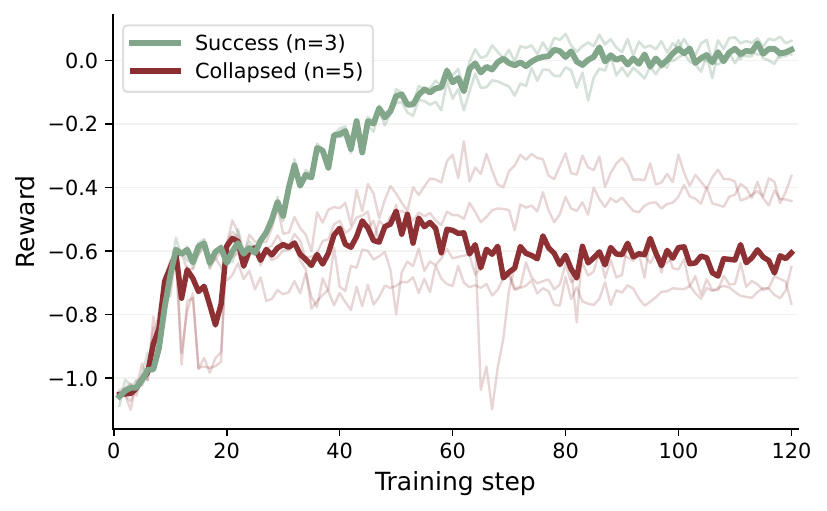}
        \caption{Training reward.}
        \label{fig:explorer_reward_bimodal}
    \end{subfigure}
    \hfill
    \begin{subfigure}[t]{0.45\linewidth}
        \centering
        \includegraphics[width=\linewidth]{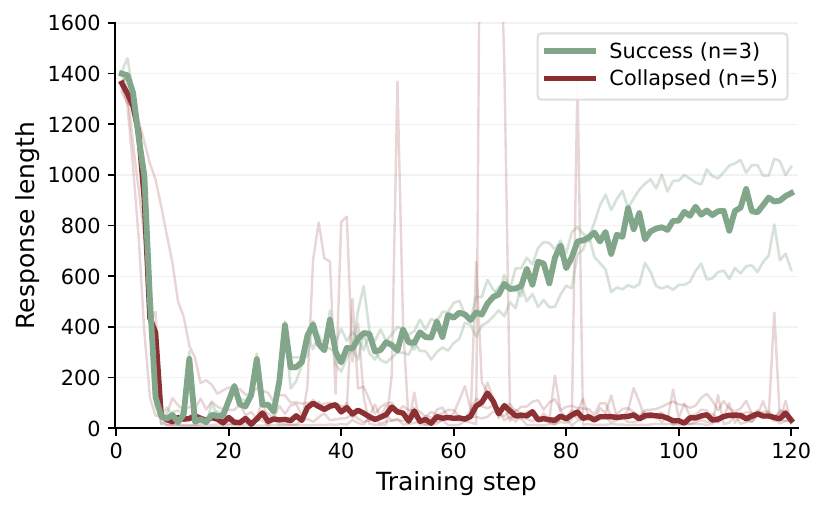}
        \caption{Response length.}
        \label{fig:explorer_length_bimodal}
    \end{subfigure}

    \caption{
    Explorer-only training exhibits a bimodal optimization pattern across
    $N=8$ independent seeds. Successful runs (green, $n=3$) recover normal
    reward growth and maintain long reasoning traces, while collapsed runs
    (red, $n=5$) fail to sustain learning and rapidly degenerate into very
    short responses. Thick curves denote group averages, and faint curves
    denote individual runs.
    }
    \label{fig:explorer_bimodal}
\end{figure*}

\subsection{Interpretation}
\label{app:normalization_interpretation}

Overall, normalization modulates the severity of subset training but does not
create or remove the entropy-defined optimization spectrum. All-token
normalization is conservative and can dilute sparse updates, while
selected-token normalization restores the per-token loss scale but can amplify
unstable signals. Low-attn anchors remain robust under both choices, whereas
high-attn explorers are either too weak under all-token normalization or too
volatile under selected-token normalization.

This supports the central interpretation of our analysis: attention entropy
does not merely select tokens with a favorable normalization scale. Instead, it
separates response tokens along a structural optimization axis. Low-entropy
tokens provide stable concentrated-support updates that can serve as an
optimization backbone. High-entropy tokens provide broader, stronger, but
higher-variance signals that are difficult to exploit through hard masking
alone. This motivates the entropy-aware soft-reweighting validation
intervention, where stable anchor-like updates are emphasized early while
broader explorer-like signals are gradually incorporated later in training.




\section{Additional Evidence-Gathering Analyses}
\label{app:evidence_gathering}

\subsection{Robustness across entropy definitions}
\label{app:entropy_variant_evidence_gathering}

In the main paper, we use normalized attention entropy as the default score for defining anchor and explorer tokens. A potential concern is that different entropy definitions may capture different aspects of attention behavior. In particular, normalized entropy divides raw entropy by $\log N_t$, where $N_t$ is the number of visible positions, and therefore measures concentration relative to the available context length. Raw entropy, in contrast, has an upper bound that grows with $N_t$ and can therefore be affected by token position and visible-context length. To examine whether our evidence-gathering conclusions are specific to one entropy definition, we repeat the analysis under four variants: normalized attention entropy, raw attention entropy, raw attention entropy restricted to the first 512 response tokens, and top-256 raw attention entropy.

Across all variants, the support-concentration pattern is consistent: anchor tokens require fewer support positions to accumulate the same amount of attention mass, whereas explorer tokens require more support positions. This confirms that the anchor--explorer distinction is robustly tied to effective support size. In contrast, spatial-span statistics are less stable across entropy definitions. Under normalized entropy, anchors can appear slightly more non-local, reflecting sparse support relative to a long visible context. Under raw entropy, explorers appear more non-local, partly because raw entropy is affected by the growth of the visible context. Restricting the analysis to the first 512 response tokens reduces this position-length effect, while top-256 raw entropy further focuses on the dominant attention support and weakens the distance-based separation. These results support the interpretation in Section~\ref{sec:evidence_gathering}: attention entropy should primarily be viewed as a measure of support concentration rather than attention distance.

\begin{figure*}[t]
    \centering
    \includegraphics[width=0.98\textwidth]{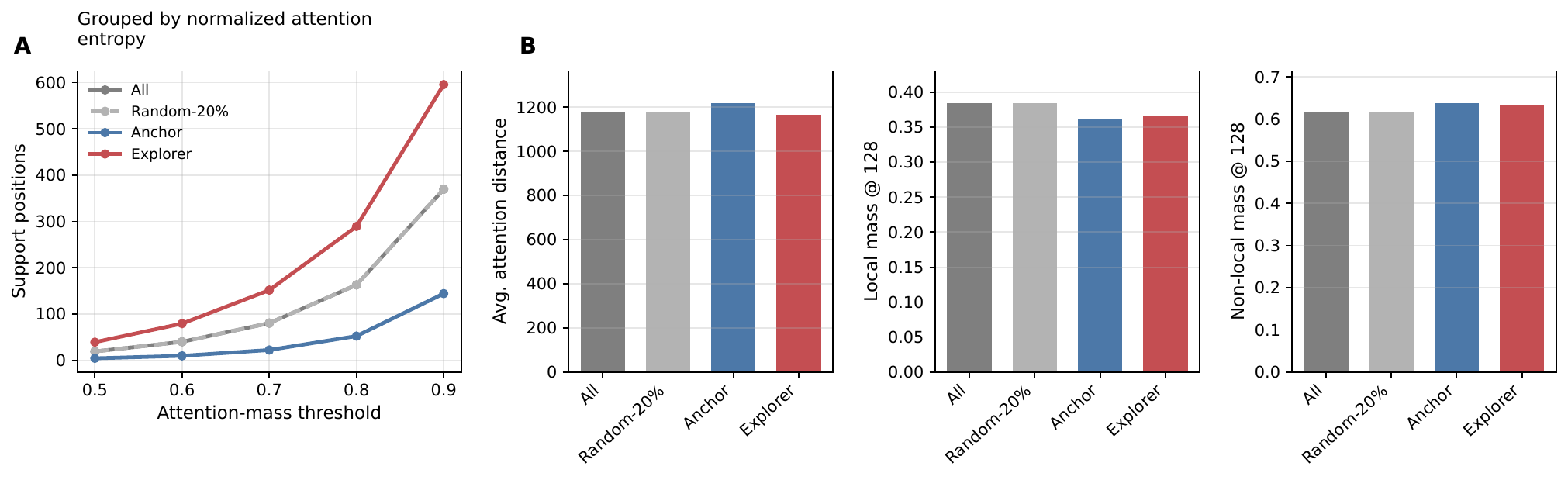}
    \caption{
    Evidence-gathering patterns grouped by normalized attention entropy.
    (A) Anchor tokens require far fewer support positions than explorer tokens to accumulate the same attention mass, indicating concentrated support.
    Explorer tokens require many more support positions, indicating diffuse aggregation.
    (B) Under normalized entropy, anchors are not more local: they have slightly larger average attention distance, lower local-window mass, and higher non-local mass.
    This suggests that normalized entropy captures relative support concentration rather than attention locality.
    }
    \label{fig:app_evidence_gathering_entropy_norm}
\end{figure*}

\paragraph{Normalized attention entropy.}
Figure~\ref{fig:app_evidence_gathering_entropy_norm} shows the evidence-gathering pattern under the default normalized attention entropy used in the main paper. Anchor tokens require substantially fewer support positions than explorer tokens at all attention-mass thresholds, confirming the sparse-versus-diffuse support distinction. Spatially, anchors are not more local under this definition; they show slightly larger average attention distance, lower local mass, and higher non-local mass. This indicates that normalized entropy captures relative support concentration rather than locality.

\begin{figure*}[t]
    \centering
    \includegraphics[width=0.98\textwidth]{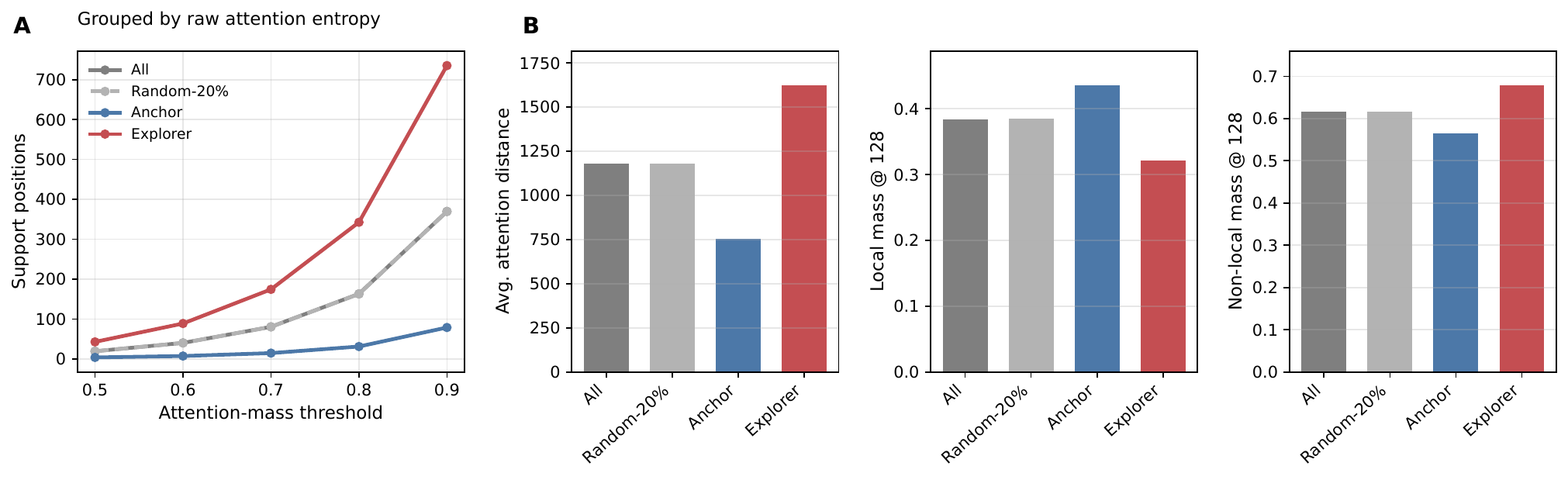}
    \caption{
    Evidence-gathering patterns grouped by raw attention entropy.
    (A) The sparse-versus-diffuse support distinction remains clear: anchors require fewer support positions, while explorers require more.
    (B) In contrast to normalized entropy, raw entropy makes explorers appear more non-local, with larger average attention distance and higher non-local mass.
    Since the upper bound of raw entropy increases with the number of visible positions, this spatial pattern may partly reflect position and visible-context-length effects.
    }
    \label{fig:app_evidence_gathering_entropy_raw}
\end{figure*}

\paragraph{Raw attention entropy.}
Figure~\ref{fig:app_evidence_gathering_entropy_raw} repeats the analysis using raw attention entropy. The concentration pattern remains unchanged: anchors require fewer support positions and explorers require more. However, the spatial-span pattern differs from normalized entropy. Under raw entropy, explorer tokens show much larger average attention distance and higher non-local mass, while anchors appear more local. This difference should be interpreted cautiously because the upper bound of raw entropy grows with the number of visible positions. Therefore, raw entropy can partially reflect token position and visible-context length.

\begin{figure*}[t]
    \centering
    \includegraphics[width=0.98\textwidth]{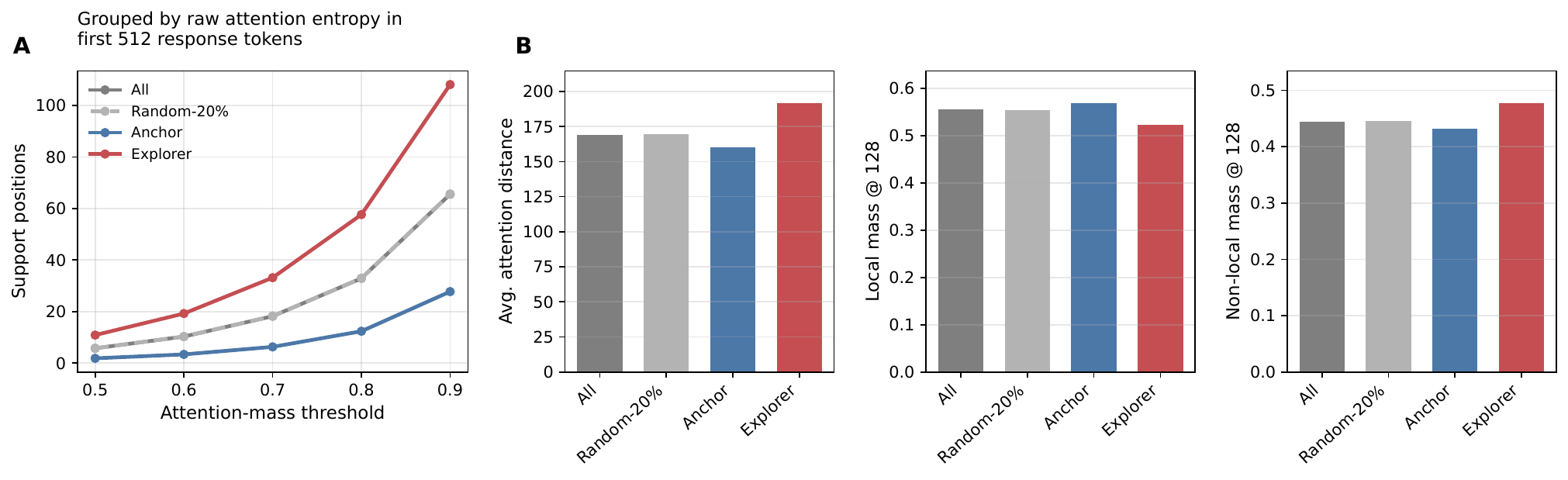}
    \caption{
    Evidence-gathering patterns grouped by raw attention entropy within the first 512 response tokens.
    (A) Anchor tokens still require fewer support positions than explorer tokens, showing that the concentration-based distinction persists when restricting the analysis to early response tokens.
    (B) The distance-based separation is weaker than in the full raw-entropy analysis, suggesting that visible-context-length and position effects contribute to the strong spatial-span pattern observed under raw entropy.
    }
    \label{fig:app_evidence_gathering_entropy_raw_first512}
\end{figure*}

\paragraph{Raw attention entropy in the first 512 response tokens.}
To reduce the visible-context-length confound in raw entropy, we repeat the raw-entropy analysis using only the first 512 response tokens. Figure~\ref{fig:app_evidence_gathering_entropy_raw_first512} shows that the concentration pattern again remains stable: anchors require fewer support positions, while explorers require more. The spatial-span gap becomes much smaller than in the full raw-entropy analysis. Explorers still show somewhat larger average attention distance and higher non-local mass than anchors, but the difference is substantially reduced. This suggests that part of the strong non-local pattern under raw entropy comes from position and context-length effects.

\begin{figure*}[t]
    \centering
    \includegraphics[width=0.98\textwidth]{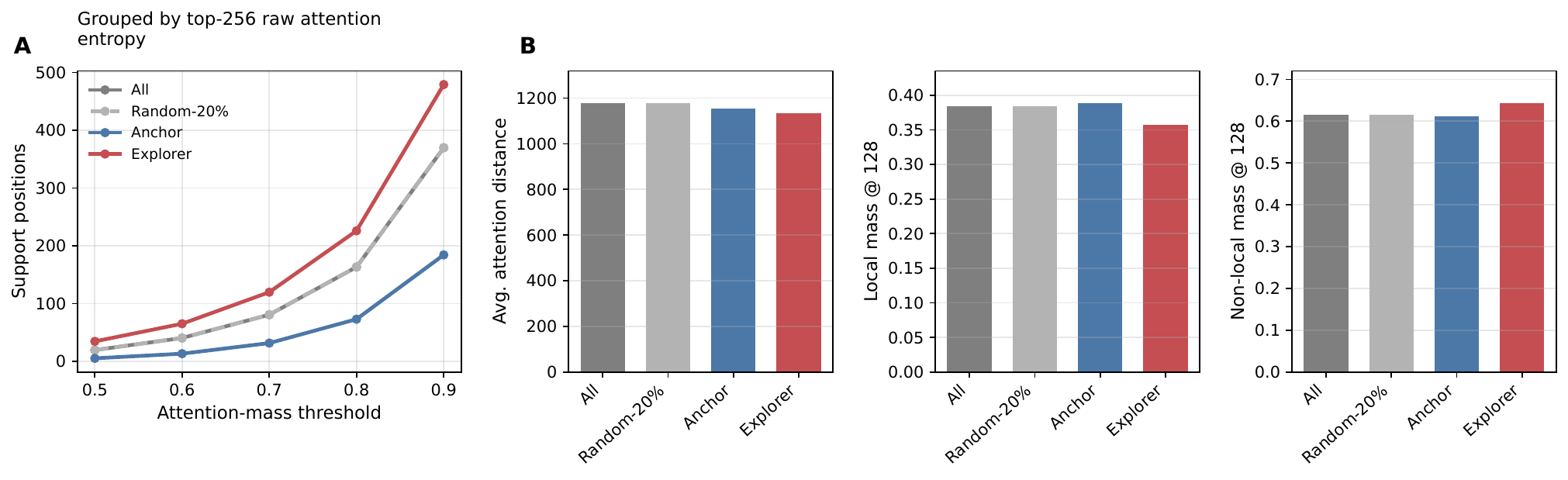}
    \caption{
    Evidence-gathering patterns grouped by top-256 raw attention entropy.
    (A) The support-concentration distinction remains robust: anchors require fewer support positions and explorers require more.
    (B) Spatial-span differences are weaker under top-256 entropy, indicating that the main stable distinction between anchors and explorers is effective support size rather than attention distance.
    }
    \label{fig:app_evidence_gathering_entropy_topk_256}
\end{figure*}

\paragraph{Top-256 raw attention entropy.}
Finally, we compute entropy using only the top-256 attention weights, after renormalizing them. This focuses the entropy measure on the dominant support positions and reduces the influence of long-tail attention mass. Figure~\ref{fig:app_evidence_gathering_entropy_topk_256} again shows the same concentration pattern: anchors require fewer support positions and explorers require more. However, distance-based differences are relatively weak. This further supports the conclusion that entropy-defined token groups are robustly separated by effective support size, while spatial locality is not the primary or stable axis of separation.

\paragraph{Summary.}
Taken together, these variants show that the anchor--explorer distinction is robust along the support-concentration axis but not along a fixed locality axis. Across normalized entropy, raw entropy, first-512 raw entropy, and top-256 raw entropy, anchors consistently use fewer effective support positions, whereas explorers aggregate over more positions. However, whether anchors or explorers appear more non-local depends on the entropy definition. We therefore avoid interpreting attention entropy as a direct measure of attention distance. Instead, we interpret it as a diagnostic for effective support concentration, which is the property most consistently associated with the observed optimization spectrum.

\subsection{Spatial-span diagnostics are secondary to support concentration}
\label{app:spatial_span_secondary}

The figures above also provide additional spatial-span diagnostics, including
average attention distance, local-window mass, and non-local mass. These
statistics are useful for checking whether attention entropy can be reduced to a
simple local-versus-global distinction. The answer is negative: spatial-span
patterns vary substantially across entropy definitions.

Under normalized attention entropy
(Figure~\ref{fig:app_evidence_gathering_entropy_norm}), anchor tokens have
slightly larger average attention distance, lower local mass, and higher
non-local mass than explorer tokens. This means that low normalized entropy
does not imply local attention. A token can have low normalized entropy by
placing highly concentrated attention on a small number of salient positions,
even if those positions are distant in the context.

Under raw attention entropy
(Figure~\ref{fig:app_evidence_gathering_entropy_raw}), the ordering reverses:
explorer tokens appear much more non-local. This is expected because raw
entropy grows with the number of visible positions and is therefore more
sensitive to response position and context length. When we restrict the raw
analysis to the first 512 response tokens
(Figure~\ref{fig:app_evidence_gathering_entropy_raw_first512}), this
distance-based gap becomes much smaller, indicating that part of the raw-entropy
spatial pattern comes from visible-context-length effects rather than from an
intrinsic locality property of the token group. When entropy is computed only
over the top-256 attention weights
(Figure~\ref{fig:app_evidence_gathering_entropy_topk_256}), distance-based
differences are further weakened.

These results reinforce the main interpretation in
Section~\ref{sec:evidence_gathering}. Attention entropy should not be treated
as a direct locality score. Its most stable diagnostic role is to distinguish
tokens with sparse selective support from tokens with diffuse multi-position
aggregation. Spatial-span statistics help rule out an overly simple
local-versus-global interpretation, but they are not the defining axis of the
anchor--explorer distinction.

\subsection{Qualitative attention-map case studies}
\label{app:qualitative_attention_maps}

We further provide qualitative attention-map examples to illustrate the
evidence-gathering modes captured by attention entropy. These examples are not
intended as standalone causal evidence; rather, they visually complement the
quantitative support-concentration analyses in
Appendix~\ref{app:entropy_variant_evidence_gathering} and
Appendix~\ref{app:spatial_span_secondary}.

\begin{figure*}[t]
    \centering

    \begin{minipage}[t]{0.98\textwidth}
        \centering
        \includegraphics[width=\linewidth]{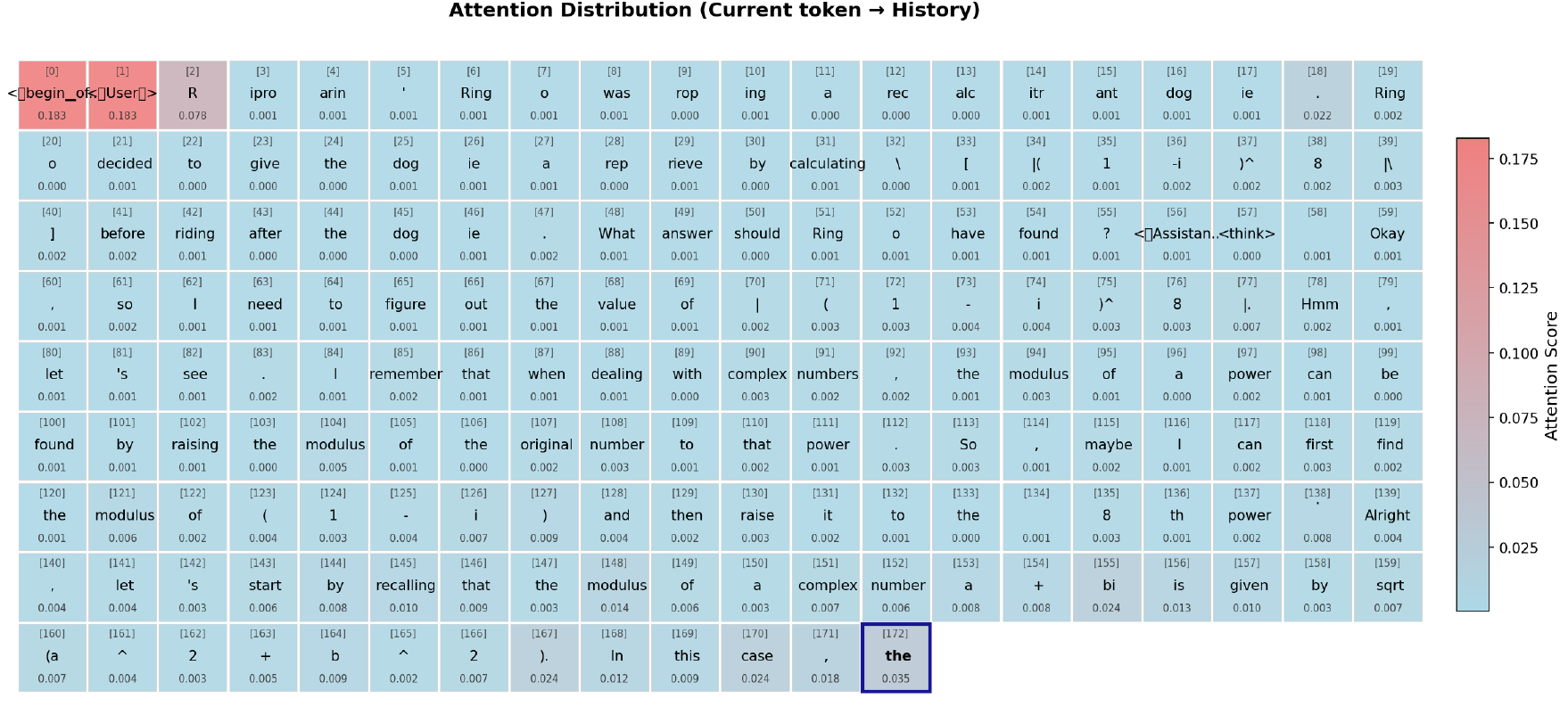}
        \vspace{1mm}

        \small
        \textbf{(a) Anchor token.}
        Low-entropy token from the same reasoning trajectory.
        The attention mass is concentrated on a small set of salient support
        positions.
    \end{minipage}
    \begin{minipage}[t]{0.98\textwidth}
        \centering
        \includegraphics[width=\linewidth]{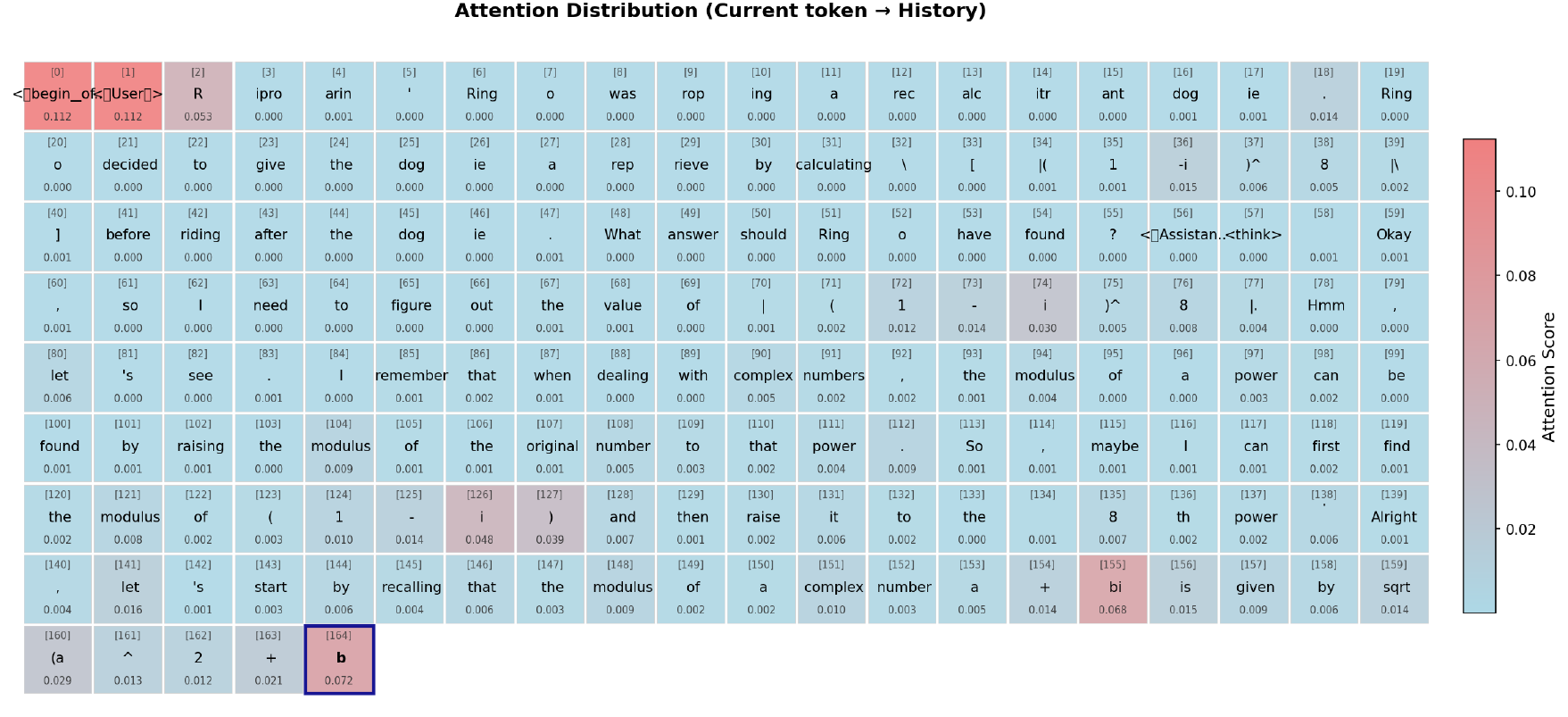}
        \vspace{1mm}

        \small
        \textbf{(b) Explorer token.}
        High-entropy token from the same reasoning trajectory.
        The attention mass is distributed across many non-adjacent support
        positions.
    \end{minipage}

    \caption{
    Qualitative attention-map case study from a single reasoning trajectory.
    We compare one anchor token and one explorer token selected by
    within-response normalized attention entropy. Each panel visualizes the
    attention distribution from the selected current token to its history tokens.
    The anchor token exhibits sparse selective support, concentrating attention
    on a small number of salient positions. The explorer token exhibits diffuse
    multi-position aggregation, spreading attention across multiple separated
    reasoning states. These examples illustrate that the anchor--explorer
    distinction is better understood as sparse versus diffuse support, rather
    than as a simple local-versus-global attention pattern.
    }
    \label{fig:app_qualitative_attention_maps}
\end{figure*}

Figure~\ref{fig:app_qualitative_attention_maps} shows two selected tokens from
the same reasoning trajectory: one anchor token from the bottom 20\% of
within-response normalized attention entropy, and one explorer token from the
top 20\%. For each selected token, we visualize its attention distribution over
previous tokens. The cells are ordered by their original positions in the
trajectory, and the color intensity indicates the attention mass assigned by
the current token to each history token.

The anchor example illustrates sparse selective support. Its attention mass is
concentrated on a small number of salient positions in the reasoning chain,
which may include nearby tokens or distant earlier states. In contrast, the
explorer example illustrates diffuse multi-position aggregation: its attention
is spread over multiple non-adjacent positions, such as problem quantities,
operators, intermediate results, and previously derived expressions. This
qualitative pattern is consistent with the quantitative observation that
anchor tokens require fewer support positions to accumulate a fixed amount of
attention mass, whereas explorer tokens require more.

\paragraph{Interpretation: evidence-gathering modes and reasoning operations.}
The evidence-gathering analyses above suggest an operational interpretation of
the anchor--explorer distinction. Anchor tokens correspond to reasoning
operations that can be supported by a small number of salient positions, such as
continuing along an established derivation, consolidating an intermediate
quantity, or confirming a previously derived result. These support positions may
be nearby or distant, depending on the entropy definition and the specific
trajectory, but the effective support remains selective. Such operations are
frequent and tend to produce reusable, low-variance optimization signals.

Explorer tokens, in contrast, correspond to operations that require diffuse
multi-position aggregation, such as combining several intermediate results,
checking consistency across distant reasoning steps, or integrating multiple
partial states before making a new inference. These operations are less
frequent and more trajectory-dependent. This interpretation helps explain the
optimization asymmetry observed in the main experiments: anchor-only training is
stable because it preserves a reliable selective-support backbone, whereas
explorer-only training is fragile because it isolates broader but more
context-sensitive signals. At the same time, the complementary information
carried by explorer tokens can be useful in successful runs, especially on
harder reasoning benchmarks.

\section{Additional Entropy Dynamics}
\label{app:additional_entropy_dynamics}

In the main paper, we report entropy dynamics using normalized attention entropy, which is the default metric used to define anchor and explorer tokens. Here we provide additional entropy-dynamics analyses under alternative entropy definitions: raw attention entropy, top-$k$ attention entropy, and fixed-position entropy. These variants serve two purposes. First, they test whether the mean separation between anchor, full, and explorer tokens is specific to normalized entropy. Second, they help separate support-concentration effects from normalization and visible-context-length effects.

Across these variants, the most robust observation is the persistence of mean entropy separation: explorer tokens remain the highest-entropy group, anchor tokens remain the lowest-entropy group, and full tokens lie in between. This confirms that RL training does not collapse all response tokens into a uniform attention regime. Instead, the separation between sparse-support and diffuse-support token groups is preserved throughout training.

At the same time, the within-group standard deviation should be interpreted carefully. It measures dispersion of scalar entropy values within a group, not optimization volatility. In several settings, explorer tokens form a relatively narrow high-entropy band and therefore have smaller entropy-value variance than full or anchor tokens. This does not imply that explorer-only training is optimization-stable. Rather, explorer instability is captured by selective-training outcomes and gradient-level alignment dynamics, not by the scalar entropy variance itself.

\begin{figure*}[t]
    \centering
    \includegraphics[width=0.95\textwidth]{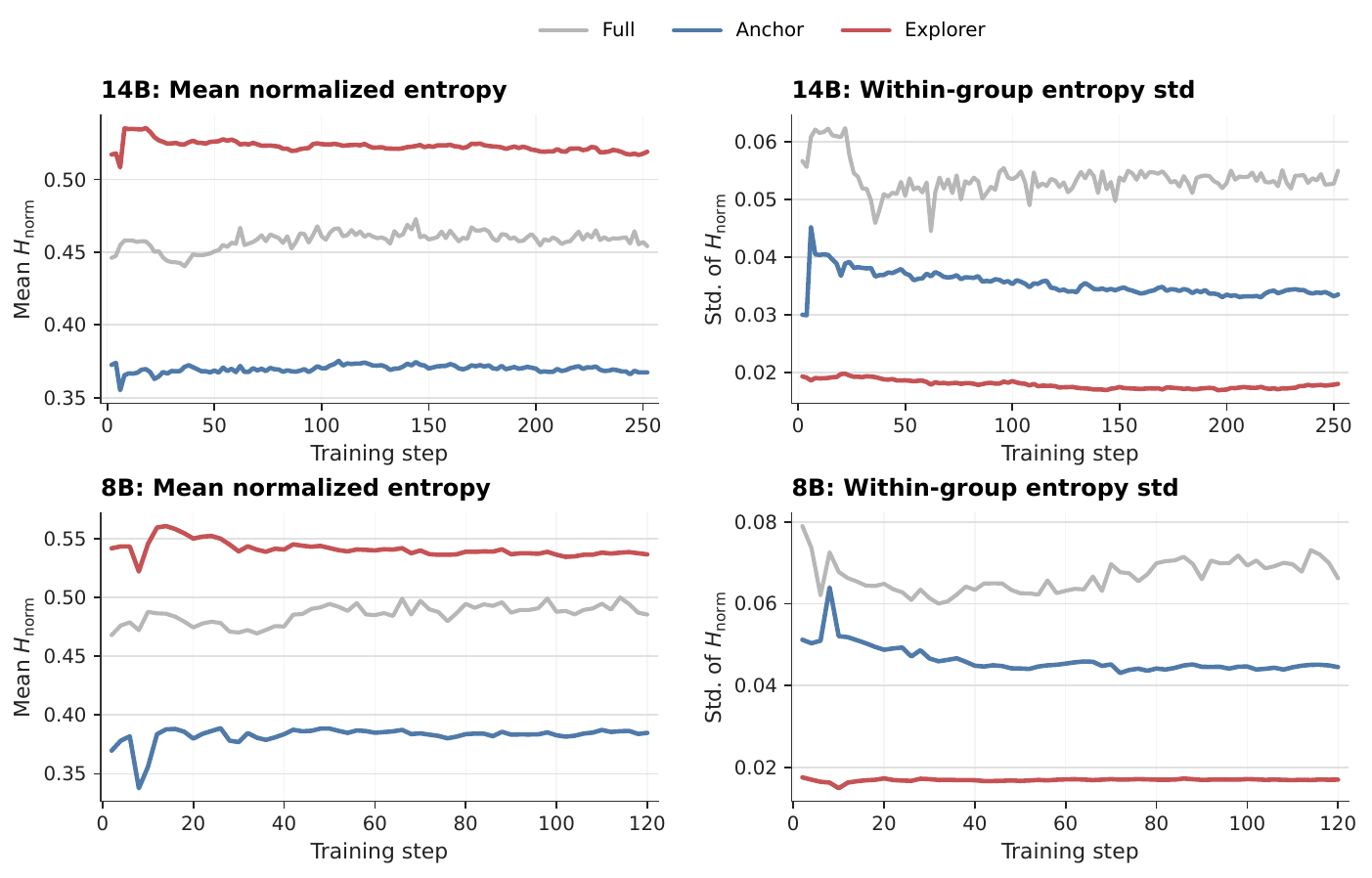}
    \caption{
    Dynamics of normalized attention entropy during RL training.
    We report the mean normalized entropy and the within-group entropy standard deviation for full response tokens, anchor tokens, and explorer tokens on Qwen3-14B and Qwen3-8B.
    Across both model scales, the mean entropy remains clearly separated throughout training: explorer tokens maintain the highest normalized entropy, anchor tokens maintain the lowest, and full tokens lie in between.
    The standard deviation measures dispersion of scalar entropy values within each group, not optimization volatility.
    }
    \label{fig:entropy_dynamics}
\end{figure*}

\subsection{Raw entropy dynamics}
\label{app:raw_entropy_dynamics}

\begin{figure*}[t]
    \centering
    \includegraphics[width=0.95\textwidth]{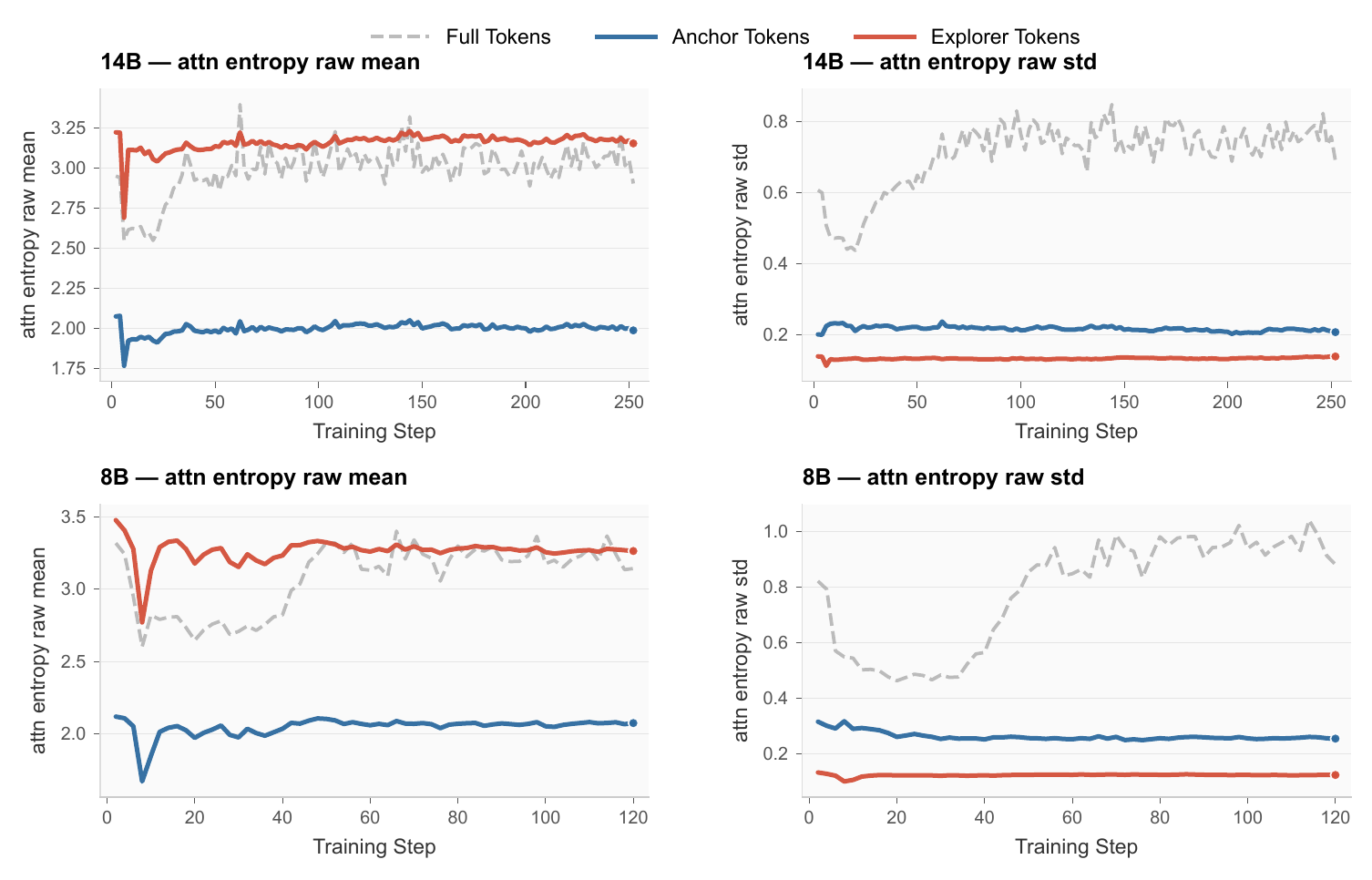}
    \caption{
    Raw attention entropy dynamics.
    We report the mean raw entropy and within-group raw-entropy standard deviation for full tokens, anchor tokens, and explorer tokens on Qwen3-14B and Qwen3-8B.
    The mean separation between explorer, full, and anchor tokens remains visible: explorer tokens generally maintain the highest raw entropy, while anchor tokens maintain the lowest.
    However, raw entropy is sensitive to visible-context length because its theoretical upper bound grows with the number of visible positions.
    As a result, the full-token population exhibits substantially larger variance and stronger non-stationarity under raw entropy.
    This supports using raw entropy as a diagnostic variant rather than as the primary grouping metric.
    }
    \label{fig:app_raw_entropy_dynamics}
\end{figure*}

Raw attention entropy provides the most direct measure of absolute attention spread, but it is also the most sensitive to sequence length. Since the number of visible positions grows during generation, the upper bound of raw entropy also increases with token position. Therefore, raw entropy mixes two effects: genuine attention dispersion and visible-context-length growth.

Despite this confound, Figure~\ref{fig:app_raw_entropy_dynamics} shows that the anchor--explorer ordering remains broadly consistent. Explorer tokens maintain higher raw entropy than anchor tokens, while full tokens lie between them or approach explorer entropy in later training. This indicates that the high-entropy group continues to represent a more diffuse support regime under the unnormalized metric.

The standard-deviation curves further illustrate why raw entropy is not ideal as the main metric. Full tokens show large variance because they contain the entire entropy spectrum and because raw entropy is strongly affected by token position. In contrast, anchor and explorer subsets are narrower by construction. Thus, raw-entropy variance should not be interpreted as optimization stability or instability. Instead, raw entropy is useful mainly as a robustness check showing that the mean separation is not unique to normalized entropy.

\subsection{Top-$k$ entropy dynamics}
\label{app:topk_entropy_dynamics}

\begin{figure*}[t]
    \centering
    \includegraphics[width=0.95\textwidth]{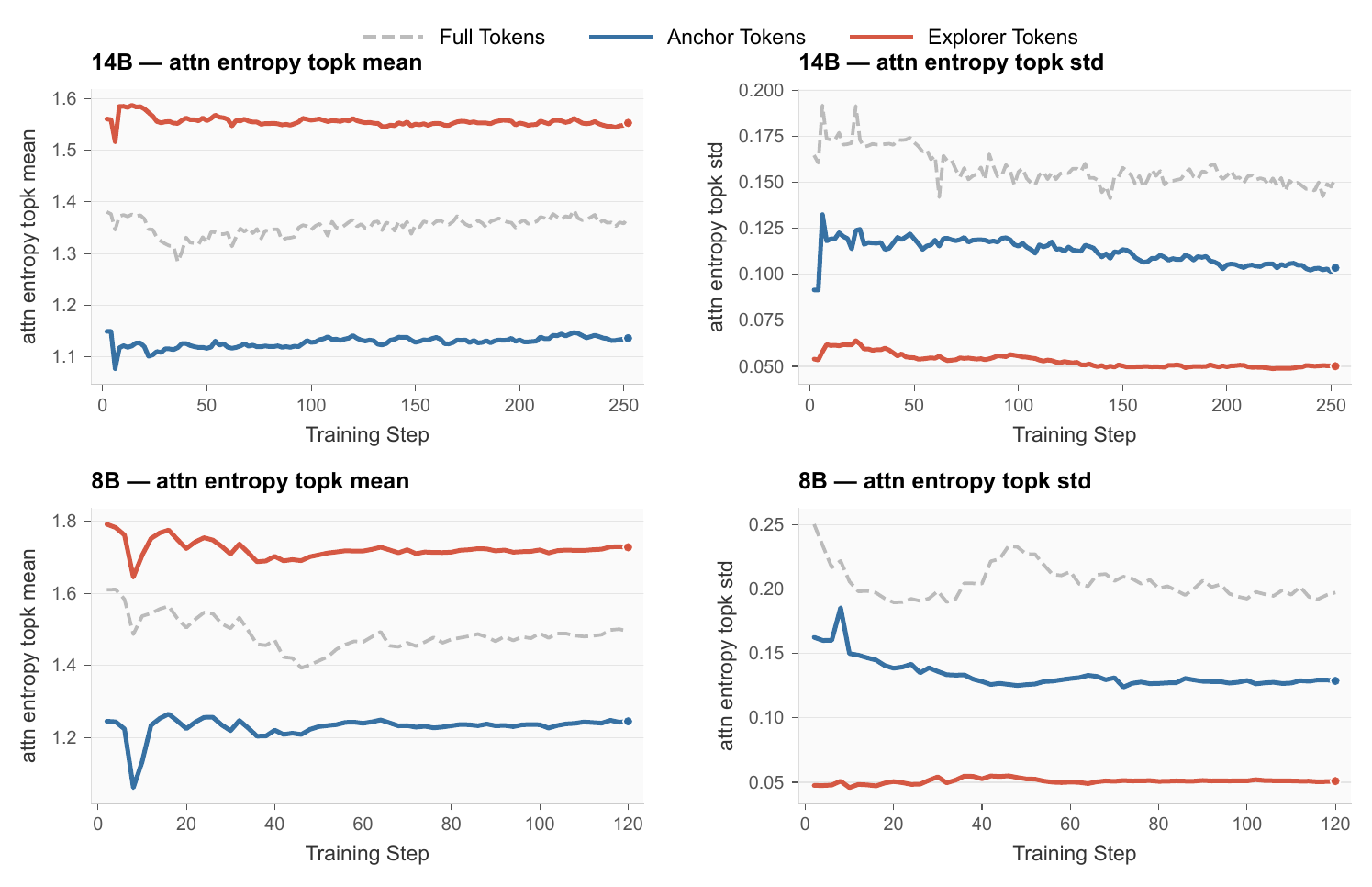}
    \caption{
    Top-$k$ attention entropy dynamics.
    We compute entropy using the top-$k$ attention weights after renormalization, focusing on the dominant attention support.
    Across both Qwen3-14B and Qwen3-8B, explorer tokens maintain the highest top-$k$ entropy, anchor tokens maintain the lowest, and full tokens remain in between.
    This indicates that the anchor--explorer separation is preserved even when measuring only the dominant support structure rather than the full attention distribution.
    The within-group standard deviation again reflects scalar entropy dispersion, not optimization volatility.
    }
    \label{fig:app_topk_entropy_dynamics}
\end{figure*}

Top-$k$ entropy focuses on the dominant attention support by discarding the long tail of very small attention weights. This makes it a useful complement to raw and normalized entropy: if the anchor--explorer separation remains visible under top-$k$ entropy, then the distinction is not driven only by low-mass attention tails.

Figure~\ref{fig:app_topk_entropy_dynamics} shows that this separation is indeed preserved. Explorer tokens consistently have higher top-$k$ entropy, indicating that even among the dominant attended positions, they aggregate over a broader support set. Anchor tokens remain low-entropy, indicating concentrated dominant support. Full tokens remain between the two groups.

The standard-deviation curves show that explorer tokens can occupy a relatively narrow high-entropy band under top-$k$ entropy. This is consistent with the main normalized-entropy dynamics: explorer tokens are not necessarily diverse in scalar entropy value, but they remain distinct in support regime. Their optimization fragility should therefore be attributed to the gradient effects of diffuse support aggregation rather than to larger entropy-value variance.

\subsection{Fixed-position entropy dynamics}
\label{app:fixed256_entropy_dynamics}

\begin{figure*}[t]
    \centering
    \includegraphics[width=0.95\textwidth]{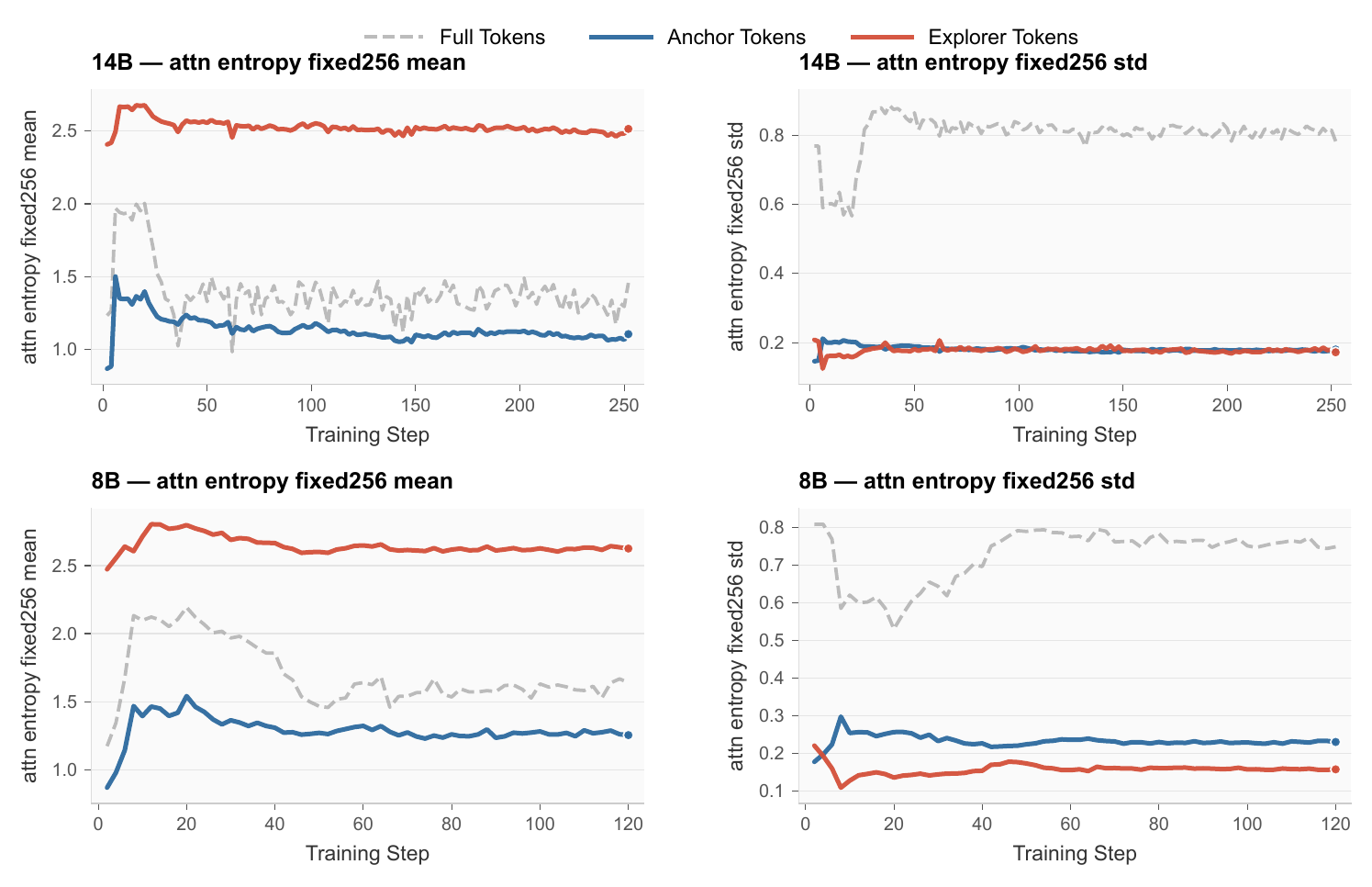}
    \caption{
    Fixed-position entropy dynamics.
    We compute entropy after restricting attention to a fixed set of visible positions, reducing the influence of visible-context-length growth.
    The mean entropy separation remains clear: explorer tokens retain higher fixed-position entropy, anchor tokens retain lower fixed-position entropy, and full tokens lie in between.
    This provides evidence that the anchor--explorer separation is not merely an artifact of increasing context length during generation.
    }
    \label{fig:app_fixed256_entropy_dynamics}
\end{figure*}

Fixed-position entropy controls for the fact that later tokens can attend to more previous positions. By restricting the computation to a fixed number of visible positions, this variant reduces the visible-context-length confound that affects raw entropy and can also influence normalized entropy.

Figure~\ref{fig:app_fixed256_entropy_dynamics} shows that the mean separation persists under this control. Explorer tokens retain the highest fixed-position entropy, anchor tokens retain the lowest, and full tokens remain in between. This indicates that the anchor--explorer distinction is not simply caused by different numbers of visible positions. Instead, it reflects a genuine difference in how concentrated or diffuse the effective attention support is.

The standard-deviation curves again reinforce the same caution: scalar entropy variance is not the same as optimization volatility. Full tokens can have high variance because they cover the whole spectrum. Anchor and explorer subsets can each be relatively narrow in entropy value while still producing very different optimization behavior. The role of fixed-position entropy is therefore not to explain explorer fragility directly, but to verify that the support-regime separation remains after reducing length-related confounds.

\subsection{Summary}

Taken together, the alternative entropy definitions support three conclusions.

First, the mean separation between anchor, full, and explorer tokens is robust. Across raw entropy, top-$k$ entropy, and fixed-position entropy, explorer tokens consistently occupy a higher-entropy regime, anchor tokens occupy a lower-entropy regime, and full tokens lie between them. This supports the claim that RL training preserves an entropy-defined support spectrum rather than collapsing all tokens into a single attention regime.

Second, entropy-value standard deviation should not be used as a direct proxy for optimization instability. Explorer tokens often form a narrow high-entropy band, meaning that their scalar entropy values can be relatively homogeneous. Their training fragility instead comes from optimization-level behavior: diffuse multi-position support can produce trajectory-sensitive gradients, low alignment with the full-token update direction, and collapse-prone selective-training dynamics.

Third, different entropy definitions emphasize different confounds. Raw entropy is sensitive to visible-context length; top-$k$ entropy focuses on dominant support and reduces the influence of long-tail attention mass; fixed-position entropy controls for context-length growth. The fact that the mean separation persists across these variants strengthens the interpretation that attention entropy captures a robust support-concentration axis.

\begin{figure*}[t]
    \centering
    \begin{subfigure}[t]{0.48\textwidth}
        \centering
        \includegraphics[width=\linewidth]{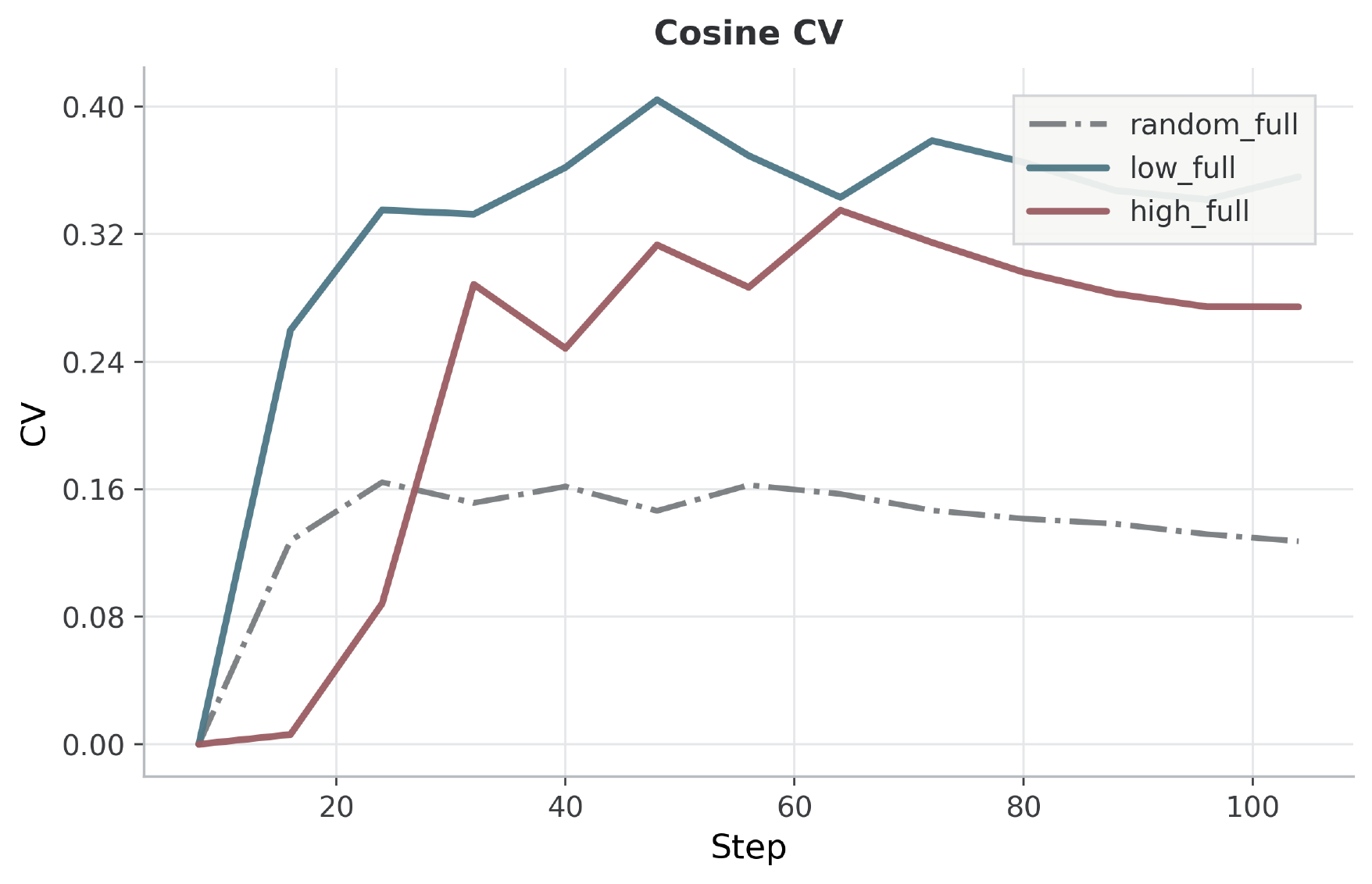}
        \caption{}
    \end{subfigure}
    \hfill
    \begin{subfigure}[t]{0.48\textwidth}
        \centering
        \includegraphics[width=\linewidth]{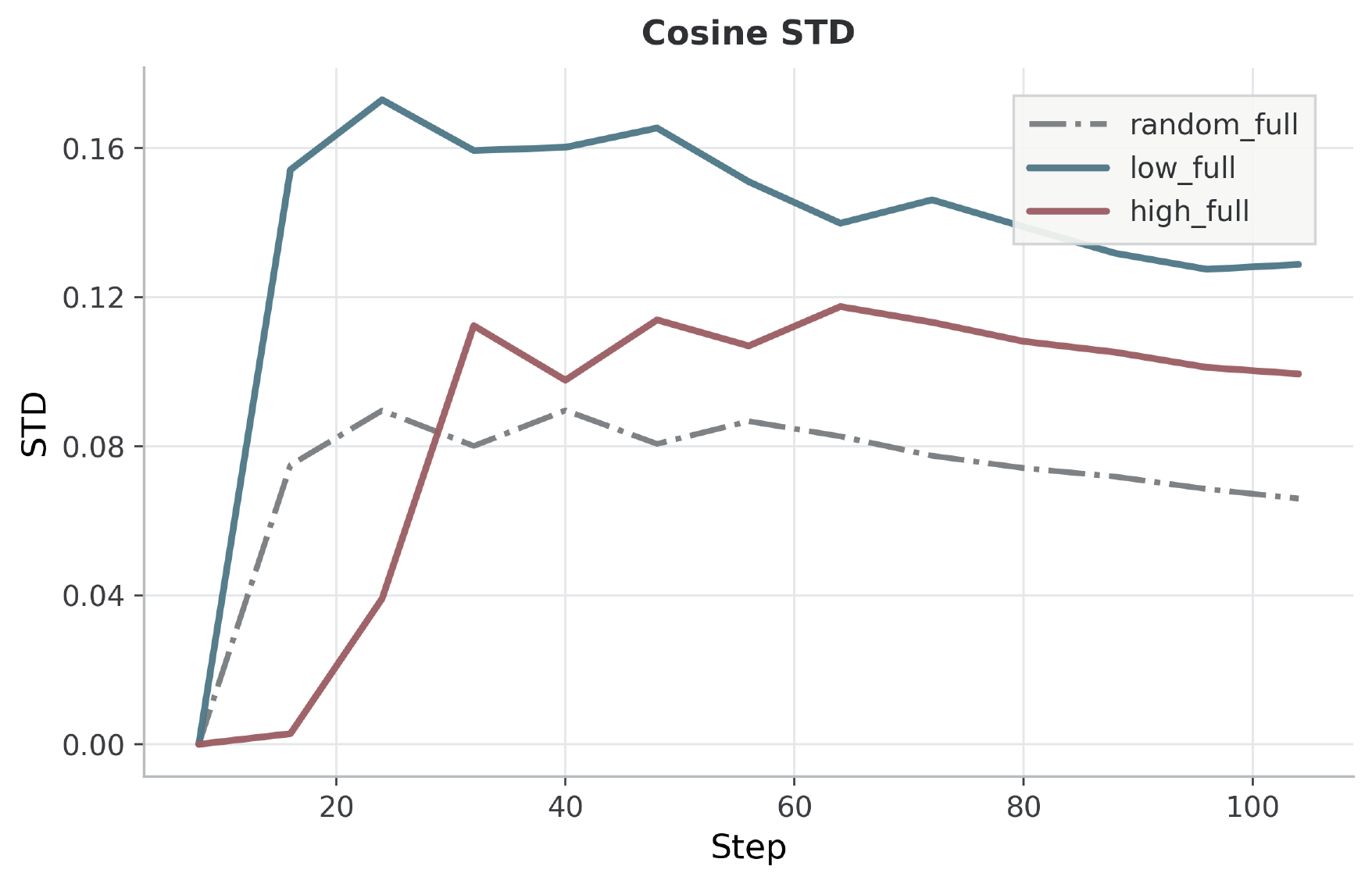}
        \caption{}
    \end{subfigure}
    \caption{
    Online directional statistics of gradient-probe trajectories.
    We summarize the temporal variation of cosine similarity between each
    subset gradient and the full-token gradient using coefficient of variation
    (CV) and standard deviation (STD). Random-20\% subsets show the lowest
    variation, consistent with stable sparse approximation of the full-token
    update. Entropy-defined subsets show larger directional non-stationarity:
    anchor gradients exhibit the strongest temporal drift, while explorer
    gradients remain less stable than random subsets and pass through the
    low-alignment regime shown in Figure~\ref{fig:gradient_probe}.
    }
    \label{fig:gradient_probe_cosine_online_stats}
\end{figure*}

\section{Additional Gradient-Geometry Analyses}
\label{app:gradient_probe_online_stats}

\subsection{Decile-Level Gradient Probe}
\label{app:decile_probe}

The anchor/explorer split is a coarse partition. To test whether this spectrum
is genuinely structured along the entropy axis rather than arising only from an
extreme bottom--top contrast, we perform a decile-level gradient probe. We sort
response tokens by attention entropy within each response, divide them into ten
entropy deciles, compute the gradient induced by each decile under the same
all-token-mean normalization, and compare each decile gradient to the full-token
gradient using the projection-ratio diagnostic.

\begin{figure}[t]
    \centering
    \begin{minipage}{0.5\linewidth}
        \centering
        \includegraphics[width=\linewidth]{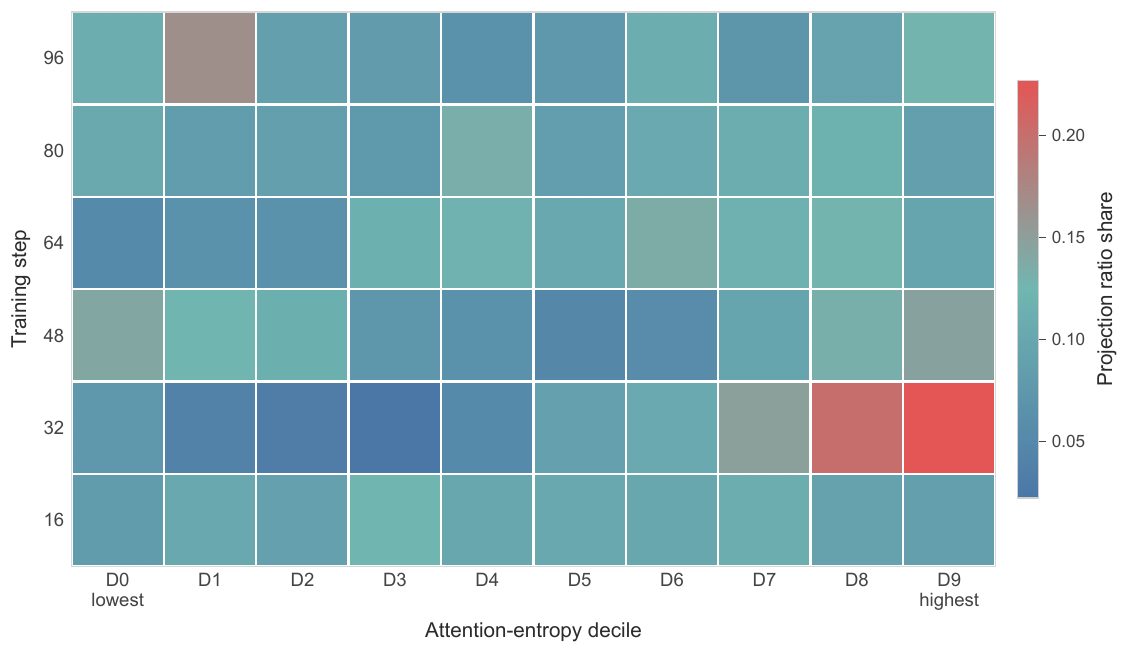}
        \subcaption[a]{Entropy-decile projection-ratio heatmap}
    \end{minipage}
    \hfill
    \begin{minipage}{0.48\linewidth}
        \centering
        \includegraphics[width=\linewidth]{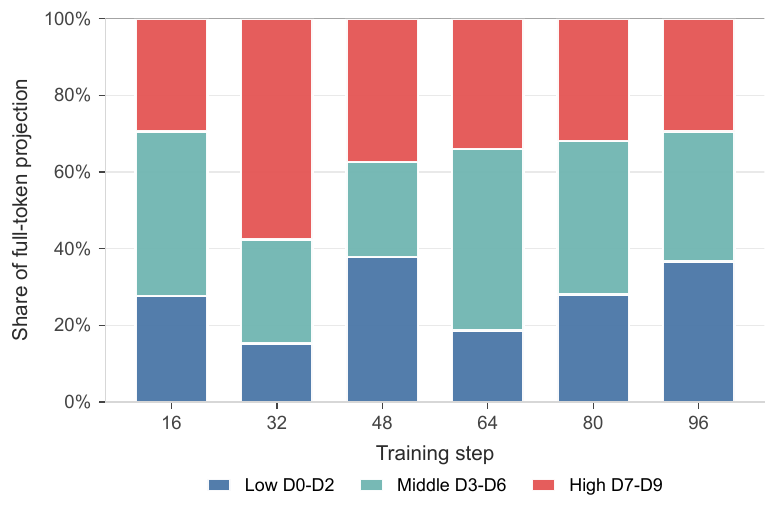}
        \subcaption[b]{Projection share by entropy band}
    \end{minipage}
    \caption{
    \textbf{Decile-level gradient decomposition along the attention-entropy axis.}
    \textbf{(a)}: projection-ratio heatmap over entropy deciles and training steps.
    Each cell shows how much the gradient induced by one entropy decile
    contributes along the full-token gradient direction.
    \textbf{(b)}: the same projection-ratio statistics aggregated into low-entropy
    (D0--D2), middle-entropy (D3--D6), and high-entropy (D7--D9) bands and
    normalized into shares. The results reveal a structured but stage-dependent
    entropy spectrum: upper-entropy deciles often contribute strongly,
    especially in early-to-middle training, but useful gradient contribution
    remains distributed across the spectrum rather than concentrated in a fixed
    subset.
    }
    \label{fig:decile_probe_main}
\end{figure}

Figure~\ref{fig:decile_probe_main} refines the coarse anchor--explorer view by
showing how gradient contribution varies across the full entropy spectrum.
Rather than collapsing tokens into only two groups, we measure the projection
ratio of each entropy decile across training. This reveals a structured but
non-monotonic pattern: upper-entropy deciles often contribute larger effective
update components, especially in early-to-middle training, but low- and
middle-entropy deciles remain non-negligible and become more prominent at other
stages.

This decile-level view supports two conclusions. First, entropy is not merely a
binary partitioning heuristic; it defines a continuous axis along which token
groups exhibit systematically different optimization roles. Second, the spectrum
is stage-dependent rather than strictly monotonic. If useful gradient
contribution were concentrated in a fixed entropy region, one could safely train
only that subset. Instead, the decile probe shows that effective update mass is
distributed across the spectrum and shifts over training, motivating
entropy-aware soft reweighting rather than fixed hard masking.

\subsection{Online Directional-Stability Statistics}
\label{app:gradient_probe_online_stats_detail}

Section~\ref{sec:gradient_quality} analyzes gradient geometry using
time-series diagnostics relative to the full-token update. Here we provide an
additional online summary of directional stability. Specifically, we summarize
the cosine-similarity trajectory between each subset gradient and the
full-token gradient using its standard deviation and coefficient of variation
(CV) across training.

For a diagnostic time series $\{c_s\}$, where $c_s$ denotes the cosine
similarity at probe step $s$, we define the online coefficient of variation as
\[
\mathrm{CV}(s)
=
\frac{
\mathrm{Std}\left(\{c_\tau:\tau \leq s\}\right)
}{
\left|\mathrm{Mean}\left(\{c_\tau:\tau \leq s\}\right)\right|+\epsilon
},
\]
where $\epsilon$ is a small constant used only to avoid division by zero at the
earliest steps. STD measures the absolute fluctuation of directional alignment,
whereas CV measures fluctuation relative to the average alignment level. A large
cosine CV therefore indicates that a subset gradient is directionally
non-stationary relative to its typical agreement with the full-token update.

Figure~\ref{fig:gradient_probe_cosine_online_stats} shows that random-20\%
subsets have the lowest cosine STD and CV throughout training. This is
consistent with their role as noisy but stable sparse approximations of the
full-token update. Entropy-defined subsets exhibit substantially larger
directional non-stationarity. Anchor gradients show the largest cosine
variation, which is consistent with the trajectory-level observation that they
are strongly aligned early in training but progressively lose alignment as
training proceeds. Thus, the large online variation of anchors reflects a
systematic drift in their optimization role rather than purely stochastic
instability.

Explorer gradients also show higher cosine variation than random subsets,
although their STD/CV is lower than that of anchors in this statistic. This
suggests that explorer-only instability should not be characterized simply as
the largest short-term fluctuation in cosine similarity. Rather, together with
the cosine trajectory in Figure~\ref{fig:gradient_probe}, the result
supports a more nuanced view: random subsets are directionally stable sparse
estimators, anchors undergo strong temporal drift, and explorers provide
directionally less reliable update components that pass through a low-alignment
middle phase.

\section{Additional Details for the Dynamic Entropy-Aware Reweighting Intervention}
\label{app:dynamic_soft_reweighting}

This appendix gives the implementation details for the dynamic entropy-aware
soft-reweighting intervention in Section~\ref{sec:soft_weighting}. The main
paper describes the intervention abstractly as a Low2High transition from
low-entropy emphasis to high-entropy emphasis. Here we provide the exact
instantiation used in our experiments. This schedule is used as a controlled
validation probe for the diagnostic, not as a fully tuned algorithmic recipe.
Static and reverse High2Low controls and remaining schedule sensitivity are
discussed below and in the limitations.

\paragraph{Schedule instantiation.}
We use one fixed continuous weighting rule for the main dynamic-reweighting
runs. The reweighter uses the decision attention entropy for each generated
token: if the captured attention-entropy tensor includes the prompt-final
position followed by response-token positions, the reweighting score for
response token $t$ uses the entropy at the generation step for that token.

For all valid response tokens $\mathcal{V}$ in the current rollout batch,
normalized decision attention entropy is converted into a temperature-scaled
softmax score,
\begin{equation}
\alpha_t
=
\frac{\exp(H_t/\tau-c)}
{\sum_{u\in\mathcal{V}}\exp(H_u/\tau-c)+\epsilon},
\qquad
c=\max_{u\in\mathcal{V}} H_u/\tau .
\end{equation}
We then rescale this score over the same valid-token set,
\begin{equation}
\bar{\alpha}_t
=
\frac{\alpha_t-\min_{u\in\mathcal{V}}\alpha_u}
{\max_{u\in\mathcal{V}}\alpha_u-\min_{u\in\mathcal{V}}\alpha_u+\epsilon},
\end{equation}
and assign the token weight by interpolation:
\begin{equation}
w_t^{(s)}
=
(1-\bar{\alpha}_t)w_{\mathrm{low}}(s)
+\bar{\alpha}_t w_{\mathrm{high}}(s).
\end{equation}
The schedule is not selected separately for individual benchmarks. Let
$S_{\mathrm{warm}}$ denote the warmup length and
$r_s=\min(s/S_{\mathrm{warm}},1)$ denote clipped warmup progress. We instantiate
$w_{\mathrm{low}}(s)$ and $w_{\mathrm{high}}(s)$ as monotonic linear functions
of $r_s$:
\begin{equation}
w_{\mathrm{low}}(s)
=
w_{\mathrm{low}}^{\mathrm{start}}
+
r_s
\left(
w_{\mathrm{low}}^{\mathrm{end}}
-
w_{\mathrm{low}}^{\mathrm{start}}
\right),
\end{equation}
\begin{equation}
w_{\mathrm{high}}(s)
=
w_{\mathrm{high}}^{\mathrm{start}}
+
r_s
\left(
w_{\mathrm{high}}^{\mathrm{end}}
-
w_{\mathrm{high}}^{\mathrm{start}}
\right).
\end{equation}
For the main Low2High intervention, we use
$w_{\mathrm{low}}^{\mathrm{start}}=1.0$,
$w_{\mathrm{low}}^{\mathrm{end}}=0.0$,
$w_{\mathrm{high}}^{\mathrm{start}}=0.0$, and
$w_{\mathrm{high}}^{\mathrm{end}}=1.0$. Thus low-entropy tokens receive larger
weights early in training and high-entropy tokens receive larger weights after
warmup. Intermediate tokens receive intermediate weights determined by their
entropy-softmax scores. The computed weights multiply the token-level GRPO
advantages; the actor loss uses all-token-weighted normalization, so the
denominator remains the number of valid response tokens rather than the sum of
weights.

\subsection{Multi-Seed Robustness of Entropy-Aware Reweighting}
\label{app:soft_reweighting_multiseed}

To assess whether the gain from the entropy-aware reweighting intervention is
driven by a favorable training seed, we repeat the primary controlled
intervention comparison with multiple independent training seeds. This
robustness check uses the fixed Layer-20 mid-layer entropy source, which is the
same entropy source used for the schedule ablations. The shallow- and deep-layer
rows in Table~\ref{tab:dynamic_soft_reweighting_main} are instead used to test
entropy-source sensitivity. All runs use the same base model, training data,
reward verifier, rollout configuration, optimization hyperparameters, and
evaluation protocol as the main experiments; the only differences are the random
seed and, for the reweighting variant, the token-weighting rule.

The Layer-20 row in the main table reports the corresponding mean $\pm$
standard deviation. Table~\ref{tab:soft_reweighting_per_seed} provides the
underlying per-seed benchmark results. This robustness check is intended to
verify that the observed improvement of the controlled mid-layer reweighting
setting is not an artifact of a single favorable run.

\begin{table}[t]
\centering
\small
\caption{
Per-seed benchmark results for full-token DAPO and Layer-20 entropy-aware
reweighting on Qwen3-8B-Base. These runs underlie the mean $\pm$ std values
reported for the Layer-20 row in Table~\ref{tab:dynamic_soft_reweighting_main}.
}
\label{tab:soft_reweighting_per_seed}
\setlength{\tabcolsep}{4pt}
\resizebox{\linewidth}{!}{
\begin{tabular}{llccccc}
\toprule
Method
& Seed
& AIME
& OlympiadBench
& Minerva
& MATH
& Avg. \\
\midrule
Full-token DAPO
& 1
& 4.86
& 35.43
& 24.95
& 72.37
& 34.40 \\

Full-token DAPO
& 2
& 6.01
& 33.35
& 27.32
& 72.73
& 34.85 \\

Full-token DAPO
& 3
& 6.62
& 33.94
& 24.80
& 70.30
& 33.92 \\

\midrule
Entropy reweighting (Layer 20)
& 1
& 10.45
& 37.42
& 27.88
& 75.06
& 37.70 \\

Entropy reweighting (Layer 20)
& 2
& 10.45
& 35.17
& 27.09
& 72.40
& 36.28 \\

Entropy reweighting (Layer 20)
& 3
& 8.35
& 36.85
& 29.69
& 75.00
& 37.47 \\
\bottomrule
\end{tabular}
}
\end{table}

The multi-seed results provide a robustness check for the fixed Layer-20
intervention comparison. Since Layer-20 entropy-aware reweighting improves the
mean held-out average over full-token DAPO across seeds, this supports the
interpretation that the attention-entropy diagnostic captures actionable
token-level structure rather than a single-run artifact. At the same time, we
treat this result as evidence for the diagnostic and the optimization-spectrum
analysis, rather than as an exhaustive claim that the particular hand-specified
schedule or entropy-source layer is globally optimal.

\begin{table}[t]
\centering
\small
\caption{Implementation details of the dynamic entropy-aware reweighting
intervention. The same configuration is used across all benchmarks for a given
run.}
\label{tab:dynamic_reweighting_hparams}
\begin{tabular}{ll}
\toprule
Item & Setting \\
\midrule
Entropy score & Normalized attention entropy \\
Entropy source & Decision entropy for each generated token \\
Entropy layer & Layer 20 for controlled mid-layer runs; varied only in entropy-source comparisons \\
Softmax temperature & $0.8$ \\
Weight signal & Batch-wise softmax over valid response-token entropy \\
Interpolation score & Min--max rescaled softmax score over valid tokens \\
Weight map & $(1-\bar{\alpha}_t)w_{\mathrm{low}}+\bar{\alpha}_t w_{\mathrm{high}}$ \\
Token coverage & All valid response tokens receive continuous weights \\
Main schedule & Low2High \\
Warmup length & $80$ steps \\
$w_{\mathrm{low}}$ init $\rightarrow$ target & $1.0 \rightarrow 0.0$ \\
$w_{\mathrm{high}}$ init $\rightarrow$ target & $0.0 \rightarrow 1.0$ \\
Applied quantity & Token-level GRPO advantages \\
Loss normalization & All-token-weighted normalization \\
Schedule selection & Fixed across benchmarks \\
\bottomrule
\end{tabular}
\end{table}

\subsection{Reweighting Schedule Ablation}
\label{app:soft_reweighting_variant_controls}

We further ablate the reweighting schedule used in the entropy-aware
soft-reweighting intervention under the fixed Layer-20 entropy source. All
variants use the same decision-attention entropy source, softmax temperature,
continuous advantage reweighting rule, and all-token-weighted normalization as
the controlled mid-layer intervention; they differ only in how the low- and
high-entropy endpoint weights are assigned over training.

Table~\ref{tab:soft_reweighting_schedule_ablation} compares static and dynamic
weighting strategies on Qwen3-8B-Base. The static variants test whether the
benefit can be explained by a fixed preference for one entropy regime. Static
anchor-biased reweighting emphasizes low-entropy tokens throughout training and
performs close to the full-token DAPO baseline, suggesting that a persistent
low-entropy preference preserves stability but does not sufficiently expand
coverage. Static explorer-biased reweighting improves the average mainly through
MATH, but does not improve AIME, indicating that emphasizing high-entropy tokens
alone is not a reliable strategy for harder reasoning benchmarks.

The dynamic variants test whether the temporal order of entropy-aware allocation
matters. Among the Layer-20 schedules we tested, the \textbf{Low2High} schedule
is the main controlled intervention and the most effective schedule within this
fixed-layer ablation: it achieves the best average in this ablation, improving
the held-out average from 34.39 to 37.15, with gains on all four benchmarks in
the Layer-20 setting. This Layer-20 schedule result
corresponds to the mid-layer row in Table~\ref{tab:dynamic_soft_reweighting_main};
the shallow- and deep-layer rows vary the entropy source rather than the
schedule. We interpret the schedule result as evidence that gradually
reallocating weight along the attention-entropy spectrum can be more effective
than using a fixed endpoint bias. In particular, starting from the lower-entropy
side provides a more stable early optimization regime, while later increasing
the emphasis on higher-entropy tokens allows the model to access broader but
more volatile signals after the training trajectory has become less fragile.

The reverse \textbf{High2Low} schedule is an ablation. It also improves the
average over the full-token baseline and obtains the highest average when AIME
is excluded, showing that the gain is not simply tied to one fixed static bias.
However, it substantially underperforms on AIME. This suggests that exposing
training to high-entropy, directionally volatile signals too early can make the
initial optimization trajectory less reliable, which may prevent the schedule
from improving consistently across all benchmarks.

Overall, these ablations support a conservative conclusion: entropy-aware
temporal reallocation is useful, and Low2High is the strongest schedule among
the tested Layer-20 variants. However, this experiment is intended as an
exploratory validation intervention rather than an exhaustive search over
scheduling strategies. We did not further explore whether alternative transition
shapes, transition timings, entropy thresholds, or weighting functions could
yield stronger results. Therefore, the result should not be interpreted as
evidence that Low2High is globally optimal; instead, it shows that
attention-entropy-aware temporal reallocation can be beneficial under the tested
fixed-layer schedules.

\begin{table}[t]
\centering
\small
\caption{
Soft-reweighting schedule ablation on Qwen3-8B-Base using the fixed Layer-20
entropy source.
``Avg.'' denotes the arithmetic mean over AIME, OlympiadBench, Minerva, and MATH.
``Avg. w/o AIME'' denotes the average over OlympiadBench, Minerva, and MATH only.
}
\label{tab:soft_reweighting_schedule_ablation}
\setlength{\tabcolsep}{4pt}
\resizebox{\linewidth}{!}{
\begin{tabular}{@{}lccrrrrrr@{}}
\toprule
Variant 
& Early emphasis 
& Late emphasis 
& AIME 
& OlympiadBench 
& Minerva 
& MATH 
& Avg. 
& Avg. w/o AIME \\
\midrule
Full-token DAPO 
& uniform 
& uniform 
& 5.83 
& 34.24 
& 25.69 
& 71.80 
& 34.39 
& 43.91 \\

Static anchor-biased 
& low entropy 
& low entropy 
& 5.78 
& 34.18 
& 25.80 
& 71.23 
& 34.25 
& 43.74 \\

Static explorer-biased 
& high entropy 
& high entropy 
& 5.59 
& 33.97 
& 25.88 
& 77.60 
& 35.76 
& 45.82 \\

\rowcolor{rowblue}
Dynamic Low2High (main) 
& low entropy 
& high entropy 
& \textbf{9.75}
& 36.48 
& 28.22 
& 74.15 
& \textbf{37.15}
& 46.28 \\

\rowcolor{gray!15}
Dynamic High2Low (ablation) 
& high entropy 
& low entropy 
& 3.78 
& \textbf{38.84}
& \textbf{29.58}
& 75.18 
& 36.85 
& \textbf{47.87} \\
\bottomrule
\end{tabular}
}
\end{table}

\begin{figure}[t]
    \centering
    \begin{minipage}{0.48\linewidth}
        \centering
        \includegraphics[width=\linewidth]{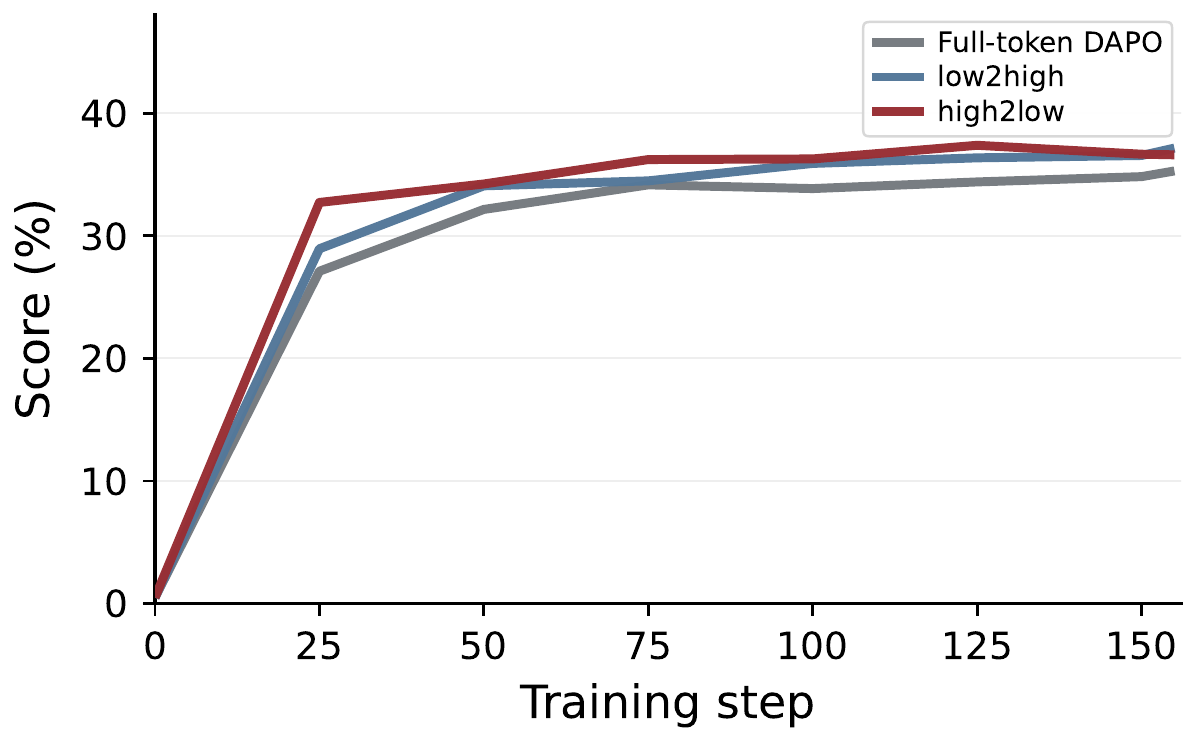}
        \vspace{-0.5em}
        \centerline{\small (a) Held-out average}
    \end{minipage}
    \hfill
    \begin{minipage}{0.48\linewidth}
        \centering
        \includegraphics[width=\linewidth]{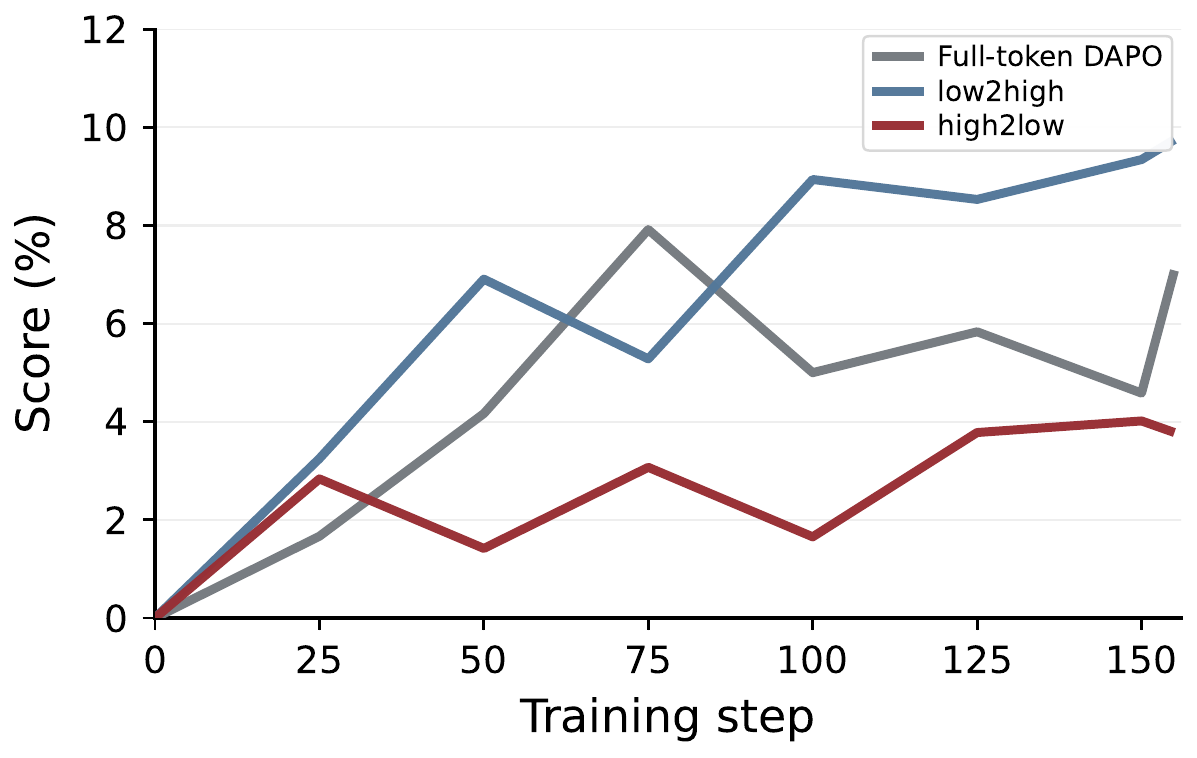}
        \vspace{-0.5em}
        \centerline{\small (b) AIME}
    \end{minipage}
    \caption{
    Training-curve comparison between the Low2High schedule and the reverse
    High2Low schedule under the fixed Layer-20 entropy source. The held-out
    average shows that both dynamic schedules can improve over full-token DAPO
    during training, indicating that entropy-aware temporal reallocation is more
    useful than a fixed uniform allocation. However, the AIME trajectory reveals a
    sharper distinction: Low2High achieves stronger late-stage AIME performance,
    whereas High2Low remains substantially weaker on this harder benchmark. This
    supports the interpretation that emphasizing high-entropy tokens too early may
    expose the model to more volatile signals before a stable optimization
    backbone has formed.
    }
    \label{fig:schedule_training_curves}
\end{figure}

\paragraph{Training-curve comparison.}
Figure~\ref{fig:schedule_training_curves} provides a trajectory-level view of the
dynamic schedule ablation in Table~\ref{tab:soft_reweighting_schedule_ablation}.
On the held-out average, both dynamic schedules improve over the full-token DAPO
baseline during training, suggesting that temporally reallocating token-level
learning signal along the attention-entropy spectrum can be beneficial. The
reverse High2Low schedule rises quickly and remains competitive on the aggregate
average, which indicates that the benefit is not simply due to one static
low-entropy or high-entropy bias.

The AIME trajectory, however, reveals an important difference between the two
temporal orders. Low2High consistently achieves stronger late-stage AIME
performance, while High2Low remains much weaker on this harder benchmark. This
pattern is consistent with the anchor--explorer interpretation: starting from
low-entropy tokens provides a more stable early optimization regime, whereas
shifting weight toward higher-entropy tokens later allows the model to access
broader but more volatile reasoning signals after the trajectory has become less
fragile. In contrast, emphasizing high-entropy tokens from the beginning can
hurt the hardest-benchmark trajectory, even if it improves the aggregate average
on easier or less brittle benchmarks.

We therefore interpret Figure~\ref{fig:schedule_training_curves} as
trajectory-level support for the Layer-20 schedule-ablation conclusion rather
than as an exhaustive schedule-selection result. Low2High is the strongest
schedule among the tested Layer-20 variants, especially on AIME, but we do not
claim that it is globally optimal over all possible transition shapes, timings,
entropy temperatures, or weighting functions.

\subsection{Entropy-Source Layer Sensitivity}
\label{app:entropy_source_sensitivity}

The main analyses use Layer 20 as a fixed mid-layer attention-entropy probe.
This choice is made to keep token grouping, support-concentration statistics,
entropy dynamics, gradient-geometry diagnostics, and intervention comparisons
under a single consistent mechanistic lens. It is not intended to claim that
Layer 20 is the globally optimal entropy source.

To examine whether the intervention depends strongly on this fixed-layer choice,
we compare representative shallow, middle, and final-layer entropy sources:
Layer 8, Layer 20, and Layer 31. Layer indices follow our implementation
convention, where Layer 31 corresponds to the final transformer layer. All other
settings are kept unchanged, including the training data, reward verifier,
rollout configuration, optimization hyperparameters, Low2High reweighting
schedule, and strict exact-match evaluation protocol. The only difference is the
layer from which normalized attention entropy is computed.

The corresponding results are reported in
Table~\ref{tab:dynamic_soft_reweighting_main}. The entropy-aware intervention
remains effective under all three entropy-source layers, suggesting that the
diagnostic is not tied to the specific Layer-20 probe used in the main
mechanistic analyses. All three representative entropy sources improve over
full-token DAPO on the held-out average and on every benchmark, but they expose
different optimization profiles: Layer 8 achieves the strongest average and
Minerva performance, Layer 20 gives the strongest AIME and MATH performance
among the entropy-source rows, and Layer 31 gives the strongest OlympiadBench
performance. This suggests that optimization-relevant attention-entropy
structure can emerge across the depth of the model, while different depths may
emphasize different parts of the reasoning signal.

We interpret this result conservatively. Since this is a representative
three-layer check rather than a full layer-wise sweep, it should not be read as
evidence that Layer 8 is universally optimal. Rather, it shows that
entropy-source selection is an important design dimension and that useful
attention-entropy signals can appear outside the fixed Layer-20 probe. A
systematic study of layer selection, neighboring-layer stability, and multi-layer
entropy aggregation is left for future work.

\subsection{Intervention Transfer to Additional Models}
\label{app:soft_reweighting_additional_models}

The main paper reports Qwen3-8B-Base as the primary model to keep the comparison
controlled. To examine whether the effect of entropy-aware soft reweighting is
specific to this model, we further evaluate the same intervention on additional
model families under the same data, reward, rollout, and evaluation protocol.

Table~\ref{tab:soft_additional_models} provides supporting transfer checks on
the tested models. On Qwen3-14B-Base, soft reweighting
improves the held-out average from 37.49 to 41.06, with AIME increasing from
10.69 to 11.83. On Qwen2.5-7B, it improves the average from 30.82 to 34.47, with
AIME increasing from 5.27 to 7.61. These results suggest that the benefit of
entropy-aware soft reweighting is not limited to the primary Qwen3-8B-Base
setting, but also appears on both a larger Qwen3 model and an earlier Qwen2.5
model.

We emphasize that these transfer results are intended as controlled supporting
evidence rather than a claim of universal effectiveness across all model
families. Nevertheless, the consistent gains across model scale and generation
provide additional support for the anchor--explorer interpretation and for using
attention entropy as a practical signal for token-level reweighting.

\begin{table}[t]
\centering
\small
\caption{Supporting transfer checks for the entropy-aware reweighting
intervention on additional models. Results use the same training data, reward,
rollout, and evaluation protocol as the primary Qwen3-8B-Base comparison. These
runs are intended as supporting evidence rather than a claim of universal
effectiveness across model families.}
\label{tab:soft_additional_models}
\setlength{\tabcolsep}{3pt}
\begin{tabular}{@{}p{0.22\linewidth}p{0.14\linewidth}p{0.14\linewidth}p{0.18\linewidth}p{0.14\linewidth}@{}}
\toprule
Model & Full-token avg. & Soft avg. & AIME(Full-token) & AIME(Soft) \\
\midrule
Qwen3-14B-Base & 37.49 & 41.06 & 10.69 & 11.83 \\
Qwen2.5-7B  & 30.82 & 34.47 & 5.27 & 7.61\\
\bottomrule
\end{tabular}
\end{table}

\section{Broader Impact.}
\label{app:broader_impact}
This work studies token-level diagnostics and allocation strategies for
RL-based reasoning post-training. Potential positive impacts include improving
the interpretability, stability, and compute efficiency of reasoning-model
training. Potential negative impacts follow from the broader improvement of
reasoning capabilities in language models, which may make misuse such as
automated problem solving, deceptive reasoning, or unsafe downstream deployment
more effective if such models are released without appropriate safeguards. Our
work does not release a new model, dataset, or user-facing system, and all
experiments are conducted on mathematical-reasoning benchmarks. We recommend
that any deployment of entropy-aware training methods be accompanied by standard
model-safety evaluation, misuse monitoring, and domain-specific risk assessment.

\section{Existing Assets and Licenses}
\label{app:asset_licenses}

This work uses existing open models, datasets, benchmarks, and software
frameworks obtained from Hugging Face Hub or the corresponding official
repositories. We credit the original creators through citations in the main
paper and appendix, and we use these assets only for research experiments under
the licenses or terms specified by their Hugging Face model cards, dataset cards,
or official repositories. We do not redistribute third-party model checkpoints,
datasets, or code as new paper assets.

The main existing assets used in this work include the Qwen model family, VeRL,
the DAPO recipe, DeepScaleR, MATH, AIME, OlympiadBench, and Minerva Math. In
particular, the Qwen model checkpoints are used from the official Qwen Hugging
Face repositories under the Apache-2.0 license. VeRL is used under the
Apache-2.0 license. The DeepScaleR-Preview-Dataset Hugging Face release is
listed under the MIT license. The MATH data are loaded from the Hugging Face
\texttt{HuggingFaceH4/MATH} release, which is listed under the MIT
license. The AIME evaluation data are loaded from the Hugging Face
\texttt{BytedTsinghua-SIA/AIME-2024} release, which is listed under the
Apache-2.0 license. OlympiadBench is loaded from the Hugging Face
\texttt{Hothan/OlympiadBench} release, which is listed under the Apache-2.0
license. Minerva Math is loaded from the Hugging Face
\texttt{svc-huggingface/minerva-math} release, which is listed under the MIT
license.

The DAPO recipe is used as an algorithmic and implementation reference for
RLVR training, together with the VeRL-based training framework. We do not use or
redistribute any third-party asset beyond the research-use scope specified by
the corresponding Hugging Face cards or official repositories.

\section*{LLM Usage Disclosure}
\label{app:LLM_usage}
We used Claude as an auxiliary tool to assist with language editing,
exploratory data analysis, and visualization scripting. The model was used to
help improve grammar, spelling, and word choice; inspect experimental logs;
draft plotting code; and generate figure layouts. All numerical results,
statistical summaries, plots, and scientific interpretations reported in the
paper were verified by the authors against the original experiment outputs.
Claude was not used to generate training data, labels, model outputs for
evaluation, or paper claims without author verification.

\clearpage
\newpage

\end{document}